\documentclass[graybox]{svmult}

\usepackage{mathptmx}       
\usepackage{helvet}         
\usepackage{courier}        
\usepackage{type1cm}                                    
\usepackage{makeidx}         
\usepackage{graphicx}                                    
\usepackage{multicol}        
\usepackage[bottom]{footmisc}
\usepackage{bm}
\usepackage{amssymb}
\usepackage{amsmath}
\usepackage{epsfig}

\makeindex
\begin{document}

\title*{Geometric Algebra Model of Distributed Representations}
\author{Agnieszka Patyk}
\institute{Agnieszka Patyk \email{patyk@mif.pg.gda.pl}
\at Faculty of Applied Physics and Mathematics,\\
Gda\'nsk University of Technology, 80-952 Gda\'nsk, Poland,\\
Centrum Leo Apostel (CLEA), Vrije Universiteit Brussel, 1050 Brussels, Belgium.}
\maketitle

\abstract*{Formalism based on GA is an alternative to distributed representation models developed so far --- Smolensky's tensor product, Holographic Reduced Representations (HRR) and Binary Spatter Code (BSC). Convolutions are replaced by geometric products, interpretable in terms of geometry which seems to be the most natural language for visualization of higher concepts. This paper recalls the main ideas behind the GA model and investigates recognition test results using both inner product and a clipped version of matrix representation. The influence of accidental blade equality on recognition is also studied. Finally, the efficiency of the GA model is compared to that of previously developed models.}

\abstract{Formalism based on GA is an alternative to distributed representation models developed so far --- Smolensky's tensor product, Holographic Reduced Representations (HRR) and Binary Spatter Code (BSC). Convolutions are replaced by geometric products, interpretable in terms of geometry which seems to be the most natural language for visualization of higher concepts. This paper recalls the main ideas behind the GA model and investigates recognition test results using both inner product and a clipped version of matrix representation. The influence of accidental blade equality on recognition is also studied. Finally, the efficiency of the GA model is compared to that of previously developed models.}


\section{Introduction}
\label{sec:1}
Since the early 1980s a new idea of representing knowledge has emerged by the name of distributed representation\index{distributed representation}. It has been the answer to the problems of recognition, reasoning and language processing --- people accomplish these everyday tasks effortlessly, often with only noisy and partial information, while computational resources required for these assignments are enormous.
To this day many models have been built, in which arbitrary variable bindings, short sequences of various lengths and predicates are all usually represented as fixed-width high dimensional vectors that encode information throughout the elements. In 1990 Smolensky \cite{Smo} described how tensor product algebra provides a framework for the distributed representation of recursive structures. Unfortunately, Smolensky's tensor product does not meet all criteria of reduced representations as the size of the tensor increases with the size of the structure. Nevertheless, Smolensky and Dolan \cite{SmoDol} have shown, that tensor product algebra can be used in some architectures as long as the size of a tensor is restricted. In 1994 Plate \cite{Plate03} worked up his Holographic Reduced Representation (HRR)\index{HRR} that uses circular convolution and vector addition to combine vectors representing elements of a domain in hierarchical structures. Elements are represented by randomly chosen high-dimensional vectors. A vector representing a structure is of the same size as the vectors representing the elements it contains. In 1997 Pentti Kanerva \cite{Kan1,Kan2} introduced Binary Spatter Code (BSC) that is very similar to HRR and is often referred to as a form of HRR. In BSC\index{BSC} objects are represented by binary vectors and the boolean exclusive OR is used instead of convolution. The clean-up memory is an important part of any distributed representation model as an auto-associative collection of all atomic objects and complex statements produced by that system. Given a noisy extracted vector such structure must be able to recall the most similar item stored or indicate, that no matching object had been found.

The geometric algebra\index{geometric algebra} (GA) model, which is the focus of this paper, is an alternative to models developed so far. It has been inspired by a well-known fact, that most people think in pictures, i.e. two- and three-dimensional shapes, not by using sequences of ones and zeroes. As far as brain functions are concerned, geometric computing has been applied thus far only in the context of primate visual system (\cite{Bay1}, Chapters 1 and 2). 

In the GA model convolutions are replaced by geometric products and superposition is performed by ordinary addition. Sentences are represented by multivectors --- superpositions of blades. The concept of GA first appeared in the 19th century works of Grassmann and Clifford, but was abandoned for almost a century until Hestenes brought up the subject in \cite{Hes} and \cite{HesSob}. The Hestenes system has recently found applications in quantum computation (Czachor \it et al. \rm \cite{Cza1}--\cite{Cza4}, \cite{Cza5}), which appears to be a promising leap from cognitive systems based on traditional computing. 

Section \ref{sec:2} of this paper recalls basic operations that can be performed on blades and multivectors, using the example Kanerva \cite{Kan2} gave to illustrate BSC. For further details on multivectors as well as interesting exercises the reader may refer to \cite{Dorst, Loun}. Section \ref{sec:3} gives rise to discussion about various ways of asking questions and investigates the percentage of correctly recognized items under two possible constructions. Section \ref{sec:4} introduces measures of similarity based not on only the inner product of a multivector, but also on its matrix representation. Finally, Section \ref{sec:5} studies the influence of accidental blade equality on recognition and Section \ref{sec:6} compares the performance of the GA model with HRR and BSC.


\section{Geometric Algebra Model}
\label{sec:2}

Distributed representation models developed so far were based on long binary or real vectors. However, most people tend to think by pictures, not by sequences of numbers. Therefore geometric algebra with its ability to describe shapes is the most natural language to mimic human thought process and to represent atomic objects as well as complex sentences. Furthermore, geometric product of two multivectors is geometrically meaningful, unlike the convolution or a binary exclusive OR operation performed on two vectors.

In this paper we consider the $C\ell_n$ algebra generated by orthonormal vectors $b_i=\{0,\dots,0,1,0\dots,0\}$ for $i\in\{1,\dots,n\}$. The inner product\index{inner product} used throughout the paper is an extension of the inner product $\langle \cdot | \cdot \rangle$ from the Euclidean space $\mathbb{R}^n$. For blades $X_{<k>} = x_1 \wedge \dots \wedge x_k$ and $Y_{<l>} = y_1 \wedge \dots \wedge y_l$ the inner product $\cdot: C\ell_n \times C\ell_n \rightarrow \mathbb{R}$ is defined as
\begin{eqnarray}
\langle X_{<k>} | Y_{<l>} \rangle &=& \left| 
\begin{matrix}
\langle x_1 | y_l\rangle & \langle x_1| y_{l-1}\rangle &\cdots&\langle x_1| y_1 \rangle\cr
\langle x_2| y_l\rangle & \langle x_2| y_{l-1}\rangle &\cdots&\langle x_2| y_1 \rangle\cr
\vdots& &\ddots&\vdots \cr
\langle x_k| y_l\rangle & \langle x_k| y_{l-1}\rangle &\cdots& \langle x_k| y_1\rangle \cr
\end{matrix}
\right|\ \textrm{ for } k=l,\\
\langle X_{<k>} | Y_{<l>} \rangle&=& 0\ \textrm{ for } k\neq l
\end{eqnarray}
and is extended by linearity to the entire algebra. 

Originally, the GA model was developed as a geometric analogue of BSC and HRR and was described by Czachor, Aerts and De Moor in \cite{Cza1} and \cite{Cza4}. Before switching from geometric product to BSC and HRR one has to realize, that geometric product is a projective representation of boolean exclusive OR\index{XOR}. Let $x_1\dots x_n$ and $y_1\dots y_n$ be binary representations of two $n$-bit numbers $x$ and $y$ and let $c_x=c_{x_1\dots x_n}=b_1^{x_1}\dots b_n^{x_n}$ and $c_y=c_{y_1 \dots y_n}=b_1^{y_1}\dots b_n^{y_n}$ be their corresponding blades, $b_i^0$ being equal to \textbf{1}. The following examples show that geometric product of two blades\index{blades} $c_x$ and $c_y$ equals, up to a sign, $c_{x\oplus y}$
\begin{eqnarray}
b_1b_1
&=&c_{10\dots 0}c_{10\dots 0}={\bf 1}=c_{0\dots 0}=c_{(10\dots 0)\oplus (10\dots 0)},\\
b_1b_{12}
&=&c_{10\dots 0}c_{110\dots 0}=b_1b_1b_2=b_2=c_{010\dots 0}=c_{(10\dots 0)\oplus (110\dots 0)},\\
b_{12}b_1
&=&c_{110\dots 0}c_{10\dots 0}=b_1b_2b_1=-b_2b_1b_1=-b_2\nonumber\\
&=&
-c_{010\dots 0}=-c_{(110\dots 0)\oplus(10\dots 0)}=(-1)^D c_{(110\dots 0)\oplus(10\dots 0)},
\end{eqnarray}
the number $D$ being calculated as follows
\begin{eqnarray}
D=y_1(x_2+\dots+x_n)+y_2(x_3+\dots+x_n)+\dots+y_{n-1}x_n=\sum_{k<l}y_kx_l.
\end{eqnarray}

The original BSC is illustrated by an example taken from \cite{Kan2, Cza4} --- atomic objects are represented by randomly chosen strings of bits, ``$\oplus$" is a componentwise addition mod $2$ and ``$\boxplus$" represents a thresholded sum producing a binary vector --- the threshold is set at one half of sentence chunks and a random string of bits is added in case of an even number of sentence chunks to break the tie. The encoded record is
\begin{eqnarray}
PSmith
&=&
(name\oplus Pat)\boxplus( sex\oplus male)\boxplus( age\oplus 66),
\end{eqnarray}
and the decoding of $name$ uses the involutive nature of XOR
\begin{eqnarray}
 Pat'
&=&
name\oplus PSmith\nonumber\\
&=&
 name\oplus
\big[( name\oplus Pat)\boxplus( sex\oplus male)\boxplus( age\oplus 66)\big]
\nonumber\\
&=&
 Pat
\boxplus( name\oplus sex\oplus male)\boxplus( name\oplus age\oplus 66)\nonumber\\
&=&
 Pat
\boxplus
{\rm noise}
\to  Pat.
\end{eqnarray}
In order to switch from BSC to HRR, the $x\mapsto c_x$ map described in \cite{Cza4} is used. 
\vspace{0.5cm}
$\\$
In the GA model, roles and fillers are represented by randomly chosen blades
\begin{eqnarray}
 PSmith&=& name\ast Pat+ sex\ast male+ age\ast 66.
\end{eqnarray}
The ``+" is an ordinary addition and ``$\ast$" written between clean-up memory items denotes the geometric product --- this notation will be traditionally omitted when writing down operations performed directly on blades and multivectors. The ``$^{+}$" written in the superscript denotes the reversion of a blade or a multivector. The whole record now corresponds to a multivector
\begin{eqnarray}
 PSmith
&=&c_{a_1\dots a_n}c_{x_1\dots x_n}+c_{b_1\dots b_n}c_{y_1\dots y_n}+c_{c_1\dots c_n}c_{z_1\dots z_n},
\end{eqnarray}
and the decoding operation ``$\sharp$" of $name$ with respect to $PSmith$ is defined as follows
\begin{eqnarray}
 PSmith\ \sharp\ name &=& name^{+}\ast PSmith \label{eq:pat1}\\
&=&
c_{a_1\dots a_n}^{+}
\big[
c_{a_1\dots a_n}c_{x_1\dots x_n}+c_{b_1\dots b_n}c_{y_1\dots y_n}+c_{c_1\dots c_n}c_{z_1\dots z_n}\big]\nonumber\\
&=&
c_{x}
\pm
c_{a\oplus b\oplus y}
\pm
c_{a\oplus c\oplus z} \label{eq:pat2}\\
&=&
 Pat
+
{\rm noise}.\label{eq:pat3}
\end{eqnarray}
It remains to employ the cleanup memory to find the element closest to $Pat'$ --- similarity is computed by the means of the inner (scalar) product. When using the decoding symbol ``$\sharp$" we assume that the reader knows which model is used at the time. Therefore, there will be no variations of the ``$\sharp$" symbol in BSC, HRR or in two possible GA models (depending on the way of asking a question).

For an actual example let us choose the following representation for roles and fillers\index{roles and fillers} of $PSmith$
\begin{eqnarray}
\left.
\begin{array}{rcl}
 Pat &=& c_{00100},\\
 male &=& c_{00111},\\
 66 &=& c_{11000},
\end{array}
\right\}{\rm fillers}\\
\left.
\begin{array}{rcl}
 name &=& c_{00010},\\
 sex &=& c_{11100},\\
 age &=& c_{10001}.
\end{array}
\right\}{\rm roles}
\end{eqnarray}
The whole record then reads
\begin{eqnarray}
 PSmith
&=&
 name\ast Pat+sex\ast male+ age\ast 66\nonumber\\
&=&
c_{00010}c_{00100} + c_{11100}c_{00111} + c_{10001}c_{11000}\nonumber\\
&=&
- c_{00110} + c_{11011} + c_{01001}.
\end{eqnarray}
The decoding of $PSmith$'s $name$ will produce the following result
\begin{eqnarray}
 name^{+}\ast PSmith &=& c_{00100} + c_{11001} - c_{01011}\nonumber\\
&=&  Pat + \textrm{noise} =  Pat'.
\end{eqnarray}
At this point, inner products between $Pat'$ and the elements of the clean-up memory need to be compared. Item in the clean-up memory\index{clean-up memory} yielding the highest inner product will be the most likely candidate for $Pat$ 
\begin{eqnarray}
\langle  Pat | Pat'\rangle  &=& c_{00100} \cdot (c_{00100} + c_{11001} - c_{01011}) = 1 \neq 0, \\
\langle male | Pat' \rangle &=& 0,\\
\langle 66 | Pat'\rangle &=& 0,\\
\langle name | Pat'\rangle &=& 0,\\ 
\langle sex | Pat'\rangle &=& 0,\\
\langle age | Pat'\rangle &=& 0,\\
\langle PSmith | Pat'\rangle &=& 0.
\end{eqnarray}

A question arises as to how to extract information from a multivector --- should a question be asked on the left-hand-side of a multivector
\begin{equation}
 name\ast  PSmith,
\end{equation}
or the right-hand-side
\begin{equation}
 PSmith\ast  name.
\end{equation}
Furthermore, should we use $name$ or rather $name^{+}$? Since we can ask about both the role and the filler, we should be able to ask both right-hand-side and left-hand-side questions according to the principles of geometric algebra. Such an approach, however, would make the rules of decomposition unclear, which is against the philosophy of distributed representations. The problem of asking reversed questions on the appropriate side of a sentence is that we should be able to distinguish roles from fillers. This implies that atomic objects should be partly hand-generated, which is not a desirable property of a distributed representation model. If we decide that a question should always be asked on one fixed side of a sentence, there is no point in reversing the blade since there is no certainty that the fixed side is the appropriate one. Independently of the hand-sidedness of questions, in test results the moduli of inner products are compared instead of their actual (possibly negative) values. For right-hand-side questions we can reformulate Equations (\ref{eq:pat1}) -- (\ref{eq:pat3}) in the following way
\begin{eqnarray}
 PSmith\ \sharp\  name &=& PSmith \ast name \\
&=&\big[
c_{a_1\dots a_n}c_{x_1\dots x_n}+c_{b_1\dots b_n}c_{y_1\dots y_n}+\nonumber\\
& & c_{c_1\dots c_n}c_{z_1\dots z_n}\big] c_{a_1\dots a_n}\\
&=& \pm c_{x} \pm c_{b\oplus y\oplus a} \pm c_{c\oplus z\oplus a} \\
&=& \pm Pat + \textrm{noise} =  Pat'.
\end{eqnarray}
The decoding of $PSmith$'s $name$ will then take the form
\begin{equation}
PSmith \ast name = - c_{00100} - c_{11001} - c_{01011} = Pat',
\end{equation}
resulting in $|\langle PSmith | Pat'\rangle| = |-1| = 1$. We will study the effects of asking questions in various ways in the next Section.


\section{Recognition}
\label{sec:3}

Before we investigate the percentage of correctly recognized items, we need to introduce the following definitions. Let $S$ and $Q$ denote the $sentence$ and the $question$ respectively. Let $\mathcal A$ be the set of all clean-up memory items $A$ for which $\langle S\ \sharp\ Q | A\rangle\neq 0$. We will call $\mathcal A$ a set of $potential\ answers$. Let $m = \textrm{max}\{|\langle S\ \sharp\ Q | A\rangle| : A \in \mathcal A\}$ and $T = \{A\in \mathcal A : |\langle S\ \sharp\ Q | A\rangle| = m\}$. A $pseudo$-$answer$ is an answer belonging to set $T$ but different than the correct answer to $S\ \sharp\ Q$ --- even if the difference is only in the meaning and not in the multivector. Of course, set $T$ might also include the correct answer --- therefore, it is called the set of $(pseudo$-$)correct$ answers and is actually the set of answers leading to the highest modulus of the inner product. We assume that a noisy statement has been recognized correctly if its counterpart in the clean-up memory is among the (pseudo-)correct answers.

There are some doubts concerning how the sentences should be built --- Plate~\cite{Plate03} adds an additional vector denoting action id (usually a verb) to a sentence, e.g.
\begin{eqnarray}
(\underline{eat} + eat_{agt}\circledast Mark + eat_{obj} \circledast theFish)/ \sqrt{3},
\end{eqnarray}
where ``$\circledast$" denotes circular convolution\index{circular convolution}. We will distinguish between two types of sentence constructions
\begin{itemize}
\item Plate construction, e.g.\quad $eat + eat_{agt}\ast Mark + eat_{obj} \ast theFish$,
\item agent-object construction, e.g.\quad $eat_{agt}\ast Mark + eat_{obj} \ast theFish$.
\end{itemize}
The agent-object construction will often be denoted as ``A-O" for short, especially in table headings. Preliminary tests conducted on the GA model were designed to investigate which type of construction suits GA better. The vocabulary set and the sentence set for these tests are included in Table~\ref{tab:chmeasurestable1}. The sentence set is especially filled with similar sentences to test sensitivity of the GA model to confusing data. Each sentence carries a number (e.g. ``(3a)") to make further equations more compact and readable.

\begin{table}[t]
\caption{Contents of the clean-up memory used in tests described throughout this paper.}
\label{tab:chmeasurestable1}  
\begin{tabular}{|p{1.3cm}|p{10cm}|} \hline
& \\
Number of blades & Contents  \\ \hline
1 & A total of $42$ atomic objects: 19 fillers, 7 single-feature roles and 8 double-feature roles \\ \hline
2 & (1a) $\quad bite_{agt}\ast Fido + bite_{obj}\ast Pat$ \\ 
 & (2a) $\quad flee_{agt}\ast Pat+flee_{obj}\ast Fido$ \\ \hline
3 & (3a) $\quad see_{agt}\ast John+see_{obj}\ast$(1a) \\ 
 & ($PSmith$) $\quad name\ast Pat+sex\ast male+age\ast 66$ \\ \hline
 4 & (1b) $\quad bite_{agt}\ast Fido+bite_{obj}\ast PSmith$ \\ 
& (2c) $\quad flee_{agt}\ast PSmith+flee_{obj}\ast Fido$ \\ 
& (4a) $\quad cause_{agt}\ast$(1a)$+cause_{obj}\ast$(2a) \\ \hline
5 & (3b) $\quad see_{agt}\ast John+see_{obj}\ast$(1b) \\ 
& (5a) $\quad see_{agt}\ast John+see_{obj}\ast$(4a) \\ \hline
6 & (4c) $\quad cause_{agt}\ast$(1b)$+cause_{obj}\ast$(2a) \\ \hline
7 & ($DogFido$) $\quad class\ast animal+type\ast dog+taste\ast chickenlike$\\
& $\quad\quad\quad\quad\quad+name\ast Fido+age\ast 7+sex\ast male+occupation\ast pet$\\ \hline
8 & (1c) $\quad bite_{agt}\ast DogFido+bite_{obj}\ast Pat$ \\ 
& (2b) $\quad flee_{agt}\ast Pat+flee_{obj}\ast DogFido$  \\ 
& (4b) $\quad cause_{agt}\ast$(1b)$+cause_{obj}\ast$(2c) \\ \hline
9 & (3c) $\quad see_{agt}\ast John+see_{obj}\ast$(1c) \\ 
& (5b) $\quad see_{agt}\ast John+see_{obj}\ast$(4b) \\ \hline
10 & (1d) $\quad bite_{agt}\ast DogFido+bite_{obj}\ast PSmith$ \\ 
& (2d) $\quad flee_{agt}\ast PSmith +flee_{obj}\ast DogFido$ \\ \hline
11 & (3d) $\quad see_{agt}\ast John+see_{obj}\ast$(1d) \\ 
\hline
\end{tabular}
$^{(*)}$ Each sentence carries a number (e.g. ``(3a)") to make further equations more readable.
\end{table}

\subsection{Right-hand-side questions}
\label{sec:3:a}

In the previous Section we commented on the use of reversions and the choice of the side of a statement that a question should be asked on. The argument for right-hand-side (generally: fixed-hand-side) questions without a reversion was that rules of decomposition of a statement should be clear and unchangeable. However, the use of right-hand-side questions poses a problem best described by the following example. Let the clean-up memory contain seven roles and fillers
\begin{eqnarray}
\begin{matrix}
see_{agt} &=& c_{00101}, &\quad\quad\quad John &=& c_{00101},\cr
see_{obj} &=& c_{01010}, &\quad\quad\quad Pat &=& c_{10000},\cr
bite_{agt} &=& c_{10110}, &\quad\quad\quad Fido &=& c_{10001},\cr
bite_{obj} &=& c_{00001},
\end{matrix}
\end{eqnarray}
and two sentences mentioned in Table \ref{tab:chmeasurestable1}
\begin{eqnarray}
\textrm{(1a) }& bite_{agt}\ast Fido +bite_{obj}\ast Pat &= c_{00111} - c_{10001}, \\
\textrm{(3a) }& see_{agt}\ast John + see_{obj}\ast\textrm{(1a)} &= -c_{00000} - c_{01101} - c_{11011}.
\end{eqnarray}
Let us now ask a question (3a)$\ \sharp\ see_{obj} = \textrm{(3a)}\ast see_{obj}$. The decoded answer
\begin{eqnarray}
\textrm{(3a)}\ast see_{obj} &=& ( -c_{00000} - c_{01101} - c_{11011})c_{01010} \nonumber\\
&=& -c_{01010} + c_{00111} + c_{10001}\\
&=& \textrm{noise} + bite_{agt}\ast Fido - bite_{obj}\ast Pat
\end{eqnarray}
results in one noisy chunks and two chunks resembling sentence (1a) but having a partially different sign than the original (1a). Furthermore, 
\begin{eqnarray}
\textrm{(1a)}\cdot(\textrm{(3a)}\ast see_{obj}) &=& ( c_{00111} - c_{10001} )\cdot( -c_{01010} + c_{00111} + c_{10001} ) \nonumber\\
&=& c_{00111} \cdot c_{00111} - c_{10001}\cdot c_{10001}\\
&=& -\textbf{1} + \textbf{1} = 0.
\end{eqnarray}
Such a situation would not have happened if we asked differently
\begin{equation}
see_{obj}^{+}\ast\textrm{(3a)},
\end{equation}
since $see_{obj}^{+} see_{obj} = \textbf{1}$ for normalized atomic objects.

The similarity of (1a) and (3a)$\ast see_{obj}$ equals zero because the nonzero similarities of blades (i.e. \textbf{1}s) belonging to these statements cancelled each other out. Cancellation could be most likely avoided, if sentence (1a) had an odd number of blades. This observation has been backed up by test results comparing the performance of Plate construction and the agent-object construction.

\definecolor{mygray1}{rgb}{0.5,0.5,0.5}
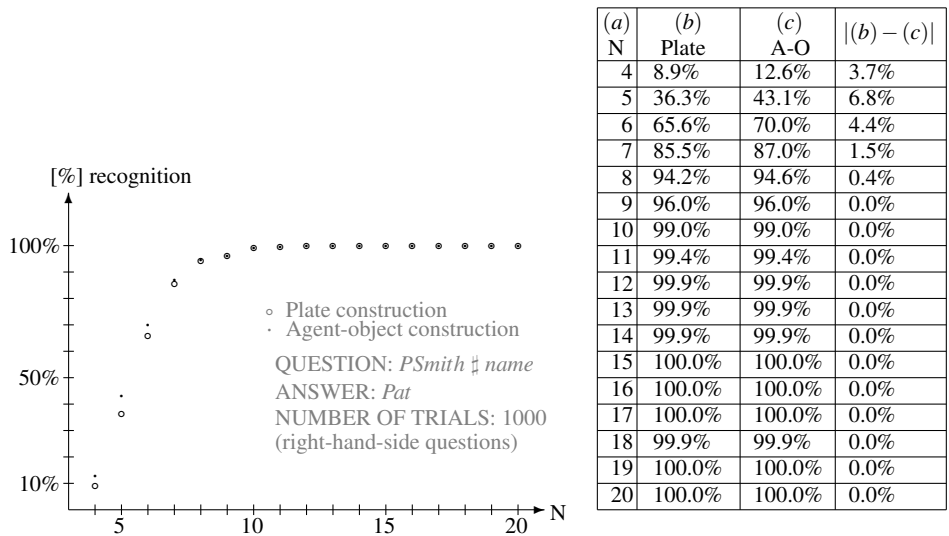
\begin{figure}[b]  
  \begin{picture}(0,195)(0,-10)
	 \put(0.0,0.0){\vector(1,0){180.0}}  
	 \put(0.0,0.0){\vector(0,1){120.0}} 
	 \put(10.0,2.0){\line(0,-1){4.0}} 
	 \put(20.0,2.0){\line(0,-1){4.0}} 
	 \put(30.0,2.0){\line(0,-1){4.0}} 
	 \put(40.0,2.0){\line(0,-1){4.0}} 
	 \put(50.0,2.0){\line(0,-1){4.0}} 
	 \put(60.0,2.0){\line(0,-1){4.0}} 
	 \put(70.0,2.0){\line(0,-1){4.0}} 
	 \put(80.0,2.0){\line(0,-1){4.0}} 
	 \put(90.0,2.0){\line(0,-1){4.0}} 
	 \put(100.0,2.0){\line(0,-1){4.0}} 
	 \put(110.0,2.0){\line(0,-1){4.0}} 
	 \put(120.0,2.0){\line(0,-1){4.0}} 
	 \put(130.0,2.0){\line(0,-1){4.0}} 
	 \put(140.0,2.0){\line(0,-1){4.0}} 
	 \put(150.0,2.0){\line(0,-1){4.0}} 
	 \put(160.0,2.0){\line(0,-1){4.0}} 
	 \put(170.0,2.0){\line(0,-1){4.0}} 
	 \put(-2.0,10.0){\line(1,0){4.0}} 
	 \put(-2.0,20.0){\line(1,0){4.0}} 
	 \put(-2.0,30.0){\line(1,0){4.0}} 
	 \put(-2.0,40.0){\line(1,0){4.0}} 
	 \put(-2.0,50.0){\line(1,0){4.0}} 
	 \put(-2.0,60.0){\line(1,0){4.0}} 
	 \put(-2.0,70.0){\line(1,0){4.0}} 
	 \put(-2.0,80.0){\line(1,0){4.0}} 
	 \put(-2.0,90.0){\line(1,0){4.0}} 
	 \put(-2.0,100.0){\line(1,0){4.0}} 
	 \put(15.0,-9.0){\makebox(0,0)[bl]{ 5}}
	 \put(63.0,-9.0){\makebox(0,0)[bl]{ 10}}
	 \put(113.0,-9.0){\makebox(0,0)[bl]{ 15}}
	 \put(163.0,-9.0){\makebox(0,0)[bl]{ 20}}
	 \put(180.0,-4.0){\makebox(0,0)[bl]{ N}}
	 \put(-21,7.0){\makebox(0,0)[bl]{ 10\%}}
	 \put(-21,47.0){\makebox(0,0)[bl]{ 50\%}}
	 \put(-25,97.0){\makebox(0,0)[bl]{ 100\%}}
	 \put(-7.0,122.0){\makebox(0,0)[bl]{[\%] recognition}}
	 \put(10.0,8.9){\circle{1.5}}
	 \put(20.0,36.3){\circle{1.5}}
	 \put(30.0,65.6){\circle{1.5}}
	 \put(40.0,85.5){\circle{1.5}}
	 \put(50.0,94.2){\circle{1.5}}
	 \put(60.0,96.0){\circle{1.5}}
	 \put(70.0,99.0){\circle{1.5}}
	 \put(80.0,99.4){\circle{1.5}}
	 \put(90.0,99.9){\circle{1.5}}
	 \put(100.0,99.9){\circle{1.5}}
	 \put(110.0,99.9){\circle{1.5}}
	 \put(120.0,100.0){\circle{1.5}}
	 \put(130.0,100.0){\circle{1.5}}
	 \put(140.0,100.0){\circle{1.5}}
	 \put(150.0,99.9){\circle{1.5}}
	 \put(160.0,100.0){\circle{1.5}}
	 \put(170.0,100.0){\circle{1.5}}
	 \put(10.0,12.6){\circle*{1}}
	 \put(20.0,43.1){\circle*{1}}
	 \put(30.0,70.0){\circle*{1}}
	 \put(40.0,87.0){\circle*{1}}
	 \put(50.0,94.6){\circle*{1}}
	 \put(60.0,96.0){\circle*{1}}
	 \put(70.0,99.0){\circle*{1}}
	 \put(80.0,99.4){\circle*{1}}
	 \put(90.0,99.9){\circle*{1}}
	 \put(100.0,99.9){\circle*{1}}
	 \put(110.0,99.9){\circle*{1}}
	 \put(120.0,100.0){\circle*{1}}
	 \put(130.0,100.0){\circle*{1}}
	 \put(140.0,100.0){\circle*{1}}
	 \put(150.0,99.9){\circle*{1}}
	 \put(160.0,100.0){\circle*{1}}
	 \put(170.0,100.0){\circle*{1}}
	 \color{mygray1}
	 \put(76,75){\circle{1.5}}
	 \put(80,73){\makebox(0,0)[bl]{ Plate construction}}
	 \put(76,68){\circle*{1}}
	 \put(80,64){\makebox(0,0)[bl]{ Agent-object construction}}
	 \put(76.0,51.0){\makebox(0,0)[bl]{ QUESTION: $PSmith\ \sharp\ name$}}
	 \put(76.0,42.0){\makebox(0,0)[bl]{ ANSWER: $Pat$}}
	 \put(76.0,32.0){\makebox(0,0)[bl]{ NUMBER OF TRIALS: 1000}}
	 \put(76.0,21.0){\makebox(0,0)[bl]{ (right-hand-side questions)}}	 	 
	 \color{black}
	 \put(200.0,190.0){\line(1,0){132.0}}  
	 \put(200.0,170.0){\line(1,0){132.0}}  
	 \put(200.0,160.0){\line(1,0){132.0}}  
	 \put(200.0,150.0){\line(1,0){132.0}}  
	 \put(200.0,140.0){\line(1,0){132.0}}  
	 \put(200.0,130.0){\line(1,0){132.0}}  
	 \put(200.0,120.0){\line(1,0){132.0}}  
	 \put(200.0,110.0){\line(1,0){132.0}}  
	 \put(200.0,100.0){\line(1,0){132.0}}  
	 \put(200.0,90.0){\line(1,0){132.0}}  
	 \put(200.0,80.0){\line(1,0){132.0}}  
	 \put(200.0,70.0){\line(1,0){132.0}}  
	 \put(200.0,60.0){\line(1,0){132.0}}  
	 \put(200.0,50.0){\line(1,0){132.0}}  
	 \put(200.0,40.0){\line(1,0){132.0}}  
	 \put(200.0,30.0){\line(1,0){132.0}}  
	 \put(200.0,20.0){\line(1,0){132.0}}  
	 \put(200.0,10.0){\line(1,0){132.0}}  
	 \put(200.0,0.0){\line(1,0){132.0}}  
	 \put(200.0,0.0){\line(0,1){190.0}}  
	 \put(215.0,0.0){\line(0,1){190.0}}  
	 \put(254.0,0.0){\line(0,1){190.0}}  
	 \put(290.0,0.0){\line(0,1){190.0}}  
	 \put(332.0,0.0){\line(0,1){190.0}}  
	 \put(203,3.0){\makebox(0,0)[bl]{ 20}}
	 \put(203,13.0){\makebox(0,0)[bl]{ 19}}
	 \put(203,23.0){\makebox(0,0)[bl]{ 18}}
	 \put(203,33.0){\makebox(0,0)[bl]{ 17}}
	 \put(203,43.0){\makebox(0,0)[bl]{ 16}}
	 \put(203,53.0){\makebox(0,0)[bl]{ 15}}
	 \put(203,63.0){\makebox(0,0)[bl]{ 14}}
	 \put(203,73.0){\makebox(0,0)[bl]{ 13}}
	 \put(203,83.0){\makebox(0,0)[bl]{ 12}}
	 \put(203,93.0){\makebox(0,0)[bl]{ 11}}
	 \put(203,103.0){\makebox(0,0)[bl]{ 10}}
	 \put(207,113.0){\makebox(0,0)[bl]{ 9}}
	 \put(207,123.0){\makebox(0,0)[bl]{ 8}}
	 \put(207,133.0){\makebox(0,0)[bl]{ 7}}
	 \put(207,143.0){\makebox(0,0)[bl]{ 6}}
	 \put(207,153.0){\makebox(0,0)[bl]{ 5}}
	 \put(207,163.0){\makebox(0,0)[bl]{ 4}}
	 \put(202,172.0){\makebox(0,0)[bl]{ N}}
	 \put(202,180.0){\makebox(0,0)[bl]{$(a)$}}
	 \put(224,172.0){\makebox(0,0)[bl]{{Plate}}}
	 \put(229,180.0){\makebox(0,0)[bl]{$(b)$}}
	 \put(219,3.0){\makebox(0,0)[bl]{ 100.0\%}} 
	 \put(219,13.0){\makebox(0,0)[bl]{ 100.0\%}}
	 \put(219,23.0){\makebox(0,0)[bl]{ 99.9\%}}
	 \put(219,33.0){\makebox(0,0)[bl]{ 100.0\%}}
	 \put(219,43.0){\makebox(0,0)[bl]{ 100.0\%}}
	 \put(219,53.0){\makebox(0,0)[bl]{ 100.0\%}}
	 \put(219,63.0){\makebox(0,0)[bl]{ 99.9\%}}
	 \put(219,73.0){\makebox(0,0)[bl]{ 99.9\%}}
	 \put(219,83.0){\makebox(0,0)[bl]{ 99.9\%}}
	 \put(219,93.0){\makebox(0,0)[bl]{ 99.4\%}}
	 \put(219,103.0){\makebox(0,0)[bl]{ 99.0\%}}
	 \put(219,113.0){\makebox(0,0)[bl]{ 96.0\%}}
	 \put(219,123.0){\makebox(0,0)[bl]{ 94.2\%}}
	 \put(219,133.0){\makebox(0,0)[bl]{ 85.5\%}}
	 \put(219,143.0){\makebox(0,0)[bl]{ 65.6\%}}
	 \put(219,153.0){\makebox(0,0)[bl]{ 36.3\%}}
	 \put(219,163.0){\makebox(0,0)[bl]{ 8.9\%}}
	 \put(265.5,172.0){\makebox(0,0)[bl]{{A-O}}}
	 \put(267.5,180.0){\makebox(0,0)[bl]{$(c)$}}
	 \put(257,3.0){\makebox(0,0)[bl]{ 100.0\%}} 
	 \put(257,13.0){\makebox(0,0)[bl]{ 100.0\%}}
	 \put(257,23.0){\makebox(0,0)[bl]{ 99.9\%}}
	 \put(257,33.0){\makebox(0,0)[bl]{ 100.0\%}}
	 \put(257,43.0){\makebox(0,0)[bl]{ 100.0\%}}
	 \put(257,53.0){\makebox(0,0)[bl]{ 100.0\%}}
	 \put(257,63.0){\makebox(0,0)[bl]{ 99.9\%}}
	 \put(257,73.0){\makebox(0,0)[bl]{ 99.9\%}}
	 \put(257,83.0){\makebox(0,0)[bl]{ 99.9\%}}
	 \put(257,93.0){\makebox(0,0)[bl]{ 99.4\%}}
	 \put(257,103.0){\makebox(0,0)[bl]{ 99.0\%}}
	 \put(257,113.0){\makebox(0,0)[bl]{ 96.0\%}}
	 \put(257,123.0){\makebox(0,0)[bl]{ 94.6\%}}
	 \put(257,133.0){\makebox(0,0)[bl]{ 87.0\%}}
	 \put(257,143.0){\makebox(0,0)[bl]{ 70.0\%}}
	 \put(257,153.0){\makebox(0,0)[bl]{ 43.1\%}}
	 \put(257,163.0){\makebox(0,0)[bl]{ 12.6\%}}
	 \put(293,176.0){\makebox(0,0)[bl]{$|(b)-(c)|$}}
	 \put(293,3.0){\makebox(0,0)[bl]{ 0.0\%}} 
	 \put(293,13.0){\makebox(0,0)[bl]{ 0.0\%}}
	 \put(293,23.0){\makebox(0,0)[bl]{ 0.0\%}}
	 \put(293,33.0){\makebox(0,0)[bl]{ 0.0\%}}
	 \put(293,43.0){\makebox(0,0)[bl]{ 0.0\%}}
	 \put(293,53.0){\makebox(0,0)[bl]{ 0.0\%}}
	 \put(293,63.0){\makebox(0,0)[bl]{ 0.0\%}}
	 \put(293,73.0){\makebox(0,0)[bl]{ 0.0\%}}
	 \put(293,83.0){\makebox(0,0)[bl]{ 0.0\%}}
	 \put(293,93.0){\makebox(0,0)[bl]{ 0.0\%}}
	 \put(293,103.0){\makebox(0,0)[bl]{ 0.0\%}}
	 \put(293,113.0){\makebox(0,0)[bl]{ 0.0\%}}	 
	 \put(293,123.0){\makebox(0,0)[bl]{ 0.4\%}}	 
	 \put(293,133.0){\makebox(0,0)[bl]{ 1.5\%}}	 
	 \put(293,143.0){\makebox(0,0)[bl]{ 4.4\%}}	 
	 \put(293,153.0){\makebox(0,0)[bl]{ 6.8\%}}	 
	 \put(293,163.0){\makebox(0,0)[bl]{ 3.7\%}}	 
  \end{picture}
  \caption{Recognition test results for $PSmith\ \sharp\ name$.}
  \label{fig:4a_PSmithName} 
\end{figure}

\definecolor{mygray1}{rgb}{0.5,0.5,0.5}
\begin{figure}[ht]  
  \begin{picture}(0,195)(0,-10)
	 \put(0.0,0.0){\vector(1,0){180.0}}  
	 \put(0.0,0.0){\vector(0,1){120.0}} 
	 \put(10.0,2.0){\line(0,-1){4.0}} 
	 \put(20.0,2.0){\line(0,-1){4.0}} 
	 \put(30.0,2.0){\line(0,-1){4.0}} 
	 \put(40.0,2.0){\line(0,-1){4.0}} 
	 \put(50.0,2.0){\line(0,-1){4.0}} 
	 \put(60.0,2.0){\line(0,-1){4.0}} 
	 \put(70.0,2.0){\line(0,-1){4.0}} 
	 \put(80.0,2.0){\line(0,-1){4.0}} 
	 \put(90.0,2.0){\line(0,-1){4.0}} 
	 \put(100.0,2.0){\line(0,-1){4.0}} 
	 \put(110.0,2.0){\line(0,-1){4.0}} 
	 \put(120.0,2.0){\line(0,-1){4.0}} 
	 \put(130.0,2.0){\line(0,-1){4.0}} 
	 \put(140.0,2.0){\line(0,-1){4.0}} 
	 \put(150.0,2.0){\line(0,-1){4.0}} 
	 \put(160.0,2.0){\line(0,-1){4.0}} 
	 \put(170.0,2.0){\line(0,-1){4.0}} 
	 \put(-2.0,10.0){\line(1,0){4.0}} 
	 \put(-2.0,20.0){\line(1,0){4.0}} 
	 \put(-2.0,30.0){\line(1,0){4.0}} 
	 \put(-2.0,40.0){\line(1,0){4.0}} 
	 \put(-2.0,50.0){\line(1,0){4.0}} 
	 \put(-2.0,60.0){\line(1,0){4.0}} 
	 \put(-2.0,70.0){\line(1,0){4.0}} 
	 \put(-2.0,80.0){\line(1,0){4.0}} 
	 \put(-2.0,90.0){\line(1,0){4.0}} 
	 \put(-2.0,100.0){\line(1,0){4.0}} 
	 \put(15.0,-9.0){\makebox(0,0)[bl]{ 5}}
	 \put(63.0,-9.0){\makebox(0,0)[bl]{ 10}}
	 \put(113.0,-9.0){\makebox(0,0)[bl]{ 15}}
	 \put(163.0,-9.0){\makebox(0,0)[bl]{ 20}}
	 \put(180.0,-4.0){\makebox(0,0)[bl]{ N}}
	 \put(-21,7.0){\makebox(0,0)[bl]{ 10\%}}
	 \put(-21,47.0){\makebox(0,0)[bl]{ 50\%}}
	 \put(-25,97.0){\makebox(0,0)[bl]{ 100\%}}
	 \put(-7.0,122.0){\makebox(0,0)[bl]{[\%] recognition}}
	 \put(10.0,0.6){\circle{1.5}}
	 \put(20.0,3.3){\circle{1.5}}
	 \put(30.0,15.3){\circle{1.5}}
	 \put(40.0,47.8){\circle{1.5}}
	 \put(50.0,72.2){\circle{1.5}}
	 \put(60.0,86.4){\circle{1.5}}
	 \put(70.0,95.1){\circle{1.5}}
	 \put(80.0,97.7){\circle{1.5}}
	 \put(90.0,98.1){\circle{1.5}}
	 \put(100.0,99.1){\circle{1.5}}
	 \put(110.0,99.9){\circle{1.5}}
	 \put(120.0,99.6){\circle{1.5}}
	 \put(130.0,100.0){\circle{1.5}}
	 \put(140.0,100.0){\circle{1.5}}
	 \put(150.0,100.0){\circle{1.5}}
	 \put(160.0,100.0){\circle{1.5}}
	 \put(170.0,100.0){\circle{1.5}}
	 \put(10.0,4.4){\circle*{1}}
	 \put(20.0,18.5){\circle*{1}}
	 \put(30.0,45.8){\circle*{1}}
	 \put(40.0,76.4){\circle*{1}}
	 \put(50.0,88.9){\circle*{1}}
	 \put(60.0,94.3){\circle*{1}}
	 \put(70.0,98.2){\circle*{1}}
	 \put(80.0,98.9){\circle*{1}}
	 \put(90.0,99.2){\circle*{1}}
	 \put(100.0,99.6){\circle*{1}}
	 \put(110.0,99.9){\circle*{1}}
	 \put(120.0,99.9){\circle*{1}}
	 \put(130.0,100.0){\circle*{1}}
	 \put(140.0,100.0){\circle*{1}}
	 \put(150.0,100.0){\circle*{1}}
	 \put(160.0,100.0){\circle*{1}}
	 \put(170.0,100.0){\circle*{1}}
	 \color{mygray1}
	 \put(76,75){\circle{1.5}}
	 \put(80,73){\makebox(0,0)[bl]{ Plate construction}}
	 \put(76,68){\circle*{1}}
	 \put(80,64){\makebox(0,0)[bl]{ Agent-object construction}}
	 \put(76.0,51.0){\makebox(0,0)[bl]{ QUESTION: (5a)$\ \sharp\ see_{agt}$}}
	 \put(76.0,43.0){\makebox(0,0)[bl]{ ANSWER: $John$}}
	 \put(76.0,33.0){\makebox(0,0)[bl]{ NUMBER OF TRIALS: 1000}}
	 \put(76.0,22.0){\makebox(0,0)[bl]{ (right-hand-side questions)}}	 	 
	 \color{black}
	 \put(200.0,190.0){\line(1,0){132.0}}  
	 \put(200.0,170.0){\line(1,0){132.0}}  
	 \put(200.0,160.0){\line(1,0){132.0}}  
	 \put(200.0,150.0){\line(1,0){132.0}}  
	 \put(200.0,140.0){\line(1,0){132.0}}  
	 \put(200.0,130.0){\line(1,0){132.0}}  
	 \put(200.0,120.0){\line(1,0){132.0}}  
	 \put(200.0,110.0){\line(1,0){132.0}}  
	 \put(200.0,100.0){\line(1,0){132.0}}  
	 \put(200.0,90.0){\line(1,0){132.0}}  
	 \put(200.0,80.0){\line(1,0){132.0}}  
	 \put(200.0,70.0){\line(1,0){132.0}}  
	 \put(200.0,60.0){\line(1,0){132.0}}  
	 \put(200.0,50.0){\line(1,0){132.0}}  
	 \put(200.0,40.0){\line(1,0){132.0}}  
	 \put(200.0,30.0){\line(1,0){132.0}}  
	 \put(200.0,20.0){\line(1,0){132.0}}  
	 \put(200.0,10.0){\line(1,0){132.0}}  
	 \put(200.0,0.0){\line(1,0){132.0}}  
	 \put(200.0,0.0){\line(0,1){190.0}}  
	 \put(215.0,0.0){\line(0,1){190.0}}  
	 \put(254.0,0.0){\line(0,1){190.0}}  
	 \put(290.0,0.0){\line(0,1){190.0}}  
	 \put(332.0,0.0){\line(0,1){190.0}}  
	 \put(203,3.0){\makebox(0,0)[bl]{ 20}}
	 \put(203,13.0){\makebox(0,0)[bl]{ 19}}
	 \put(203,23.0){\makebox(0,0)[bl]{ 18}}
	 \put(203,33.0){\makebox(0,0)[bl]{ 17}}
	 \put(203,43.0){\makebox(0,0)[bl]{ 16}}
	 \put(203,53.0){\makebox(0,0)[bl]{ 15}}
	 \put(203,63.0){\makebox(0,0)[bl]{ 14}}
	 \put(203,73.0){\makebox(0,0)[bl]{ 13}}
	 \put(203,83.0){\makebox(0,0)[bl]{ 12}}
	 \put(203,93.0){\makebox(0,0)[bl]{ 11}}
	 \put(203,103.0){\makebox(0,0)[bl]{ 10}}
	 \put(207,113.0){\makebox(0,0)[bl]{ 9}}
	 \put(207,123.0){\makebox(0,0)[bl]{ 8}}
	 \put(207,133.0){\makebox(0,0)[bl]{ 7}}
	 \put(207,143.0){\makebox(0,0)[bl]{ 6}}
	 \put(207,153.0){\makebox(0,0)[bl]{ 5}}
	 \put(207,163.0){\makebox(0,0)[bl]{ 4}}
	 \put(202,172.0){\makebox(0,0)[bl]{ N}}
	 \put(202,180.0){\makebox(0,0)[bl]{$(a)$}}
	 \put(224,172.0){\makebox(0,0)[bl]{{Plate}}}
	 \put(229,180.0){\makebox(0,0)[bl]{$(b)$}}
	 \put(219,3.0){\makebox(0,0)[bl]{ 100.0\%}} 
	 \put(219,13.0){\makebox(0,0)[bl]{ 100.0\%}}
	 \put(219,23.0){\makebox(0,0)[bl]{ 100.0\%}}
	 \put(219,33.0){\makebox(0,0)[bl]{ 100.0\%}}
	 \put(219,43.0){\makebox(0,0)[bl]{ 100.0\%}}
	 \put(219,53.0){\makebox(0,0)[bl]{ 99.6\%}}
	 \put(219,63.0){\makebox(0,0)[bl]{ 99.9\%}}
	 \put(219,73.0){\makebox(0,0)[bl]{ 99.1\%}}
	 \put(219,83.0){\makebox(0,0)[bl]{ 98.1\%}}
	 \put(219,93.0){\makebox(0,0)[bl]{ 97.7\%}}
	 \put(219,103.0){\makebox(0,0)[bl]{ 95.1\%}}
	 \put(219,113.0){\makebox(0,0)[bl]{ 86.4\%}}
	 \put(219,123.0){\makebox(0,0)[bl]{ 72.2\%}}
	 \put(219,133.0){\makebox(0,0)[bl]{ 47.8\%}}
	 \put(219,143.0){\makebox(0,0)[bl]{ 15.3\%}}
	 \put(219,153.0){\makebox(0,0)[bl]{ 3.3\%}}
	 \put(219,163.0){\makebox(0,0)[bl]{ 0.6\%}}
	 \put(265.5,172.0){\makebox(0,0)[bl]{{A-O}}}
	 \put(267.5,180.0){\makebox(0,0)[bl]{$(c)$}}
	 \put(257,3.0){\makebox(0,0)[bl]{ 100.0\%}} 
	 \put(257,13.0){\makebox(0,0)[bl]{ 100.0\%}}
	 \put(257,23.0){\makebox(0,0)[bl]{ 100.0\%}}
	 \put(257,33.0){\makebox(0,0)[bl]{ 100.0\%}}
	 \put(257,43.0){\makebox(0,0)[bl]{ 100.0\%}}
	 \put(257,53.0){\makebox(0,0)[bl]{ 99.9\%}}
	 \put(257,63.0){\makebox(0,0)[bl]{ 99.9\%}}
	 \put(257,73.0){\makebox(0,0)[bl]{ 99.6\%}}
	 \put(257,83.0){\makebox(0,0)[bl]{ 99.2\%}}
	 \put(257,93.0){\makebox(0,0)[bl]{ 98.9\%}}
	 \put(257,103.0){\makebox(0,0)[bl]{ 98.2\%}}
	 \put(257,113.0){\makebox(0,0)[bl]{ 94.3\%}}
	 \put(257,123.0){\makebox(0,0)[bl]{ 88.9\%}}
	 \put(257,133.0){\makebox(0,0)[bl]{ 76.4\%}}
	 \put(257,143.0){\makebox(0,0)[bl]{ 45.8\%}}
	 \put(257,153.0){\makebox(0,0)[bl]{ 18.5\%}}
	 \put(257,163.0){\makebox(0,0)[bl]{ 4.4\%}}
	 \put(293,176.0){\makebox(0,0)[bl]{$|(b)-(c)|$}}
	 \put(293,3.0){\makebox(0,0)[bl]{ 0.0\%}} 
	 \put(293,13.0){\makebox(0,0)[bl]{ 0.0\%}}
	 \put(293,23.0){\makebox(0,0)[bl]{ 0.0\%}}
	 \put(293,33.0){\makebox(0,0)[bl]{ 0.0\%}}
	 \put(293,43.0){\makebox(0,0)[bl]{ 0.0\%}}
	 \put(293,53.0){\makebox(0,0)[bl]{ 0.3\%}}
	 \put(293,63.0){\makebox(0,0)[bl]{ 0.0\%}}
	 \put(293,73.0){\makebox(0,0)[bl]{ 0.5\%}}
	 \put(293,83.0){\makebox(0,0)[bl]{ 1.1\%}}
	 \put(293,93.0){\makebox(0,0)[bl]{ 1.2\%}}
	 \put(293,103.0){\makebox(0,0)[bl]{ 3.1\%}}
	 \put(293,113.0){\makebox(0,0)[bl]{ 7.9\%}}	 
	 \put(293,123.0){\makebox(0,0)[bl]{ 16.7\%}}	 
	 \put(293,133.0){\makebox(0,0)[bl]{ 28.6\%}}	 
	 \put(293,143.0){\makebox(0,0)[bl]{ 30.5\%}}	 
	 \put(293,153.0){\makebox(0,0)[bl]{ 15.2\%}}	 
	 \put(293,163.0){\makebox(0,0)[bl]{ 3.8\%}}	 
  \end{picture}
  \caption{Recognition test results for (5a)$\ \sharp\ see_{agt}$.}
  \label{fig:4b_5aSeeagt} 
\end{figure}
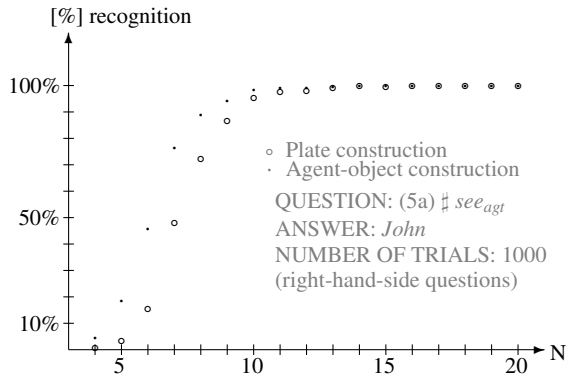

\definecolor{mygray1}{rgb}{0.5,0.5,0.5}
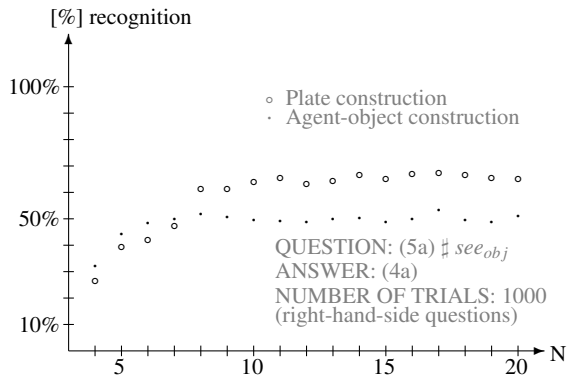
\begin{figure}[ht]  
  \begin{picture}(0,195)(0,-10)
	 \put(0.0,0.0){\vector(1,0){180.0}}  
	 \put(0.0,0.0){\vector(0,1){120.0}} 
	 \put(10.0,2.0){\line(0,-1){4.0}} 
	 \put(20.0,2.0){\line(0,-1){4.0}} 
	 \put(30.0,2.0){\line(0,-1){4.0}} 
	 \put(40.0,2.0){\line(0,-1){4.0}} 
	 \put(50.0,2.0){\line(0,-1){4.0}} 
	 \put(60.0,2.0){\line(0,-1){4.0}} 
	 \put(70.0,2.0){\line(0,-1){4.0}} 
	 \put(80.0,2.0){\line(0,-1){4.0}} 
	 \put(90.0,2.0){\line(0,-1){4.0}} 
	 \put(100.0,2.0){\line(0,-1){4.0}} 
	 \put(110.0,2.0){\line(0,-1){4.0}} 
	 \put(120.0,2.0){\line(0,-1){4.0}} 
	 \put(130.0,2.0){\line(0,-1){4.0}} 
	 \put(140.0,2.0){\line(0,-1){4.0}} 
	 \put(150.0,2.0){\line(0,-1){4.0}} 
	 \put(160.0,2.0){\line(0,-1){4.0}} 
	 \put(170.0,2.0){\line(0,-1){4.0}} 
	 \put(-2.0,10.0){\line(1,0){4.0}} 
	 \put(-2.0,20.0){\line(1,0){4.0}} 
	 \put(-2.0,30.0){\line(1,0){4.0}} 
	 \put(-2.0,40.0){\line(1,0){4.0}} 
	 \put(-2.0,50.0){\line(1,0){4.0}} 
	 \put(-2.0,60.0){\line(1,0){4.0}} 
	 \put(-2.0,70.0){\line(1,0){4.0}} 
	 \put(-2.0,80.0){\line(1,0){4.0}} 
	 \put(-2.0,90.0){\line(1,0){4.0}} 
	 \put(-2.0,100.0){\line(1,0){4.0}} 
	 \put(15.0,-9.0){\makebox(0,0)[bl]{ 5}}
	 \put(63.0,-9.0){\makebox(0,0)[bl]{ 10}}
	 \put(113.0,-9.0){\makebox(0,0)[bl]{ 15}}
	 \put(163.0,-9.0){\makebox(0,0)[bl]{ 20}}
	 \put(180.0,-4.0){\makebox(0,0)[bl]{ N}}
	 \put(-21,7.0){\makebox(0,0)[bl]{ 10\%}}
	 \put(-21,47.0){\makebox(0,0)[bl]{ 50\%}}
	 \put(-25,97.0){\makebox(0,0)[bl]{ 100\%}}
	 \put(-7.0,122.0){\makebox(0,0)[bl]{[\%] recognition}}
	 \put(10.0,26.4){\circle{1.5}}
	 \put(20.0,39.2){\circle{1.5}}
	 \put(30.0,41.8){\circle{1.5}}
	 \put(40.0,47.3){\circle{1.5}}
	 \put(50.0,61.4){\circle{1.5}}
	 \put(60.0,61.2){\circle{1.5}}
	 \put(70.0,64.0){\circle{1.5}}
	 \put(80.0,65.4){\circle{1.5}}
	 \put(90.0,63.2){\circle{1.5}}
	 \put(100.0,64.3){\circle{1.5}}
	 \put(110.0,66.4){\circle{1.5}}
	 \put(120.0,64.9){\circle{1.5}}
	 \put(130.0,66.9){\circle{1.5}}
	 \put(140.0,67.3){\circle{1.5}}
	 \put(150.0,66.4){\circle{1.5}}
	 \put(160.0,65.4){\circle{1.5}}
	 \put(170.0,65.2){\circle{1.5}}
	 \put(10.0,32.2){\circle*{1}}
	 \put(20.0,44.2){\circle*{1}}
	 \put(30.0,48.6){\circle*{1}}
	 \put(40.0,50.1){\circle*{1}}
	 \put(50.0,51.7){\circle*{1}}
	 \put(60.0,50.6){\circle*{1}}
	 \put(70.0,49.5){\circle*{1}}
	 \put(80.0,49.3){\circle*{1}}
	 \put(90.0,48.7){\circle*{1}}
	 \put(100.0,49.9){\circle*{1}}
	 \put(110.0,50.4){\circle*{1}}
	 \put(120.0,48.8){\circle*{1}}
	 \put(130.0,49.8){\circle*{1}}
	 \put(140.0,53.3){\circle*{1}}
	 \put(150.0,49.4){\circle*{1}}
	 \put(160.0,48.8){\circle*{1}}
	 \put(170.0,51.1){\circle*{1}}
	 \color{mygray1}
	 \put(76,95){\circle{1.5}}
	 \put(80,93){\makebox(0,0)[bl]{ Plate construction}}
	 \put(76,88){\circle*{1}}
	 \put(80,84){\makebox(0,0)[bl]{ Agent-object construction}}
	 \put(76.0,34.0){\makebox(0,0)[bl]{ QUESTION: (5a)$\ \sharp\ see_{obj}$}}
	 \put(76.0,26.0){\makebox(0,0)[bl]{ ANSWER: (4a)}}
	 \put(76.0,18.0){\makebox(0,0)[bl]{ NUMBER OF TRIALS: 1000}}
	 \put(76.0,9.0){\makebox(0,0)[bl]{ (right-hand-side questions)}}	 	 
	 \color{black}
	 \put(200.0,190.0){\line(1,0){132.0}}  
	 \put(200.0,170.0){\line(1,0){132.0}}  
	 \put(200.0,160.0){\line(1,0){132.0}}  
	 \put(200.0,150.0){\line(1,0){132.0}}  
	 \put(200.0,140.0){\line(1,0){132.0}}  
	 \put(200.0,130.0){\line(1,0){132.0}}  
	 \put(200.0,120.0){\line(1,0){132.0}}  
	 \put(200.0,110.0){\line(1,0){132.0}}  
	 \put(200.0,100.0){\line(1,0){132.0}}  
	 \put(200.0,90.0){\line(1,0){132.0}}  
	 \put(200.0,80.0){\line(1,0){132.0}}  
	 \put(200.0,70.0){\line(1,0){132.0}}  
	 \put(200.0,60.0){\line(1,0){132.0}}  
	 \put(200.0,50.0){\line(1,0){132.0}}  
	 \put(200.0,40.0){\line(1,0){132.0}}  
	 \put(200.0,30.0){\line(1,0){132.0}}  
	 \put(200.0,20.0){\line(1,0){132.0}}  
	 \put(200.0,10.0){\line(1,0){132.0}}  
	 \put(200.0,0.0){\line(1,0){132.0}}  
	 \put(200.0,0.0){\line(0,1){190.0}}  
	 \put(215.0,0.0){\line(0,1){190.0}}  
	 \put(254.0,0.0){\line(0,1){190.0}}  
	 \put(290.0,0.0){\line(0,1){190.0}}  
	 \put(332.0,0.0){\line(0,1){190.0}}  
	 \put(203,3.0){\makebox(0,0)[bl]{ 20}}
	 \put(203,13.0){\makebox(0,0)[bl]{ 19}}
	 \put(203,23.0){\makebox(0,0)[bl]{ 18}}
	 \put(203,33.0){\makebox(0,0)[bl]{ 17}}
	 \put(203,43.0){\makebox(0,0)[bl]{ 16}}
	 \put(203,53.0){\makebox(0,0)[bl]{ 15}}
	 \put(203,63.0){\makebox(0,0)[bl]{ 14}}
	 \put(203,73.0){\makebox(0,0)[bl]{ 13}}
	 \put(203,83.0){\makebox(0,0)[bl]{ 12}}
	 \put(203,93.0){\makebox(0,0)[bl]{ 11}}
	 \put(203,103.0){\makebox(0,0)[bl]{ 10}}
	 \put(207,113.0){\makebox(0,0)[bl]{ 9}}
	 \put(207,123.0){\makebox(0,0)[bl]{ 8}}
	 \put(207,133.0){\makebox(0,0)[bl]{ 7}}
	 \put(207,143.0){\makebox(0,0)[bl]{ 6}}
	 \put(207,153.0){\makebox(0,0)[bl]{ 5}}
	 \put(207,163.0){\makebox(0,0)[bl]{ 4}}
	 \put(202,172.0){\makebox(0,0)[bl]{ N}}
	 \put(202,180.0){\makebox(0,0)[bl]{$(a)$}}
	 \put(224,172.0){\makebox(0,0)[bl]{{Plate}}}
	 \put(229,180.0){\makebox(0,0)[bl]{$(b)$}}
	 \put(219,3.0){\makebox(0,0)[bl]{ 65.2\%}} 
	 \put(219,13.0){\makebox(0,0)[bl]{ 65.4\%}}
	 \put(219,23.0){\makebox(0,0)[bl]{ 66.4\%}}
	 \put(219,33.0){\makebox(0,0)[bl]{ 67.3\%}}
	 \put(219,43.0){\makebox(0,0)[bl]{ 66.9\%}}
	 \put(219,53.0){\makebox(0,0)[bl]{ 64.9\%}}
	 \put(219,63.0){\makebox(0,0)[bl]{ 66.4\%}}
	 \put(219,73.0){\makebox(0,0)[bl]{ 64.3\%}}
	 \put(219,83.0){\makebox(0,0)[bl]{ 63.2\%}}
	 \put(219,93.0){\makebox(0,0)[bl]{ 65.4\%}}
	 \put(219,103.0){\makebox(0,0)[bl]{ 64.0\%}}
	 \put(219,113.0){\makebox(0,0)[bl]{ 61.2\%}}
	 \put(219,123.0){\makebox(0,0)[bl]{ 61.4\%}}
	 \put(219,133.0){\makebox(0,0)[bl]{ 47.3\%}}
	 \put(219,143.0){\makebox(0,0)[bl]{ 41.8\%}}
	 \put(219,153.0){\makebox(0,0)[bl]{ 39.2\%}}
	 \put(219,163.0){\makebox(0,0)[bl]{ 26.4\%}}
	 \put(265.5,172.0){\makebox(0,0)[bl]{{A-O}}}
	 \put(267.5,180.0){\makebox(0,0)[bl]{$(c)$}}
	 \put(257,3.0){\makebox(0,0)[bl]{ 51.1\%}} 
	 \put(257,13.0){\makebox(0,0)[bl]{ 48.8\%}}
	 \put(257,23.0){\makebox(0,0)[bl]{ 49.4\%}}
	 \put(257,33.0){\makebox(0,0)[bl]{ 53.3\%}}
	 \put(257,43.0){\makebox(0,0)[bl]{ 49.8\%}}
	 \put(257,53.0){\makebox(0,0)[bl]{ 48.8\%}}
	 \put(257,63.0){\makebox(0,0)[bl]{ 50.4\%}}
	 \put(257,73.0){\makebox(0,0)[bl]{ 49.9\%}}
	 \put(257,83.0){\makebox(0,0)[bl]{ 48.7\%}}
	 \put(257,93.0){\makebox(0,0)[bl]{ 49.3\%}}
	 \put(257,103.0){\makebox(0,0)[bl]{ 49.5\%}}
	 \put(257,113.0){\makebox(0,0)[bl]{ 50.6\%}}
	 \put(257,123.0){\makebox(0,0)[bl]{ 51.7\%}}
	 \put(257,133.0){\makebox(0,0)[bl]{ 50.1\%}}
	 \put(257,143.0){\makebox(0,0)[bl]{ 48.6\%}}
	 \put(257,153.0){\makebox(0,0)[bl]{ 44.2\%}}
	 \put(257,163.0){\makebox(0,0)[bl]{ 32.2\%}}
	 \put(293,176.0){\makebox(0,0)[bl]{$|(b)-(c)|$}}
	 \put(293,3.0){\makebox(0,0)[bl]{ 14.1\%}} 
	 \put(293,13.0){\makebox(0,0)[bl]{ 16.6\%}}
	 \put(293,23.0){\makebox(0,0)[bl]{ 17.0\%}}
	 \put(293,33.0){\makebox(0,0)[bl]{ 14.0\%}}
	 \put(293,43.0){\makebox(0,0)[bl]{ 17.1\%}}
	 \put(293,53.0){\makebox(0,0)[bl]{ 16.1\%}}
	 \put(293,63.0){\makebox(0,0)[bl]{ 16.0\%}}
	 \put(293,73.0){\makebox(0,0)[bl]{ 14.4\%}}
	 \put(293,83.0){\makebox(0,0)[bl]{ 14.5\%}}
	 \put(293,93.0){\makebox(0,0)[bl]{ 16.1\%}}
	 \put(293,103.0){\makebox(0,0)[bl]{ 14.5\%}}
	 \put(293,113.0){\makebox(0,0)[bl]{ 10.6\%}}	 
	 \put(293,123.0){\makebox(0,0)[bl]{ 9.7\%}}	 
	 \put(293,133.0){\makebox(0,0)[bl]{ 2.8\%}}	 
	 \put(293,143.0){\makebox(0,0)[bl]{ 6.8\%}}	 
	 \put(293,153.0){\makebox(0,0)[bl]{ 5.0\%}}	 
	 \put(293,163.0){\makebox(0,0)[bl]{ 5.8\%}}	 
  \end{picture}
  \caption{Recognition test results for (5a)$\ \sharp\ see_{obj}$.}
  \label{fig:4c_5aSeeobj} 
\end{figure}

\definecolor{mygray1}{rgb}{0.5,0.5,0.5}
\begin{figure}[ht]  
  \begin{picture}(0,195)(0,-10)
	 \put(0.0,0.0){\vector(1,0){180.0}}  
	 \put(0.0,0.0){\vector(0,1){120.0}} 
	 \put(10.0,2.0){\line(0,-1){4.0}} 
	 \put(20.0,2.0){\line(0,-1){4.0}} 
	 \put(30.0,2.0){\line(0,-1){4.0}} 
	 \put(40.0,2.0){\line(0,-1){4.0}} 
	 \put(50.0,2.0){\line(0,-1){4.0}} 
	 \put(60.0,2.0){\line(0,-1){4.0}} 
	 \put(70.0,2.0){\line(0,-1){4.0}} 
	 \put(80.0,2.0){\line(0,-1){4.0}} 
	 \put(90.0,2.0){\line(0,-1){4.0}} 
	 \put(100.0,2.0){\line(0,-1){4.0}} 
	 \put(110.0,2.0){\line(0,-1){4.0}} 
	 \put(120.0,2.0){\line(0,-1){4.0}} 
	 \put(130.0,2.0){\line(0,-1){4.0}} 
	 \put(140.0,2.0){\line(0,-1){4.0}} 
	 \put(150.0,2.0){\line(0,-1){4.0}} 
	 \put(160.0,2.0){\line(0,-1){4.0}} 
	 \put(170.0,2.0){\line(0,-1){4.0}} 
	 \put(-2.0,10.0){\line(1,0){4.0}} 
	 \put(-2.0,20.0){\line(1,0){4.0}} 
	 \put(-2.0,30.0){\line(1,0){4.0}} 
	 \put(-2.0,40.0){\line(1,0){4.0}} 
	 \put(-2.0,50.0){\line(1,0){4.0}} 
	 \put(-2.0,60.0){\line(1,0){4.0}} 
	 \put(-2.0,70.0){\line(1,0){4.0}} 
	 \put(-2.0,80.0){\line(1,0){4.0}} 
	 \put(-2.0,90.0){\line(1,0){4.0}} 
	 \put(-2.0,100.0){\line(1,0){4.0}} 
	 \put(15.0,-9.0){\makebox(0,0)[bl]{ 5}}
	 \put(63.0,-9.0){\makebox(0,0)[bl]{ 10}}
	 \put(113.0,-9.0){\makebox(0,0)[bl]{ 15}}
	 \put(163.0,-9.0){\makebox(0,0)[bl]{ 20}}
	 \put(180.0,-4.0){\makebox(0,0)[bl]{ N}}
	 \put(-21,7.0){\makebox(0,0)[bl]{ 10\%}}
	 \put(-21,47.0){\makebox(0,0)[bl]{ 50\%}}
	 \put(-25,97.0){\makebox(0,0)[bl]{ 100\%}}
	 \put(-7.0,122.0){\makebox(0,0)[bl]{[\%] recognition}}
	 \put(10.0,22.5){\circle{1.5}}
	 \put(20.0,36.3){\circle{1.5}}
	 \put(30.0,54.6){\circle{1.5}}
	 \put(40.0,72.6){\circle{1.5}}
	 \put(50.0,85.8){\circle{1.5}}
	 \put(60.0,93.5){\circle{1.5}}
	 \put(70.0,96.4){\circle{1.5}}
	 \put(80.0,98.0){\circle{1.5}}
	 \put(90.0,99.4){\circle{1.5}}
	 \put(100.0,99.5){\circle{1.5}}
	 \put(110.0,99.7){\circle{1.5}}
	 \put(120.0,100.0){\circle{1.5}}
	 \put(130.0,99.9){\circle{1.5}}
	 \put(140.0,100.0){\circle{1.5}}
	 \put(150.0,100.0){\circle{1.5}}
	 \put(160.0,99.9){\circle{1.5}}
	 \put(170.0,100.0){\circle{1.5}}
	 \put(10.0,28.6){\circle*{1}}
	 \put(20.0,44.6){\circle*{1}}
	 \put(30.0,66.5){\circle*{1}}
	 \put(40.0,79.5){\circle*{1}}
	 \put(50.0,89.4){\circle*{1}}
	 \put(60.0,96.1){\circle*{1}}
	 \put(70.0,97.2){\circle*{1}}
	 \put(80.0,98.3){\circle*{1}}
	 \put(90.0,99.5){\circle*{1}}
	 \put(100.0,99.7){\circle*{1}}
	 \put(110.0,99.8){\circle*{1}}
	 \put(120.0,100.0){\circle*{1}}
	 \put(130.0,100.0){\circle*{1}}
	 \put(140.0,100.0){\circle*{1}}
	 \put(150.0,100.0){\circle*{1}}
	 \put(160.0,100.0){\circle*{1}}
	 \put(170.0,100.0){\circle*{1}}
	 \color{mygray1}
	 \put(76,75){\circle{1.5}}
	 \put(80,73){\makebox(0,0)[bl]{ Plate construction}}
	 \put(76,68){\circle*{1}}
	 \put(80,64){\makebox(0,0)[bl]{ Agent-object construction}}
	 \put(76.0,51.0){\makebox(0,0)[bl]{ QUESTION: (1b)$\ \sharp\ bite_{obj}$}}
	 \put(76.0,43.0){\makebox(0,0)[bl]{ ANSWER: $PSmith$}}
	 \put(76.0,33.0){\makebox(0,0)[bl]{ NUMBER OF TRIALS: 1000}}
	 \put(76.0,22.0){\makebox(0,0)[bl]{ (right-hand-side questions)}}	 	 
	 \color{black}
	 \put(200.0,190.0){\line(1,0){132.0}}  
	 \put(200.0,170.0){\line(1,0){132.0}}  
	 \put(200.0,160.0){\line(1,0){132.0}}  
	 \put(200.0,150.0){\line(1,0){132.0}}  
	 \put(200.0,140.0){\line(1,0){132.0}}  
	 \put(200.0,130.0){\line(1,0){132.0}}  
	 \put(200.0,120.0){\line(1,0){132.0}}  
	 \put(200.0,110.0){\line(1,0){132.0}}  
	 \put(200.0,100.0){\line(1,0){132.0}}  
	 \put(200.0,90.0){\line(1,0){132.0}}  
	 \put(200.0,80.0){\line(1,0){132.0}}  
	 \put(200.0,70.0){\line(1,0){132.0}}  
	 \put(200.0,60.0){\line(1,0){132.0}}  
	 \put(200.0,50.0){\line(1,0){132.0}}  
	 \put(200.0,40.0){\line(1,0){132.0}}  
	 \put(200.0,30.0){\line(1,0){132.0}}  
	 \put(200.0,20.0){\line(1,0){132.0}}  
	 \put(200.0,10.0){\line(1,0){132.0}}  
	 \put(200.0,0.0){\line(1,0){132.0}}  
	 \put(200.0,0.0){\line(0,1){190.0}}  
	 \put(215.0,0.0){\line(0,1){190.0}}  
	 \put(254.0,0.0){\line(0,1){190.0}}  
	 \put(290.0,0.0){\line(0,1){190.0}}  
	 \put(332.0,0.0){\line(0,1){190.0}}  
	 \put(203,3.0){\makebox(0,0)[bl]{ 20}}
	 \put(203,13.0){\makebox(0,0)[bl]{ 19}}
	 \put(203,23.0){\makebox(0,0)[bl]{ 18}}
	 \put(203,33.0){\makebox(0,0)[bl]{ 17}}
	 \put(203,43.0){\makebox(0,0)[bl]{ 16}}
	 \put(203,53.0){\makebox(0,0)[bl]{ 15}}
	 \put(203,63.0){\makebox(0,0)[bl]{ 14}}
	 \put(203,73.0){\makebox(0,0)[bl]{ 13}}
	 \put(203,83.0){\makebox(0,0)[bl]{ 12}}
	 \put(203,93.0){\makebox(0,0)[bl]{ 11}}
	 \put(203,103.0){\makebox(0,0)[bl]{ 10}}
	 \put(207,113.0){\makebox(0,0)[bl]{ 9}}
	 \put(207,123.0){\makebox(0,0)[bl]{ 8}}
	 \put(207,133.0){\makebox(0,0)[bl]{ 7}}
	 \put(207,143.0){\makebox(0,0)[bl]{ 6}}
	 \put(207,153.0){\makebox(0,0)[bl]{ 5}}
	 \put(207,163.0){\makebox(0,0)[bl]{ 4}}
	 \put(202,172.0){\makebox(0,0)[bl]{ N}}
	 \put(202,180.0){\makebox(0,0)[bl]{$(a)$}}
	 \put(224,172.0){\makebox(0,0)[bl]{{Plate}}}
	 \put(229,180.0){\makebox(0,0)[bl]{$(b)$}}
	 \put(219,3.0){\makebox(0,0)[bl]{ 100.0\%}} 
	 \put(219,13.0){\makebox(0,0)[bl]{ 99.9\%}}
	 \put(219,23.0){\makebox(0,0)[bl]{ 100.0\%}}
	 \put(219,33.0){\makebox(0,0)[bl]{ 100.0\%}}
	 \put(219,43.0){\makebox(0,0)[bl]{ 99.9\%}}
	 \put(219,53.0){\makebox(0,0)[bl]{ 100.0\%}}
	 \put(219,63.0){\makebox(0,0)[bl]{ 99.7\%}}
	 \put(219,73.0){\makebox(0,0)[bl]{ 99.5\%}}
	 \put(219,83.0){\makebox(0,0)[bl]{ 99.4\%}}
	 \put(219,93.0){\makebox(0,0)[bl]{ 98.0\%}}
	 \put(219,103.0){\makebox(0,0)[bl]{ 96.4\%}}
	 \put(219,113.0){\makebox(0,0)[bl]{ 93.5\%}}
	 \put(219,123.0){\makebox(0,0)[bl]{ 85.8\%}}
	 \put(219,133.0){\makebox(0,0)[bl]{ 72.6\%}}
	 \put(219,143.0){\makebox(0,0)[bl]{ 54.6\%}}
	 \put(219,153.0){\makebox(0,0)[bl]{ 36.3\%}}
	 \put(219,163.0){\makebox(0,0)[bl]{ 22.5\%}}
	 \put(265.5,172.0){\makebox(0,0)[bl]{{A-O}}}
	 \put(267.5,180.0){\makebox(0,0)[bl]{$(c)$}}
	 \put(257,3.0){\makebox(0,0)[bl]{ 100.0\%}} 
	 \put(257,13.0){\makebox(0,0)[bl]{ 100.0\%}}
	 \put(257,23.0){\makebox(0,0)[bl]{ 100.0\%}}
	 \put(257,33.0){\makebox(0,0)[bl]{ 100.0\%}}
	 \put(257,43.0){\makebox(0,0)[bl]{ 100.0\%}}
	 \put(257,53.0){\makebox(0,0)[bl]{ 100.0\%}}
	 \put(257,63.0){\makebox(0,0)[bl]{ 99.8\%}}
	 \put(257,73.0){\makebox(0,0)[bl]{ 99.7\%}}
	 \put(257,83.0){\makebox(0,0)[bl]{ 99.5\%}}
	 \put(257,93.0){\makebox(0,0)[bl]{ 98.3\%}}
	 \put(257,103.0){\makebox(0,0)[bl]{ 97.2\%}}
	 \put(257,113.0){\makebox(0,0)[bl]{ 96.1\%}}
	 \put(257,123.0){\makebox(0,0)[bl]{ 89.4\%}}
	 \put(257,133.0){\makebox(0,0)[bl]{ 79.5\%}}
	 \put(257,143.0){\makebox(0,0)[bl]{ 66.5\%}}
	 \put(257,153.0){\makebox(0,0)[bl]{ 44.6\%}}
	 \put(257,163.0){\makebox(0,0)[bl]{ 28.6\%}}
	 \put(293,176.0){\makebox(0,0)[bl]{$|(b)-(c)|$}}
	 \put(293,3.0){\makebox(0,0)[bl]{ 0.0\%}} 
	 \put(293,13.0){\makebox(0,0)[bl]{ 0.1\%}}
	 \put(293,23.0){\makebox(0,0)[bl]{ 0.0\%}}
	 \put(293,33.0){\makebox(0,0)[bl]{ 0.0\%}}
	 \put(293,43.0){\makebox(0,0)[bl]{ 0.1\%}}
	 \put(293,53.0){\makebox(0,0)[bl]{ 0.0\%}}
	 \put(293,63.0){\makebox(0,0)[bl]{ 0.1\%}}
	 \put(293,73.0){\makebox(0,0)[bl]{ 0.2\%}}
	 \put(293,83.0){\makebox(0,0)[bl]{ 0.1\%}}
	 \put(293,93.0){\makebox(0,0)[bl]{ 0.3\%}}
	 \put(293,103.0){\makebox(0,0)[bl]{ 0.8\%}}
	 \put(293,113.0){\makebox(0,0)[bl]{ 2.6\%}}	 
	 \put(293,123.0){\makebox(0,0)[bl]{ 3.6\%}}	 
	 \put(293,133.0){\makebox(0,0)[bl]{ 6.9\%}}	 
	 \put(293,143.0){\makebox(0,0)[bl]{ 11.9\%}}	 
	 \put(293,153.0){\makebox(0,0)[bl]{ 8.3\%}}	 
	 \put(293,163.0){\makebox(0,0)[bl]{ 6.1\%}}	 
  \end{picture}
  \caption{Recognition test results for (1b)$\ \sharp\ bite_{obj}$.}
  \label{fig:4d_1bBiteobj} 
\end{figure}
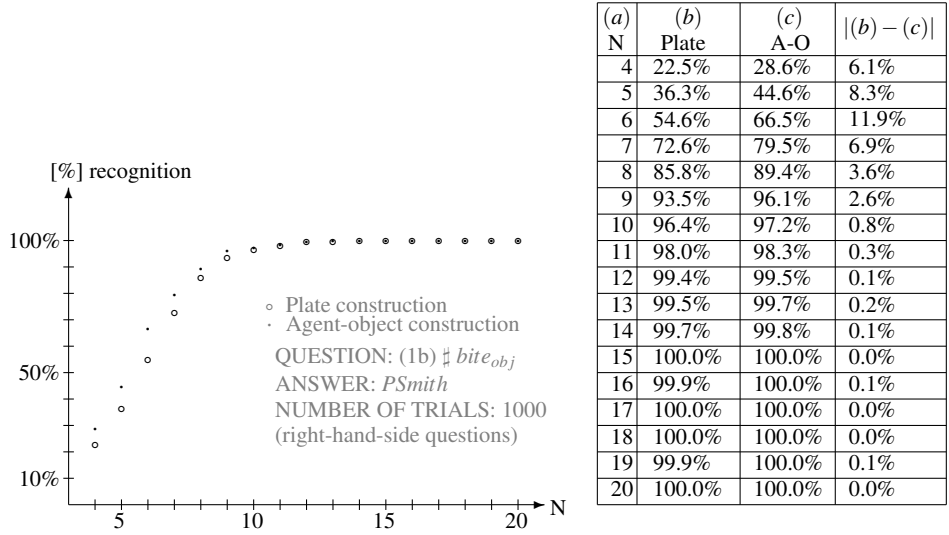

As expected, the agent-object construction seems to work better for sentences from which a rather simple information is to be derived, e.g. $PSmith\ \sharp\ name$ or (5a)$\ \sharp\ see_{obj}$ (Figures~\ref{fig:4a_PSmithName} and~\ref{fig:4b_5aSeeagt} respectively). However, when the information asked was more complex, e.g. (5a)$\ \sharp\ see_{obj}$, the Plate construction seemed more appropriate (Figure~\ref{fig:4c_5aSeeobj}). This might be so for at least two reasons:
\begin{itemize}
\item some of the blades belonging to the answer of (5a)$\ \sharp\ see_{obj}$ appear in numerous entries in the cleanup memory listed in Table~\ref{tab:chmeasurestable1}, causing the GA model to misinterpret the answers it receives after the inner products have been computed,
\item the number of blades in (5a)$\ \sharp\ see_{obj}$ is uneven when using Plate construction, hence the possibility that blades' similarities cancel each other out is smaller than in case of the agent-object construction --- such hypothesis would be backed up by test results depicted in Figure~\ref{fig:4d_1bBiteobj}.
\end{itemize}
These hypotheses led to a conclusion that perhaps a random blade should be added to those sentences that have an even number of blades, similarly to BSC.

A correct answer might not be recognized for two reasons:
\begin{itemize}
\item the correct answer has an even number of blades and their similarities cancelled each other out completely because of having opposite signs - hence such an answer does not even appear within the set of potential answers,
\item there are some pseudo-answers leading to a higher inner product because the similarities of blades of a correct answer cancelled each other partially.
\end{itemize}

Adding random extra blades that make the number of blades in a multivector odd (for short: $odding\ blades$) is a solution to the first reason why a correct answer is not recognized. Further, an odding blade acts as a distinct marker belonging only to one sentence (for sufficiently large data size) distinguishing it from other sentences, unlike the extra blade representing action in Plate construction which may appear in numerous sentences. Unfortunately, to address the second problem, we need to employ some other measurement of similarity than the inner product. We will show in Section \ref{sec:4}, that Hamming and Euclidean measures perform very well in that case.

Observation of preliminary recognition test results led to a conclusion, that sentences with an even number of blades behave quite differently than sentences with an odd number of blades. In the following tests we inspected  the average number of times that blades' similarities cancelled each other out completely during the computation of similarity via inner product.

\definecolor{myred1}{rgb}{1,0.5,0.5}
\definecolor{mygray1}{rgb}{0.5,0.5,0.5}
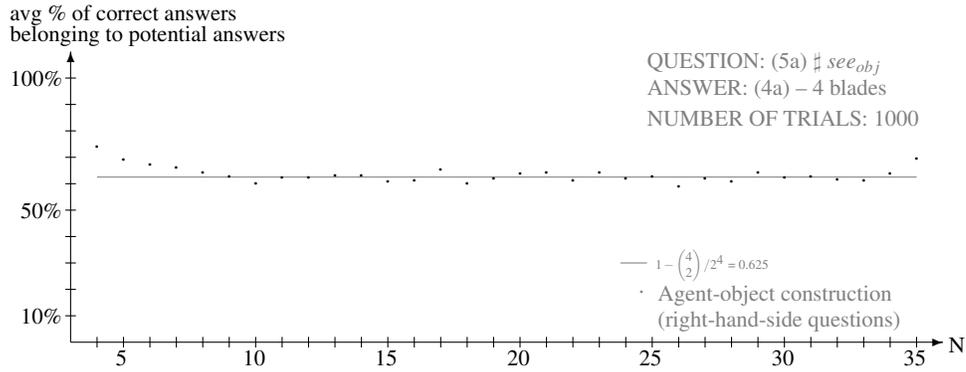
\begin{figure}[p]  
  \begin{picture}(0,135)(0,-10)
	 \put(0.0,0.0){\vector(1,0){330.0}}  
	 \put(0.0,0.0){\vector(0,1){110.0}} 
	 \put(10.0,2.0){\line(0,-1){4.0}} 
	 \put(20.0,2.0){\line(0,-1){4.0}} 
	 \put(30.0,2.0){\line(0,-1){4.0}} 
	 \put(40.0,2.0){\line(0,-1){4.0}} 
	 \put(50.0,2.0){\line(0,-1){4.0}} 
	 \put(60.0,2.0){\line(0,-1){4.0}} 
	 \put(70.0,2.0){\line(0,-1){4.0}} 
	 \put(80.0,2.0){\line(0,-1){4.0}} 
	 \put(90.0,2.0){\line(0,-1){4.0}} 
	 \put(100.0,2.0){\line(0,-1){4.0}} 
	 \put(110.0,2.0){\line(0,-1){4.0}} 
	 \put(120.0,2.0){\line(0,-1){4.0}} 
	 \put(130.0,2.0){\line(0,-1){4.0}} 
	 \put(140.0,2.0){\line(0,-1){4.0}} 
	 \put(150.0,2.0){\line(0,-1){4.0}} 
	 \put(160.0,2.0){\line(0,-1){4.0}} 
	 \put(170.0,2.0){\line(0,-1){4.0}}  
	 \put(180.0,2.0){\line(0,-1){4.0}} 
	 \put(190.0,2.0){\line(0,-1){4.0}} 
	 \put(200.0,2.0){\line(0,-1){4.0}} 
	 \put(210.0,2.0){\line(0,-1){4.0}} 
	 \put(220.0,2.0){\line(0,-1){4.0}}  
	 \put(230.0,2.0){\line(0,-1){4.0}} 
	 \put(240.0,2.0){\line(0,-1){4.0}} 
	 \put(250.0,2.0){\line(0,-1){4.0}} 
	 \put(260.0,2.0){\line(0,-1){4.0}} 
	 \put(270.0,2.0){\line(0,-1){4.0}}  
	 \put(280.0,2.0){\line(0,-1){4.0}} 
	 \put(290.0,2.0){\line(0,-1){4.0}} 
	 \put(300.0,2.0){\line(0,-1){4.0}} 
	 \put(310.0,2.0){\line(0,-1){4.0}} 
	 \put(320.0,2.0){\line(0,-1){4.0}}  
	 \put(-2.0,10.0){\line(1,0){4.0}} 
	 \put(-2.0,20.0){\line(1,0){4.0}} 
	 \put(-2.0,30.0){\line(1,0){4.0}} 
	 \put(-2.0,40.0){\line(1,0){4.0}} 
	 \put(-2.0,50.0){\line(1,0){4.0}} 
	 \put(-2.0,60.0){\line(1,0){4.0}} 
	 \put(-2.0,70.0){\line(1,0){4.0}} 
	 \put(-2.0,80.0){\line(1,0){4.0}} 
	 \put(-2.0,90.0){\line(1,0){4.0}} 
	 \put(-2.0,100.0){\line(1,0){4.0}} 
	 \put(15.0,-9.0){\makebox(0,0)[bl]{ 5}}
	 \put(63.0,-9.0){\makebox(0,0)[bl]{ 10}}
	 \put(113.0,-9.0){\makebox(0,0)[bl]{ 15}}
	 \put(163.0,-9.0){\makebox(0,0)[bl]{ 20}}
	 \put(213.0,-9.0){\makebox(0,0)[bl]{ 25}}
	 \put(263.0,-9.0){\makebox(0,0)[bl]{ 30}}
	 \put(313.0,-9.0){\makebox(0,0)[bl]{ 35}}
	 \put(330.0,-4.0){\makebox(0,0)[bl]{ N}}
	 \put(-21,7.0){\makebox(0,0)[bl]{ 10\%}}
	 \put(-21,47.0){\makebox(0,0)[bl]{ 50\%}}
	 \put(-25,97.0){\makebox(0,0)[bl]{ 100\%}}
	 \put(-25.0,120.0){\makebox(0,0)[bl]{ avg \% of correct answers }}
	 \put(-25.0,112.0){\makebox(0,0)[bl]{ belonging to potential answers}}
	 \color{mygray1}
	 \put(10.0,62.5){\line(1,0){310.0}}
	 \color{black}
	 \put(10.0,73.9){\circle*{1}}
	 \put(20.0,69.0){\circle*{1}}
	 \put(30.0,67.3){\circle*{1}}
	 \put(40.0,66.0){\circle*{1}}
	 \put(50.0,64.3){\circle*{1}}
	 \put(60.0,62.8){\circle*{1}}
	 \put(70.0,60.2){\circle*{1}}
	 \put(80.0,62.4){\circle*{1}}
	 \put(90.0,62.3){\circle*{1}}
	 \put(100.0,63.1){\circle*{1}}
	 \put(110.0,63.1){\circle*{1}}
	 \put(120.0,60.7){\circle*{1}}
	 \put(130.0,61.1){\circle*{1}}
	 \put(140.0,65.3){\circle*{1}}
	 \put(150.0,60.1){\circle*{1}}
	 \put(160.0,62.1){\circle*{1}}
	 \put(170.0,64.0){\circle*{1}}
 	 \put(180.0,64.4){\circle*{1}}
 	 \put(190.0,61.4){\circle*{1}}
 	 \put(200.0,64.2){\circle*{1}}
 	 \put(210.0,62.1){\circle*{1}}
 	 \put(220.0,62.6){\circle*{1}}
 	 \put(230.0,59.1){\circle*{1}}
 	 \put(240.0,62.1){\circle*{1}}
 	 \put(250.0,60.9){\circle*{1}}
 	 \put(260.0,64.4){\circle*{1}}
 	 \put(270.0,62.3){\circle*{1}}
 	 \put(280.0,62.8){\circle*{1}}
 	 \put(290.0,61.8){\circle*{1}}
 	 \put(300.0,61.1){\circle*{1}}
 	 \put(310.0,63.7){\circle*{1}}
 	 \put(320.0,69.7){\circle*{1}}
	 \color{mygray1}
	 \put(208,30){\line(1,0){10.0}}
	 \color{mygray1}
	 \put(220,23){\makebox(0,0)[bl]{\tiny{ $1 - \begin{pmatrix}4 \cr 2\end{pmatrix}/2^{4}$} = 0.625}}	 
	 \put(216,19){\circle*{1}}
	 \put(220,14){\makebox(0,0)[bl]{ Agent-object construction}}
	 \put(220,4){\makebox(0,0)[bl]{ (right-hand-side questions)}}	 	 
	 \put(216.0,101.0){\makebox(0,0)[bl]{ QUESTION: (5a)$\ \sharp\ see_{obj}$}}
	 \put(216.0,92.0){\makebox(0,0)[bl]{ ANSWER: (4a) -- 4 blades}}
	 \put(216.0,82.0){\makebox(0,0)[bl]{ NUMBER OF TRIALS: 1000}}
 	 \color{black}
  \end{picture}
  \caption{The average number of times a correct answer appears within the set of potential answers ((5a)$\ \sharp\ see_{obj}$).}
  \label{fig:wasnonzeroG} 
\end{figure}

\definecolor{myred1}{rgb}{1,0.5,0.5}
\definecolor{mygray1}{rgb}{0.5,0.5,0.5}
\begin{figure}[p]  
  \begin{picture}(0,137)(0,-10)
	 \put(0.0,0.0){\vector(1,0){330.0}}  
	 \put(0.0,0.0){\vector(0,1){110.0}} 
	 \put(10.0,2.0){\line(0,-1){4.0}} 
	 \put(20.0,2.0){\line(0,-1){4.0}} 
	 \put(30.0,2.0){\line(0,-1){4.0}} 
	 \put(40.0,2.0){\line(0,-1){4.0}} 
	 \put(50.0,2.0){\line(0,-1){4.0}} 
	 \put(60.0,2.0){\line(0,-1){4.0}} 
	 \put(70.0,2.0){\line(0,-1){4.0}} 
	 \put(80.0,2.0){\line(0,-1){4.0}} 
	 \put(90.0,2.0){\line(0,-1){4.0}} 
	 \put(100.0,2.0){\line(0,-1){4.0}} 
	 \put(110.0,2.0){\line(0,-1){4.0}} 
	 \put(120.0,2.0){\line(0,-1){4.0}} 
	 \put(130.0,2.0){\line(0,-1){4.0}} 
	 \put(140.0,2.0){\line(0,-1){4.0}} 
	 \put(150.0,2.0){\line(0,-1){4.0}} 
	 \put(160.0,2.0){\line(0,-1){4.0}} 
	 \put(170.0,2.0){\line(0,-1){4.0}}  
	 \put(180.0,2.0){\line(0,-1){4.0}} 
	 \put(190.0,2.0){\line(0,-1){4.0}} 
	 \put(200.0,2.0){\line(0,-1){4.0}} 
	 \put(210.0,2.0){\line(0,-1){4.0}} 
	 \put(220.0,2.0){\line(0,-1){4.0}}  
	 \put(230.0,2.0){\line(0,-1){4.0}} 
	 \put(240.0,2.0){\line(0,-1){4.0}} 
	 \put(250.0,2.0){\line(0,-1){4.0}} 
	 \put(260.0,2.0){\line(0,-1){4.0}} 
	 \put(270.0,2.0){\line(0,-1){4.0}}  
	 \put(280.0,2.0){\line(0,-1){4.0}} 
	 \put(290.0,2.0){\line(0,-1){4.0}} 
	 \put(300.0,2.0){\line(0,-1){4.0}} 
	 \put(310.0,2.0){\line(0,-1){4.0}} 
	 \put(320.0,2.0){\line(0,-1){4.0}}  
	 \put(-2.0,10.0){\line(1,0){4.0}} 
	 \put(-2.0,20.0){\line(1,0){4.0}} 
	 \put(-2.0,30.0){\line(1,0){4.0}} 
	 \put(-2.0,40.0){\line(1,0){4.0}} 
	 \put(-2.0,50.0){\line(1,0){4.0}} 
	 \put(-2.0,60.0){\line(1,0){4.0}} 
	 \put(-2.0,70.0){\line(1,0){4.0}} 
	 \put(-2.0,80.0){\line(1,0){4.0}} 
	 \put(-2.0,90.0){\line(1,0){4.0}} 
	 \put(-2.0,100.0){\line(1,0){4.0}} 
	 \put(15.0,-9.0){\makebox(0,0)[bl]{ 5}}
	 \put(63.0,-9.0){\makebox(0,0)[bl]{ 10}}
	 \put(113.0,-9.0){\makebox(0,0)[bl]{ 15}}
	 \put(163.0,-9.0){\makebox(0,0)[bl]{ 20}}
	 \put(213.0,-9.0){\makebox(0,0)[bl]{ 25}}
	 \put(263.0,-9.0){\makebox(0,0)[bl]{ 30}}
	 \put(313.0,-9.0){\makebox(0,0)[bl]{ 35}}
	 \put(330.0,-4.0){\makebox(0,0)[bl]{ N}}
	 \put(-21,7.0){\makebox(0,0)[bl]{ 10\%}}
	 \put(-21,47.0){\makebox(0,0)[bl]{ 50\%}}
	 \put(-25,97.0){\makebox(0,0)[bl]{ 100\%}}
	 \put(-25.0,120.0){\makebox(0,0)[bl]{ avg \% of correct answers }}
	 \put(-25.0,112.0){\makebox(0,0)[bl]{ belonging to potential answers}}
	 \color{mygray1}
	 \put(10.0,72.6563){\line(1,0){310.0}}
	 \color{black}
	 \put(10.0,87.3){\circle*{1}}
	 \put(20.0,83.8){\circle*{1}}
	 
	 \put(30.0,77.7){\circle*{1}}
	 \put(40.0,75.5){\circle*{1}}
	 \put(50.0,71.8){\circle*{1}}
	 \put(60.0,73.7){\circle*{1}}
	 \put(70.0,72.5){\circle*{1}}
	 
	 \put(80.0,73.5){\circle*{1}}
	 \put(90.0,71.2){\circle*{1}}
	 \put(100.0,70.0){\circle*{1}}
	 \put(110.0,73.6){\circle*{1}}
	 \put(120.0,73.7){\circle*{1}}
	 
	 \put(130.0,75.1){\circle*{1}}
	 \put(140.0,72.5){\circle*{1}}
	 \put(150.0,74.8){\circle*{1}}
	 \put(160.0,74.3){\circle*{1}}
	 \put(170.0,73.8){\circle*{1}}

 	 \put(180.0,73.7){\circle*{1}}
 	 \put(190.0,72.4){\circle*{1}}
 	 \put(200.0,74.2){\circle*{1}}
 	 \put(210.0,74.9){\circle*{1}}
 	 \put(220.0,74.3){\circle*{1}}
 	 
 	 \put(230.0,69.8){\circle*{1}}
 	 \put(240.0,71.9){\circle*{1}}
 	 \put(250.0,71.5){\circle*{1}}
 	 \put(260.0,74.3){\circle*{1}}
 	 \put(270.0,72.4){\circle*{1}}
 	 
 	 \put(280.0,72.4){\circle*{1}}
 	 \put(290.0,72.3){\circle*{1}}
 	 \put(300.0,71.4){\circle*{1}}
 	 \put(310.0,75.4){\circle*{1}}
 	 \put(320.0,71.1){\circle*{1}}
 	 
	 \color{mygray1}
	 \put(208,30){\line(1,0){10.0}}
	 \color{mygray1}
	 \put(220,23){\makebox(0,0)[bl]{\tiny{ $1 - \begin{pmatrix}8 \cr 4\end{pmatrix}/2^{8}$} = 0.726563}}	 
	 \put(216,19){\circle*{1}}
	 \put(220,14){\makebox(0,0)[bl]{ Agent-object construction}}
	 \put(220,4){\makebox(0,0)[bl]{ (right-hand-side questions)}}	 	 
	 \put(216.0,101.0){\makebox(0,0)[bl]{ QUESTION: (5b)$\ \sharp\ see_{obj}$}}
	 \put(216.0,92.0){\makebox(0,0)[bl]{ ANSWER: (4b) -- 8 blades}}
	 \put(216.0,82.0){\makebox(0,0)[bl]{ NUMBER OF TRIALS: 1000}}
 	 \color{black}
  \end{picture}
  \caption{The average number of times a correct answer appears within the set of potential answers ((5b)$\ \sharp\ see_{obj}$).}
  \label{fig:wasnonzeroG1} 
\end{figure}
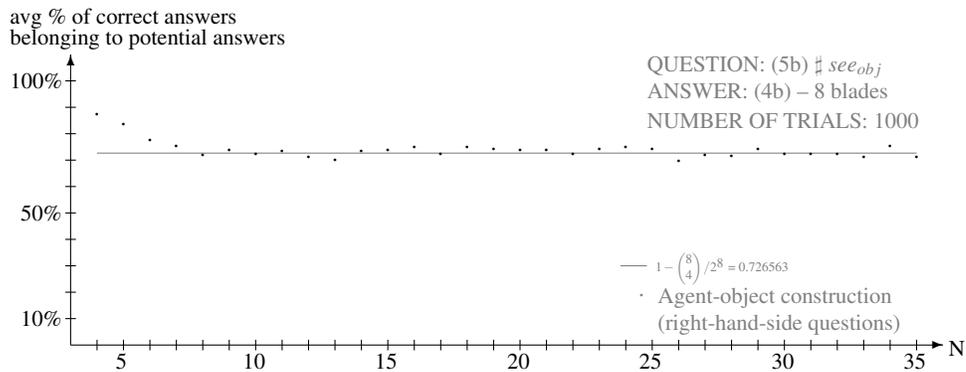

\definecolor{myred1}{rgb}{1,0.5,0.5}
\definecolor{mygray1}{rgb}{0.5,0.5,0.5}
\begin{figure}[p]  
  \begin{picture}(0,137)(0,-10)
	 \put(0.0,0.0){\vector(1,0){330.0}}  
	 \put(0.0,0.0){\vector(0,1){110.0}} 
	 \put(10.0,2.0){\line(0,-1){4.0}} 
	 \put(20.0,2.0){\line(0,-1){4.0}} 
	 \put(30.0,2.0){\line(0,-1){4.0}} 
	 \put(40.0,2.0){\line(0,-1){4.0}} 
	 \put(50.0,2.0){\line(0,-1){4.0}} 
	 \put(60.0,2.0){\line(0,-1){4.0}} 
	 \put(70.0,2.0){\line(0,-1){4.0}} 
	 \put(80.0,2.0){\line(0,-1){4.0}} 
	 \put(90.0,2.0){\line(0,-1){4.0}} 
	 \put(100.0,2.0){\line(0,-1){4.0}} 
	 \put(110.0,2.0){\line(0,-1){4.0}} 
	 \put(120.0,2.0){\line(0,-1){4.0}} 
	 \put(130.0,2.0){\line(0,-1){4.0}} 
	 \put(140.0,2.0){\line(0,-1){4.0}} 
	 \put(150.0,2.0){\line(0,-1){4.0}} 
	 \put(160.0,2.0){\line(0,-1){4.0}} 
	 \put(170.0,2.0){\line(0,-1){4.0}}  
	 \put(180.0,2.0){\line(0,-1){4.0}} 
	 \put(190.0,2.0){\line(0,-1){4.0}} 
	 \put(200.0,2.0){\line(0,-1){4.0}} 
	 \put(210.0,2.0){\line(0,-1){4.0}} 
	 \put(220.0,2.0){\line(0,-1){4.0}}  
	 \put(230.0,2.0){\line(0,-1){4.0}} 
	 \put(240.0,2.0){\line(0,-1){4.0}} 
	 \put(250.0,2.0){\line(0,-1){4.0}} 
	 \put(260.0,2.0){\line(0,-1){4.0}} 
	 \put(270.0,2.0){\line(0,-1){4.0}}  
	 \put(280.0,2.0){\line(0,-1){4.0}} 
	 \put(290.0,2.0){\line(0,-1){4.0}} 
	 \put(300.0,2.0){\line(0,-1){4.0}} 
	 \put(310.0,2.0){\line(0,-1){4.0}} 
	 \put(320.0,2.0){\line(0,-1){4.0}}  
	 \put(-2.0,10.0){\line(1,0){4.0}} 
	 \put(-2.0,20.0){\line(1,0){4.0}} 
	 \put(-2.0,30.0){\line(1,0){4.0}} 
	 \put(-2.0,40.0){\line(1,0){4.0}} 
	 \put(-2.0,50.0){\line(1,0){4.0}} 
	 \put(-2.0,60.0){\line(1,0){4.0}} 
	 \put(-2.0,70.0){\line(1,0){4.0}} 
	 \put(-2.0,80.0){\line(1,0){4.0}} 
	 \put(-2.0,90.0){\line(1,0){4.0}} 
	 \put(-2.0,100.0){\line(1,0){4.0}} 
	 \put(15.0,-9.0){\makebox(0,0)[bl]{ 5}}
	 \put(63.0,-9.0){\makebox(0,0)[bl]{ 10}}
	 \put(113.0,-9.0){\makebox(0,0)[bl]{ 15}}
	 \put(163.0,-9.0){\makebox(0,0)[bl]{ 20}}
	 \put(213.0,-9.0){\makebox(0,0)[bl]{ 25}}
	 \put(263.0,-9.0){\makebox(0,0)[bl]{ 30}}
	 \put(313.0,-9.0){\makebox(0,0)[bl]{ 35}}
	 \put(330.0,-4.0){\makebox(0,0)[bl]{ N}}
	 \put(-21,7.0){\makebox(0,0)[bl]{ 10\%}}
	 \put(-21,47.0){\makebox(0,0)[bl]{ 50\%}}
	 \put(-25,97.0){\makebox(0,0)[bl]{ 100\%}}
	 \put(-25.0,120.0){\makebox(0,0)[bl]{ avg \% of correct answers }}
	 \put(-25.0,112.0){\makebox(0,0)[bl]{ belonging to potential answers}}
	 \color{mygray1}
	 \put(10.0,75.3906){\line(1,0){310.0}}
	 \color{black}
	 \put(10.0,90.8){\circle*{1}}
	 \put(20.0,86.7){\circle*{1}}
	 
	 \put(30.0,80.4){\circle*{1}}
	 \put(40.0,77.3){\circle*{1}}
	 \put(50.0,77.3){\circle*{1}}
	 \put(60.0,76.3){\circle*{1}}
	 \put(70.0,73.2){\circle*{1}}
	 
	 \put(80.0,75.8){\circle*{1}}
	 \put(90.0,73.0){\circle*{1}}
	 \put(100.0,75.9){\circle*{1}}
	 \put(110.0,74.6){\circle*{1}}
	 \put(120.0,75.9){\circle*{1}}
	 
	 \put(130.0,73.3){\circle*{1}}
	 \put(140.0,74.9){\circle*{1}}
	 \put(150.0,76.1){\circle*{1}}
	 \put(160.0,74.8){\circle*{1}}
	 \put(170.0,74.6){\circle*{1}}

 	 \put(180.0,75.4){\circle*{1}}
 	 \put(190.0,75.0){\circle*{1}}
 	 \put(200.0,74.2){\circle*{1}}
 	 \put(210.0,71.7){\circle*{1}}
 	 \put(220.0,76.8){\circle*{1}}
 	 
 	 \put(230.0,75.0){\circle*{1}}
 	 \put(240.0,76.7){\circle*{1}}
 	 \put(250.0,73.8){\circle*{1}}
 	 \put(260.0,76.4){\circle*{1}}
 	 \put(270.0,77.5){\circle*{1}}
 	 
 	 \put(280.0,73.5){\circle*{1}}
 	 \put(290.0,77.2){\circle*{1}}
 	 \put(300.0,75.3){\circle*{1}}
 	 \put(310.0,78.9){\circle*{1}}
 	 \put(320.0,73.5){\circle*{1}}
 	 
	 \color{mygray1}
	 \put(208,30){\line(1,0){10.0}}
	 \color{mygray1}
	 \put(220,23){\makebox(0,0)[bl]{\tiny{ $1 - \begin{pmatrix}10 \cr 5\end{pmatrix}/2^{10}$} = 0.753906}}	 
	 \put(216,19){\circle*{1}}
	 \put(220,14){\makebox(0,0)[bl]{ Agent-object construction}}
	 \put(220,4){\makebox(0,0)[bl]{ (right-hand-side questions)}}	 
	 \put(216.0,106.0){\makebox(0,0)[bl]{ QUESTION: (3d)$\ \sharp\ see_{obj}$}}
	 \put(216.0,97.0){\makebox(0,0)[bl]{ ANSWER: (1d) -- 10 blades}}
	 \put(216.0,87.0){\makebox(0,0)[bl]{ NUMBER OF TRIALS: 1000}}
 	 \color{black}
  \end{picture}
  \caption{The average number of times a correct answer appears within the set of potential answers ((3d)$\ \sharp\ see_{obj}$).}
  \label{fig:wasnonzero3d} 
\end{figure}
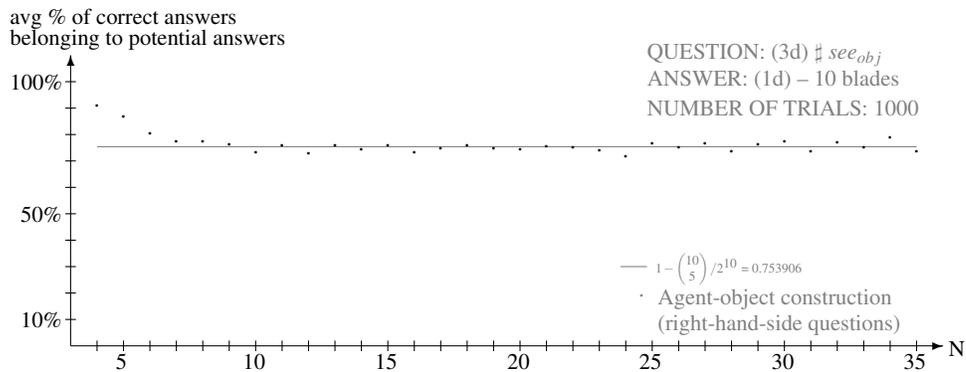

The complete cancellation of similarities takes place only when exactly half of blades of the correct answer carry a plus sign and the other half carry a minus sign. If the correct answer has $2K\geq2$ blades, then the probability of exactly half of blades having the same sign is 
\begin{equation}
\frac{\begin{pmatrix}2K \cr K\end{pmatrix}}{2^{2K}}
\end{equation}
under assumption, that the sentence set is chosen completely at random without the interference of the experimenter. Figures~\ref{fig:wasnonzeroG} through \ref{fig:wasnonzero3d} show three examples of questions yielding an even-number blade answer and the average number of times their blades' similarities cancelled each other out completely.

\subsection{Appropriate-hand-side reversed questions}
\label{sec:3:b}

Let us recall some roles and fillers 
\begin{eqnarray}
\begin{matrix}
see_{agt} &=& c_{00101}, &\quad\quad\quad John &=& c_{00101},\cr
see_{obj} &=& c_{01010}, &\quad\quad\quad Pat &=& c_{10000},\cr
bite_{agt} &=& c_{10110}, &\quad\quad\quad Fido &=& c_{10001},\cr
bite_{obj} &=& c_{00001},
\end{matrix}
\end{eqnarray}
as well as two sentences mentioned in Table \ref{tab:chmeasurestable1}
\begin{eqnarray}
\textrm{(1a) }\ \ & bite_{agt}\ast Fido +bite_{obj}\ast Pat &= c_{00111} - c_{10001}, \\
\textrm{(3a) }\ \ & see_{agt}\ast John + see_{obj}\ast\textrm{(1a)} &= -c_{00000} - c_{01101} - c_{11011}.
\end{eqnarray}
The answers to questions (3a)$\ \sharp\ see_{obj}$ and (3a)$\ \sharp\ John$ should be computed in different ways
\begin{eqnarray}
\textrm{(3a)}\ \sharp\ see_{obj} &=& see_{obj}^{+} \ast \textrm{(3a)}\ \approx\ \textrm{(1a)}, \\
\textrm{(3a)}\ \sharp\ John &=& \textrm{(3a)}\ast John^{+}\ \approx\ see_{agt}.
\end{eqnarray}
We will concentrate only on the first question
\begin{eqnarray}
see_{obj}^{+} \ast \textrm{(3a)} &=& c_{01010}^{+}( -c_{00000} - c_{01101} - c_{11011} )\nonumber\\
&=& c_{01010}( c_{00000} + c_{01101} + c_{11011} )\nonumber\\
&=& c_{01010} + c_{00111} - c_{10001}\\
&=& \textrm{noise} + \textrm{(1a)} = \textrm{(1a)}'.
\end{eqnarray}
Only two elements of the clean-up memory are similar to (1a)$'$
\begin{eqnarray}
|\langle see_{obj} | \textrm{(1a)}'\rangle| &=& 1,\\
|\langle\textrm{(1a)} | \textrm{(1a)}'\rangle| &=& |( c_{00111} - c_{10001} )\cdot(c_{01010} + c_{00111} - c_{10001})|\nonumber\\
&=& |c_{00111} \cdot c_{00111} + c_{10001}\cdot c_{10001}| \nonumber\\
&=& |-\textbf{1} - \textbf{1}| = 2,
\end{eqnarray}
where $see_{obj}$ is similar to the noise term only by accident.

Asking reversed questions on the appropriate side has one huge advantage over fixed-hand-side questions: no similarities cancel each other out neither completely nor partially while similarity is being computed, hence there is no need for adding odding vectors. For small data size blades may cancel each other out at the moment a sentence is created. Nevertheless, in all cases recognition will quickly reach 100\% and the only problem that might appear is that several items of the clean-up memory might be equally similar.


\section{Other measures of similarity}
\label{sec:4}

The inner product is not the only way to measure the similarity of concepts stored in the clean-up memory. This Section comments on the use of matrix representation and its advantages in the unavoidable presence of similarity cancellation and many equally probable answers. We will show, that comparison by Hamming and Euclidean measures gives promising results such cases.

\subsection{Matrix representation}
\label{sec:3:matrix}

\begin{figure}[ht!]
  \begin{picture}(0,45)(0,10)
  \includegraphics[width=\textwidth]{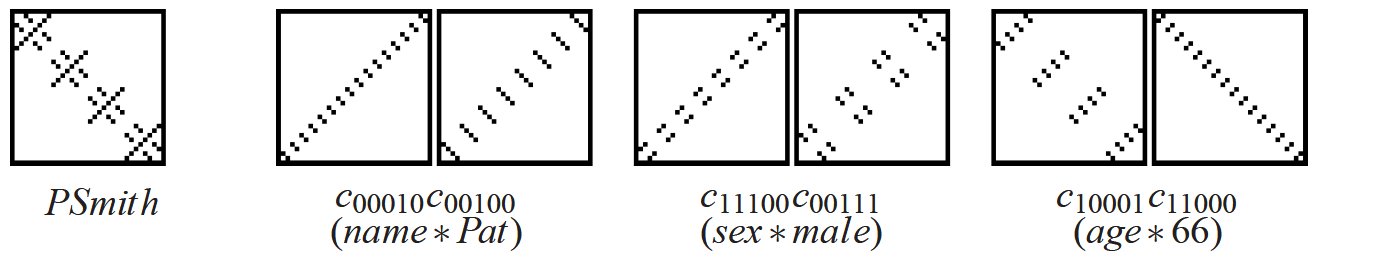}	  
  \end{picture}
\caption{$PSmith$ and its blades for $n = 5$.}
\label{fig:1}
\end{figure}

\begin{figure}[ht!]
  \begin{picture}(0,175)(0,10)
      \includegraphics[width=\textwidth]{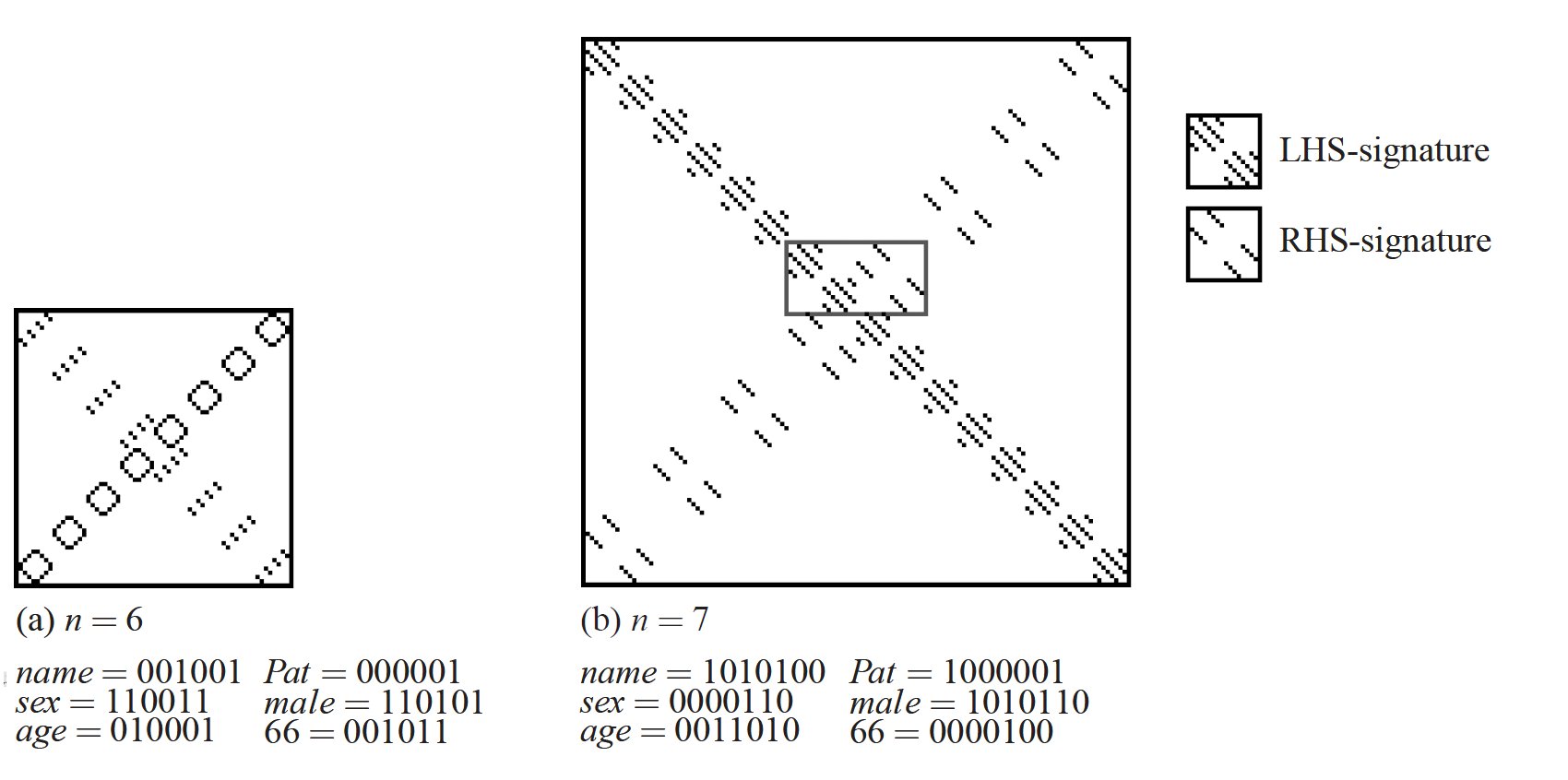}	  
  \end{picture}
  \caption{An example of matrix representation of $PSmith$ for $n\in\{6,7\}$.}
  \label{fig:2} 
\end{figure}

Matrix representations\index{matrix representation} of GA, although not efficient, are useful for performing cross-checks of various GA constructions and algorithms. An arbitrary $n$-bit record can be encoded into the matrix algebra known as Cartan representation\index{Cartan representation} of Clifford algebras as follows
\begin{eqnarray}
b_{2k}
&=&
\underbrace{\sigma_1\otimes\dots\otimes \sigma_1}_{n-k}
\otimes\,\sigma_2\otimes
\underbrace{1\otimes\dots\otimes 1}_{k-1},\label{eq:cartan1}\\
b_{2k-1}
&=&
\underbrace{\sigma_1\otimes\dots\otimes \sigma_1}_{n-k}
\otimes\,\sigma_3\otimes
\underbrace{1\otimes\dots\otimes 1}_{k-1},\label{eq:cartan2}
\end{eqnarray}
using Pauli's matrices
\begin{eqnarray}
\sigma_1
=
\left(
\begin{array}{cc}
0 & 1\\
1 & 0
\end{array}
\right),
\quad
\sigma_2
=
\left(
\begin{array}{cc}
0 & -i\\
i & 0
\end{array}
\right)
,
\quad
\sigma_3
=
\left(
\begin{array}{cc}
1 & 0\\
0 & -1
\end{array}
\right).
\end{eqnarray}
Let us once again consider the roles and fillers of $PSmith$
\begin{eqnarray}
\left.
\begin{array}{rcl}
Pat &=& c_{00100},\\
male &=& c_{00111},\\
66 &=& c_{11000},
\end{array}
\right\}{\rm fillers}\\
\left.
\begin{array}{rcl}
name &=& c_{00010},\\
sex &=& c_{11100},\\
age &=& c_{10001},
\end{array}
\right\}{\rm roles}
\end{eqnarray}
as described in Section \ref{sec:2}. Their explicit matrix representations are
\begin{eqnarray}
Pat &=& c_{00100}=b_3=
\sigma_1 \otimes \sigma_1 \otimes \sigma_1 \otimes \sigma_3 \otimes 1, \\
male &=& c_{00111}=b_3 b_4 b_5 \nonumber\\
&=& 
(\sigma_1 \otimes \sigma_1 \otimes \sigma_1 \otimes \sigma_3 \otimes 1)
(\sigma_1 \otimes \sigma_1 \otimes \sigma_1 \otimes \sigma_2 \otimes 1)
(\sigma_1 \otimes \sigma_1 \otimes \sigma_3 \otimes 1 \otimes 1)\nonumber\\
&=& \sigma_1 \otimes \sigma_1 \otimes \sigma_3 \otimes (-i\sigma_1) \otimes 1,\\
66 &=& c_{11000} = b_1 b_2 =
(\sigma_1 \otimes \sigma_1 \otimes \sigma_1 \otimes \sigma_1 \otimes \sigma_3)
(\sigma_1 \otimes \sigma_1 \otimes \sigma_1 \otimes \sigma_1 \otimes \sigma_2)\nonumber\\
&=& 1 \otimes 1 \otimes 1 \otimes 1 \otimes (-i\sigma_1),\\
name &=& c_{00010}=b_4=\sigma_1 \otimes \sigma_1 \otimes \sigma_1 \otimes \sigma_2 \otimes 1,\\
sex &=& c_{11100}=b_1 b_2 b_3\nonumber\\
&=&
(\sigma_1 \otimes \sigma_1 \otimes \sigma_1 \otimes \sigma_1 \otimes \sigma_3)
(\sigma_1 \otimes \sigma_1 \otimes \sigma_1 \otimes \sigma_1 \otimes \sigma_2)
(\sigma_1 \otimes \sigma_1 \otimes \sigma_1 \otimes \sigma_3 \otimes 1)\nonumber\\
&=& \sigma_1 \otimes \sigma_1 \otimes \sigma_3 \otimes 1 \otimes (-i\sigma_1),\\
age &=& c_{10001}=b_1 b_5=
(\sigma_1 \otimes \sigma_1 \otimes \sigma_1 \otimes \sigma_1 \otimes \sigma_3)
(\sigma_1 \otimes \sigma_1 \otimes \sigma_3 \otimes 1 \otimes 1)\nonumber\\
&=&
1 \otimes 1 \otimes (-i\sigma_2) \otimes \sigma_1 \otimes \sigma_3.
\end{eqnarray}
Figure \ref{fig:1} shows six blades making up $PSmith$ for $n=5$ and Figure \ref{fig:2} shows the matrix representation of $PSmith$ for $n\in\{6,7\}$, black dots indicate nonzero matrix entries.

The regularity of patterns placed along the diagonals is not an accident. Consider Cartan representation of blades $b_{2k}$ and $b_{2k-1}$ --- the shortest sequence of $n-k$ $\sigma_1$'s will occur for $k=\lceil \frac{n}{2} \rceil$ (in other words, in blade $b_n$). Therefore each blade $b_1,\dots,b_n$ has at least $\lfloor \frac{n}{2} \rfloor$ of $\sigma_1$'s placed at the beginning of the formula describing its representation. Hence, there are exactly $2^{\lfloor \frac{n}{2} \rfloor}$ ``boxes" of patterns placed along one of the diagonals, each one of dimensions $2^{\lceil \frac{n}{2} \rceil} \times 2^{\lceil \frac{n}{2} \rceil}$. To extract a part individual for a given blade, one needs to consider only the last $\lceil \frac{n}{2} \rceil +1$ of $\sigma$'s or unit matrices belonging to its representation --- the extra $\sigma_1$ bearing the number $n-\lceil \frac{n}{2} \rceil$ is needed to preserve the direction of the diagonal --- either ``top left to bottom right" or ``top right to bottom left". If $c=b_{\alpha_1}\dots b_{\alpha_m}$ is a blade representing an atomic object in the clean-up memory, say $Pat$, then such an object has a ``top left to bottom right" orientation if and only if $m\equiv 0 (\textrm{mod}\ 2)$. Therefore we can reformulate equations (\ref{eq:cartan1}) and (\ref{eq:cartan2}) in the following way
\begin{equation}
b_{2k}
=
\underbrace{
\sigma_1\otimes\dots\otimes \sigma_1
}_{\lceil \frac{n}{2}\rceil-k+1}\otimes\,\sigma_2\otimes
\underbrace{\textbf{1}\otimes\dots\otimes \textbf{1}}_{k-1},
\end{equation}
\begin{equation}
b_{2k-1}
=
\underbrace{
\sigma_1\otimes\dots\otimes \sigma_1
}_{\lceil \frac{n}{2} \rceil-k +1}\otimes\,\sigma_3\otimes
\underbrace{\textbf{1}\otimes\dots\otimes \textbf{1}}_{k-1}.
\end{equation}

Consider once again the representation of $PSmith$ depicted in Figure~\ref{fig:2}b. To distinguish $PSmith$ from other object we only need to store two of its ``boxes" --- each ``box" lying along a different diagonal, Figure~\ref{fig:2}b shows such two parts. We will call the two different ``boxes" \textit{left-hand-side signatures} and \textit{right-hand-side signatures} depending on the corner the diagonal is anchored to at the top of the matrix. It is worth noticing that signatures for $n=2k-1$ and $n=2k$ are of the same size, which causes some test results diagrams to resemble step functions.

Note that the use of tensor products\index{tensor product} in GA bears no resemblance to Smolensky's model, as the rank of a tensor does not increase with the growing complexity of a sentence.

\subsection{The Hamming measure of similarity}
\label{sec:3:hamming}

The first most obvious method of comparing two matrices or their signatures would be to compute the number of entries they have in common and the number of entries they differ by. Let $X=[x_{ij}]$ and $Y=[y_{ij}]$ be signatures of matrices, i.e. $i\in\{1,\dots 2^{\lceil \frac{n}{2} \rceil+1}\}, j\in\{1,\dots 2^{\lceil \frac{n}{2} \rceil}\}$. Let
\begin{eqnarray}
c(x_{ij},y_{ij}) &=& \left\{ 
\begin{array}{ll}
1 &\quad\textrm{ if}\ x_{ij}\neq 0\ \textrm{and}\ y_{ij}\neq 0, \\
0 &\quad\textrm{ otherwise},
\end{array}
\right. \\
u(x_{ij},y_{ij}) &=& 1 - c(x_{ij},y_{ij}).
\end{eqnarray}
Now let us count the number of $common\ points$ and $uncommon\ points$
\begin{eqnarray}
C(X,Y) = \sum\limits_{i,j} c(x_{ij},y_{ij}), \\
U(X,Y) = \sum\limits_{i,j} u(x_{ij},y_{ij}). 
\end{eqnarray}
Finally, the Hamming\index{Hamming distance} measure for comparing the signatures of matrices computes the ratio of common and uncommon points
\begin{equation}
H(X,Y) = \left\{ \begin{array}{ll}
\frac{C(X,Y)}{U(X,Y)} & \quad\textrm{if}\ U(X,Y)\neq 0, \\
\infty & \quad\textrm{otherwise}.
\end{array}
\right.
\end{equation}
Such a measure of similarity is fairly fast to calculate since it does not involve computing any mathematical operations except addition and the final division of C(X,Y) and U(X,Y).

\definecolor{mygray1}{rgb}{0.5,0.5,0.5}
\definecolor{mygreen1}{rgb}{0,0.8,0}
\definecolor{myblue1}{rgb}{0,0,0.9}
\begin{figure}[p]		
  %
  \caption{Recognition test via inner product, Hamming and Euclidean measure for (4a)$\ \sharp\ cause_{obj}$.}
  \label{fig:HEF}   
\end{figure}

\definecolor{mygray1}{rgb}{0.5,0.5,0.5}
\definecolor{mygreen1}{rgb}{0,0.8,0}
\definecolor{myblue1}{rgb}{0,0,0.9}
\begin{figure}[p]		
  %
  \caption{Recognition test via inner product, Hamming and Euclidean measure for (3b)$\ \sharp\ see_{obj}$.}
  \label{fig:HED1}   
\end{figure}

\definecolor{mygray1}{rgb}{0.5,0.5,0.5}
\definecolor{mygreen1}{rgb}{0,0.8,0}
\definecolor{myblue1}{rgb}{0,0,0.9}
\begin{figure}[p]		
  %
  \caption{Recognition test via inner product, Hamming and Euclidean measure for (5b)$\ \sharp\ see_{obj}$.}
  \label{fig:HEG1}   
\end{figure}

\subsection{The Euclidean measure of similarity}
\label{sec:3:euclidean}

The second most obvious method for computing matrix similarity is via Euclidean distance\index{Euclidean distance}. Again, let $X=[x_{ij}]$ and $Y=[y_{ij}]$ be signatures of matrices for $i\in\{1,\dots 2^{\lceil \frac{n}{2} \rceil+1}\}, j\in\{1,\dots 2^{\lceil \frac{n}{2} \rceil}\}$. Let
\begin{equation}
E(X,Y) = \left\{ \begin{array}{ll}
\frac{1}{\sum\limits_{i,j} \sqrt{||x_{ij}|^2 - |y_{ij}|^2|}} &  \quad\textrm{if}\ \sum\limits_{i,j} \sqrt{||x_{ij}|^2 - |y_{ij}|^2|}\neq 0, \\
\infty &  \quad\textrm{otherwise}.
\end{array}
\right.
\end{equation}
This kind of measure uses more mathematical operations requiring greater time to compute --- the modulus of a complex number, multiplication and the square root. The Hamming measure involved calculating only addition and the ratio of common and uncommon points. Calculating the ratio in both measures results in those measures taking on a role of ``probability" that the matrices are alike rather than describing the distance between them, therefore one should avoid calling those measures ``metrics".

\subsection{Performance of Hamming and Euclidean measures}
\label{sec:4:HE}

In this Section we present some test results comparing the effectiveness of Hamming and Euclidean measures against the computation of similarity by inner product. These tests are conducted on the data set presented in Table~\ref{tab:chmeasurestable1}. Once the inner product test indicates more than one potential answer, Hamming and Euclidean measures are employed upon the subset of the potential answers --- not upon the whole clean-up memory. Figures \ref{fig:HEF} through \ref{fig:HEG1} show test results for sentences with various numbers of blades using two types of construction: agent-object construction and agent-object construction with odding blades.

There was no significant difference between results obtained using the agent-object construction with odding blades and those obtained with the help of Plate construction, therefore results for Plate construction are not presented in the diagrams. Nevertheless, it is more in the spirit of distributed representations to use agent-object construction with odding blades since the additional blade is drawn at random, whereas the use of Plate construction makes data more predictable. Poor recognition in case of the agent-object construction without odding blades results from complete or partial similarity cancellation.

It becomes apparent that the best types of construction of sentences for GA are agent-object construction with odding blades and the Plate construction, as they ensure that sentences have an odd number of blades. Further, it is advisable to compute similarity by the means of Hamming measure or the Euclidean measure instead of the inner product. The Euclidean measure recognizes 100\% of items much faster (i.e. for smaller data size), but for large data size both measures behave identically. Therefore Hamming measure should be used to calculate similarity since its computation requires less time. The success of those measures is due to the fact that the differences between matrices or their signatures lessen the similarity, whereas differences in blades did not lessen the value of the inner product considerably.


\section{The Average Number of Potential Answers}
\label{sec:5}

Our point of interest in this Section will be to analyze the influence of accidental blade equality on the number of potential answers under agent-object construction with appropriate-hand-side reversed questions. The following estimates assume $ideal\ conditions$, i.e.$\ $no two chunks of a sentence are identical up to a constant at any time. Intuitively, such conditions could be met for sufficiently large lengths of the input vectors, whereas vectors of short length will with high probability be linearly dependent. We will estimate the average number of times that a nonzero inner product comes up when a noisy output is compared with items stored in the clean-up memory. We will deal with the following issues:
\begin{itemize}
\item[] How often does the system produce identical blades representing atomic objects? 
\item[] How often does the system produce identical sentence chunks from different blades?
\item[] How do the two above problems affect the number of potential answers?
\end{itemize}

Let $V$ be the the set of multivectors over $\mathbb{R}^N$ stored in the clean-up memory and let $\omega(V)$ be the maximum number of blades stored in a multivector in $V$. The set of all multivectors having the number of blades equal to $k$ is denoted by $S_k$ ($S_1$ being the set of atomic objects). Naturally, $V = S_1\cup\dots\cup S_{\omega(V)}$. Let $\tilde n$ be a noisy answer to some question. Under ideal conditions for every $c\in V$
\begin{equation}
|\langle \tilde n | c \rangle| \neq 0 \Leftrightarrow \tilde n \ \textrm{and} \ c \ \textrm{share\ a\ common\ blade}.
\end{equation}

$\\$We will begin with a simple example of a multivector with one meaningful blade and $L$ noisy blades. Let $r_0,\dots,r_L$ be roles and $f_0,\dots,f_L$ be fillers for some $L > 0$. Consider a question
\begin{equation}
(r_0 \ast f_0 + r_1 \ast f_1 + \dots + r_L \ast f_L)\ \sharp\ r_0
\end{equation}
which results in the following noisy answer
\begin{equation}
f_0 + \tilde n_1+\dots+\tilde n_L,\quad \tilde n_i = r_0^{+}\ast r_i \ast f_i,\quad 0<i\leq L.
\end{equation}
Surely, the original answer $f_0$ belongs to $S_1$. Let $s\in V$ be an arbitrary element of the clean-up memory.

\definecolor{mygray1}{rgb}{0.5,0.5,0.5}
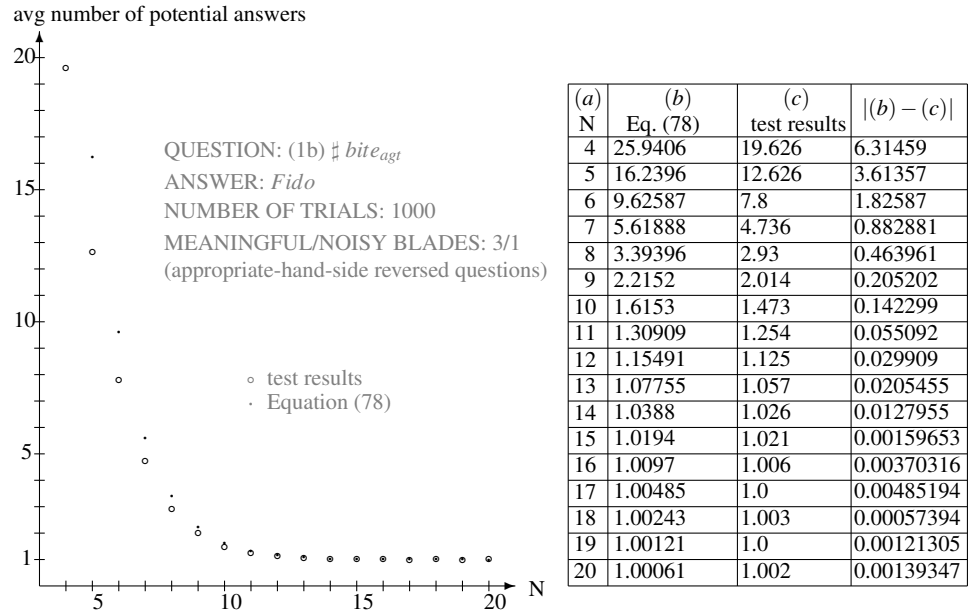
\begin{figure}[p]
  \begin{picture}(0,220)(-20,-5)
	 \put(0.0,0.0){\vector(1,0){180.0}}  
	 \put(0.0,0.0){\vector(0,1){210.0}} 
	 \put(10.0,2.0){\line(0,-1){4.0}} 
	 \put(20.0,2.0){\line(0,-1){4.0}} 
	 \put(30.0,2.0){\line(0,-1){4.0}} 
	 \put(40.0,2.0){\line(0,-1){4.0}} 
	 \put(50.0,2.0){\line(0,-1){4.0}} 
	 \put(60.0,2.0){\line(0,-1){4.0}} 
	 \put(70.0,2.0){\line(0,-1){4.0}} 
	 \put(80.0,2.0){\line(0,-1){4.0}} 
	 \put(90.0,2.0){\line(0,-1){4.0}} 
	 \put(100.0,2.0){\line(0,-1){4.0}} 
	 \put(110.0,2.0){\line(0,-1){4.0}} 
	 \put(120.0,2.0){\line(0,-1){4.0}} 
	 \put(130.0,2.0){\line(0,-1){4.0}} 
	 \put(140.0,2.0){\line(0,-1){4.0}} 
	 \put(150.0,2.0){\line(0,-1){4.0}} 
	 \put(160.0,2.0){\line(0,-1){4.0}} 
	 \put(170.0,2.0){\line(0,-1){4.0}} 
	 \put(-2.0,10.0){\line(1,0){4.0}} 
	 \put(-2.0,20.0){\line(1,0){4.0}} 
	 \put(-2.0,30.0){\line(1,0){4.0}} 
	 \put(-2.0,40.0){\line(1,0){4.0}} 
	 \put(-2.0,50.0){\line(1,0){4.0}} 
	 \put(-2.0,60.0){\line(1,0){4.0}} 
	 \put(-2.0,70.0){\line(1,0){4.0}} 
	 \put(-2.0,80.0){\line(1,0){4.0}} 
	 \put(-2.0,90.0){\line(1,0){4.0}} 
	 \put(-2.0,100.0){\line(1,0){4.0}} 
	 \put(-2.0,110.0){\line(1,0){4.0}} 
	 \put(-2.0,120.0){\line(1,0){4.0}} 
	 \put(-2.0,130.0){\line(1,0){4.0}} 
	 \put(-2.0,140.0){\line(1,0){4.0}} 
	 \put(-2.0,150.0){\line(1,0){4.0}} 
	 \put(-2.0,160.0){\line(1,0){4.0}}
	 \put(-2.0,170.0){\line(1,0){4.0}} 
	 \put(-2.0,180.0){\line(1,0){4.0}} 
	 \put(-2.0,190.0){\line(1,0){4.0}} 
	 \put(-2.0,200.0){\line(1,0){4.0}} 
	 \put(18.0,-9.0){\makebox(0,0)[bl]{ 5}}
	 \put(66.0,-9.0){\makebox(0,0)[bl]{ 10}}
	 \put(116.0,-9.0){\makebox(0,0)[bl]{ 15}}
	 \put(166.0,-9.0){\makebox(0,0)[bl]{ 20}}
	 \put(183.0,-4.0){\makebox(0,0)[bl]{ N}}
	 \put(-9,8.0){\makebox(0,0)[bl]{ 1}}
	 \put(-9,48.0){\makebox(0,0)[bl]{ 5}}
	 \put(-12,98.0){\makebox(0,0)[bl]{ 10}}
	 \put(-12,148.0){\makebox(0,0)[bl]{ 15}}
	 \put(-12,198.0){\makebox(0,0)[bl]{ 20}}
	 \put(-12.0,212.0){\makebox(0,0)[bl]{ avg number of potential answers}}
	 \put(10.0,196.26){\circle{1.5}}  
	 \put(20.0,126.26){\circle{1.5}}
	 \put(30.0,78){\circle{1.5}}
	 \put(40.0,47.36){\circle{1.5}}
	 \put(50.0,29.3){\circle{1.5}}
	 \put(60.0,20.14){\circle{1.5}}
	 \put(70.0,14.73){\circle{1.5}}
	 \put(80.0,12.54){\circle{1.5}}
	 \put(90.0,11.25){\circle{1.5}}
	 \put(100.0,10.57){\circle{1.5}}
	 \put(110.0,10.26){\circle{1.5}}
	 \put(120.0,10.21){\circle{1.5}}
	 \put(130.0,10.06){\circle{1.5}}
	 \put(140.0,10){\circle{1.5}}
	 \put(150.0,10.03){\circle{1.5}}
	 \put(160.0,10){\circle{1.5}}
	 \put(170.0,10.02){\circle{1.5}}
	 \put(20.0,162.396){\circle*{1}}
	 \put(30.0,96.2587){\circle*{1}}
	 \put(40.0,56.1888){\circle*{1}}
	 \put(50.0,33.9396){\circle*{1}}
	 \put(60.0,22.192){\circle*{1}}
	 \put(70.0,16.153){\circle*{1}}
	 \put(80.0,13.0909){\circle*{1}}
	 \put(90.0,11.5491){\circle*{1}}
	 \put(100.0,10.7755){\circle*{1}}
	 \put(110.0,10.388){\circle*{1}}
	 \put(120.0,10.194){\circle*{1}}
	 \put(130.0,10.097){\circle*{1}}
	 \put(140.0,10.0485){\circle*{1}}
	 \put(150.0,10.0243){\circle*{1}}
	 \put(160.0,10.0121){\circle*{1}}
	 \put(170.0,10.0061){\circle*{1}}
	 \color{mygray1}
	 \put(45.0,160.0){\makebox(0,0)[bl]{ QUESTION: (1b)$\ \sharp\ bite_{agt}$}}
	 \put(45.0,150.0){\makebox(0,0)[bl]{ ANSWER: $Fido$}}
	 \put(45.0,139.0){\makebox(0,0)[bl]{ NUMBER OF TRIALS: 1000}}
	 \put(45.0,127.0){\makebox(0,0)[bl]{ MEANINGFUL/NOISY BLADES: 3/1}}
	 \put(45.0,115.0){\makebox(0,0)[bl]{ (appropriate-hand-side reversed questions)}}	 
	 \put(80,78){\circle{1.5}}
	 \put(84,76){\makebox(0,0)[bl]{ test results}}
	 \put(80,69){\circle*{1}}
	 \put(84,65){\makebox(0,0)[bl]{ Equation (\ref{eq:simpleexamplecase12})}}
	 \color{black}
	 \put(200.0,190.0){\line(1,0){151.0}}  
	 \put(200.0,170.0){\line(1,0){151.0}}  
	 \put(200.0,160.0){\line(1,0){151.0}}  
	 \put(200.0,150.0){\line(1,0){151.0}}  
	 \put(200.0,140.0){\line(1,0){151.0}}  
	 \put(200.0,130.0){\line(1,0){151.0}}  
	 \put(200.0,120.0){\line(1,0){151.0}}  
	 \put(200.0,110.0){\line(1,0){151.0}}  
	 \put(200.0,100.0){\line(1,0){151.0}}  
	 \put(200.0,90.0){\line(1,0){151.0}}  
	 \put(200.0,80.0){\line(1,0){151.0}}  
	 \put(200.0,70.0){\line(1,0){151.0}}  
	 \put(200.0,60.0){\line(1,0){151.0}}  
	 \put(200.0,50.0){\line(1,0){151.0}}  
	 \put(200.0,40.0){\line(1,0){151.0}}  
	 \put(200.0,30.0){\line(1,0){151.0}}  
	 \put(200.0,20.0){\line(1,0){151.0}}  
	 \put(200.0,10.0){\line(1,0){151.0}}  
	 \put(200.0,0.0){\line(1,0){151.0}}  
	 \put(200.0,0.0){\line(0,1){190.0}}  
	 \put(215.0,0.0){\line(0,1){190.0}}  
	 \put(264.0,0.0){\line(0,1){190.0}}  
	 \put(307.0,0.0){\line(0,1){190.0}}  
	 \put(351.0,0.0){\line(0,1){190.0}}  	
	 \put(200,3.0){\makebox(0,0)[bl]{ 20}}
	 \put(200,13.0){\makebox(0,0)[bl]{ 19}}
	 \put(200,23.0){\makebox(0,0)[bl]{ 18}}
	 \put(200,33.0){\makebox(0,0)[bl]{ 17}}
	 \put(200,43.0){\makebox(0,0)[bl]{ 16}}
	 \put(200,53.0){\makebox(0,0)[bl]{ 15}}
	 \put(200,63.0){\makebox(0,0)[bl]{ 14}}
	 \put(200,73.0){\makebox(0,0)[bl]{ 13}}
	 \put(200,83.0){\makebox(0,0)[bl]{ 12}}
	 \put(200,93.0){\makebox(0,0)[bl]{ 11}}
	 \put(200,103.0){\makebox(0,0)[bl]{ 10}}
	 \put(204,113.0){\makebox(0,0)[bl]{ 9}}
	 \put(204,123.0){\makebox(0,0)[bl]{ 8}}
	 \put(204,133.0){\makebox(0,0)[bl]{ 7}}
	 \put(204,143.0){\makebox(0,0)[bl]{ 6}}
	 \put(204,153.0){\makebox(0,0)[bl]{ 5}}
	 \put(204,163.0){\makebox(0,0)[bl]{ 4}}
	 \put(202,172.0){\makebox(0,0)[bl]{ N}}
	 \put(200,180.0){\makebox(0,0)[bl]{ $(a)$}}
	 \put(222,170.5){\makebox(0,0)[bl]{{Eq. (\ref{eq:simpleexamplecase12})}}}
	 \put(234,180.0){\makebox(0,0)[bl]{ $(b)$}}
	 \put(215,3.0){\makebox(0,0)[bl]{ 1.00061}} 
	 \put(215,13.0){\makebox(0,0)[bl]{ 1.00121}}
	 \put(215,23.0){\makebox(0,0)[bl]{ 1.00243}}
	 \put(215,33.0){\makebox(0,0)[bl]{ 1.00485}}
	 \put(215,43.0){\makebox(0,0)[bl]{ 1.0097}}
	 \put(215,53.0){\makebox(0,0)[bl]{ 1.0194}}
	 \put(215,63.0){\makebox(0,0)[bl]{ 1.0388}}
	 \put(215,73.0){\makebox(0,0)[bl]{ 1.07755}}
	 \put(215,83.0){\makebox(0,0)[bl]{ 1.15491}}
	 \put(215,93.0){\makebox(0,0)[bl]{ 1.30909}}
	 \put(215,103.0){\makebox(0,0)[bl]{ 1.6153}}
	 \put(215,113.0){\makebox(0,0)[bl]{ 2.2152}}
	 \put(215,123.0){\makebox(0,0)[bl]{ 3.39396}}
	 \put(215,133.0){\makebox(0,0)[bl]{ 5.61888}}
	 \put(215,143.0){\makebox(0,0)[bl]{ 9.62587}}
	 \put(215,153.0){\makebox(0,0)[bl]{ 16.2396}}
	 \put(215,163.0){\makebox(0,0)[bl]{ 25.9406}}
	 \put(267,172.0){\makebox(0,0)[bl]{ test results}}
	 \put(279,180.0){\makebox(0,0)[bl]{ $(c)$}}
	 \put(263,3.0){\makebox(0,0)[bl]{ 1.002}} 
	 \put(263,13.0){\makebox(0,0)[bl]{ 1.0}}
	 \put(263,23.0){\makebox(0,0)[bl]{ 1.003}}
	 \put(263,33.0){\makebox(0,0)[bl]{ 1.0}}
	 \put(263,43.0){\makebox(0,0)[bl]{ 1.006}}
	 \put(263,53.0){\makebox(0,0)[bl]{ 1.021}}
	 \put(263,63.0){\makebox(0,0)[bl]{ 1.026}}
	 \put(263,73.0){\makebox(0,0)[bl]{ 1.057}}
	 \put(263,83.0){\makebox(0,0)[bl]{ 1.125}}
	 \put(263,93.0){\makebox(0,0)[bl]{ 1.254}}
	 \put(263,103.0){\makebox(0,0)[bl]{ 1.473}}
	 \put(263,113.0){\makebox(0,0)[bl]{ 2.014}}
	 \put(263,123.0){\makebox(0,0)[bl]{ 2.93}}
	 \put(263,133.0){\makebox(0,0)[bl]{ 4.736}}
	 \put(263,143.0){\makebox(0,0)[bl]{ 7.8}}
	 \put(263,153.0){\makebox(0,0)[bl]{ 12.626}}
	 \put(263,163.0){\makebox(0,0)[bl]{ 19.626}}
	 \put(309,176.0){\makebox(0,0)[bl]{ $|(b)-(c)|$}}
	 \put(306,3.0){\makebox(0,0)[bl]{ 0.00139347}} 
	 \put(306,13.0){\makebox(0,0)[bl]{ 0.00121305}}
	 \put(306,23.0){\makebox(0,0)[bl]{ 0.00057394}}
	 \put(306,33.0){\makebox(0,0)[bl]{ 0.00485194}}
	 \put(306,43.0){\makebox(0,0)[bl]{ 0.00370316}}
	 \put(306,53.0){\makebox(0,0)[bl]{ 0.00159653}}
	 \put(306,63.0){\makebox(0,0)[bl]{ 0.0127955}}
	 \put(306,73.0){\makebox(0,0)[bl]{ 0.0205455}}
	 \put(306,83.0){\makebox(0,0)[bl]{ 0.029909}}
	 \put(306,93.0){\makebox(0,0)[bl]{ 0.055092}}
	 \put(306,103.0){\makebox(0,0)[bl]{ 0.142299}}
	 \put(306,113.0){\makebox(0,0)[bl]{ 0.205202}}	 
	 \put(306,123.0){\makebox(0,0)[bl]{ 0.463961}}	 
	 \put(306,133.0){\makebox(0,0)[bl]{ 0.882881}}	 
	 \put(306,143.0){\makebox(0,0)[bl]{ 1.82587}}	 
	 \put(306,153.0){\makebox(0,0)[bl]{ 3.61357}}	 
	 \put(306,163.0){\makebox(0,0)[bl]{ 6.31459}}	 
  \end{picture}
  \caption{Average number of potential answers per 1000 trials with a 1:3 meaningful-to-noisy blades ratio.}
  \label{fig:simpleexamplecase13}   
\end{figure}

\definecolor{mygray1}{rgb}{0.5,0.5,0.5}
\begin{figure}[p]
  \begin{picture}(0,220)(-20,-5)
	 \put(0.0,0.0){\vector(1,0){180.0}}  
	 \put(0.0,0.0){\vector(0,1){210.0}} 
	 \put(10.0,2.0){\line(0,-1){4.0}} 
	 \put(20.0,2.0){\line(0,-1){4.0}} 
	 \put(30.0,2.0){\line(0,-1){4.0}} 
	 \put(40.0,2.0){\line(0,-1){4.0}} 
	 \put(50.0,2.0){\line(0,-1){4.0}} 
	 \put(60.0,2.0){\line(0,-1){4.0}} 
	 \put(70.0,2.0){\line(0,-1){4.0}} 
	 \put(80.0,2.0){\line(0,-1){4.0}} 
	 \put(90.0,2.0){\line(0,-1){4.0}} 
	 \put(100.0,2.0){\line(0,-1){4.0}} 
	 \put(110.0,2.0){\line(0,-1){4.0}} 
	 \put(120.0,2.0){\line(0,-1){4.0}} 
	 \put(130.0,2.0){\line(0,-1){4.0}} 
	 \put(140.0,2.0){\line(0,-1){4.0}} 
	 \put(150.0,2.0){\line(0,-1){4.0}} 
	 \put(160.0,2.0){\line(0,-1){4.0}} 
	 \put(170.0,2.0){\line(0,-1){4.0}} 
	 \put(-2.0,10.0){\line(1,0){4.0}} 
	 \put(-2.0,20.0){\line(1,0){4.0}} 
	 \put(-2.0,30.0){\line(1,0){4.0}} 
	 \put(-2.0,40.0){\line(1,0){4.0}} 
	 \put(-2.0,50.0){\line(1,0){4.0}} 
	 \put(-2.0,60.0){\line(1,0){4.0}} 
	 \put(-2.0,70.0){\line(1,0){4.0}} 
	 \put(-2.0,80.0){\line(1,0){4.0}} 
	 \put(-2.0,90.0){\line(1,0){4.0}} 
	 \put(-2.0,100.0){\line(1,0){4.0}} 
	 \put(-2.0,110.0){\line(1,0){4.0}} 
	 \put(-2.0,120.0){\line(1,0){4.0}} 
	 \put(-2.0,130.0){\line(1,0){4.0}} 
	 \put(-2.0,140.0){\line(1,0){4.0}} 
	 \put(-2.0,150.0){\line(1,0){4.0}} 
	 \put(-2.0,160.0){\line(1,0){4.0}}
	 \put(-2.0,170.0){\line(1,0){4.0}} 
	 \put(-2.0,180.0){\line(1,0){4.0}} 
	 \put(-2.0,190.0){\line(1,0){4.0}} 
	 \put(-2.0,200.0){\line(1,0){4.0}} 
	 \put(18.0,-9.0){\makebox(0,0)[bl]{ 5}}
	 \put(66.0,-9.0){\makebox(0,0)[bl]{ 10}}
	 \put(116.0,-9.0){\makebox(0,0)[bl]{ 15}}
	 \put(166.0,-9.0){\makebox(0,0)[bl]{ 20}}
	 \put(183.0,-4.0){\makebox(0,0)[bl]{ N}}
	 \put(-9,8.0){\makebox(0,0)[bl]{ 1}}
	 \put(-9,48.0){\makebox(0,0)[bl]{ 5}}
	 \put(-12,98.0){\makebox(0,0)[bl]{ 10}}
	 \put(-12,148.0){\makebox(0,0)[bl]{ 15}}
	 \put(-12,198.0){\makebox(0,0)[bl]{ 20}}
	 \put(-12.0,212.0){\makebox(0,0)[bl]{ avg number of potential answers}}
	 \put(10.0,194.79){\circle{1.5}}  
	 \put(20.0,126.98){\circle{1.5}}
	 \put(30.0,81.75){\circle{1.5}}
	 \put(40.0,53.68){\circle{1.5}}
	 \put(50.0,37.85){\circle{1.5}}
	 \put(60.0,29.01){\circle{1.5}}
	 \put(70.0,24.41){\circle{1.5}}
	 \put(80.0,22.49){\circle{1.5}}
	 \put(90.0,21.36){\circle{1.5}}
	 \put(100.0,20.45){\circle{1.5}}
	 \put(110.0,20.32){\circle{1.5}}
	 \put(120.0,20.21){\circle{1.5}}
	 \put(130.0,20.15){\circle{1.5}}
	 \put(140.0,20.04){\circle{1.5}}
	 \put(150.0,20.03){\circle{1.5}}
	 \put(160.0,20){\circle{1.5}}
	 \put(170.0,20){\circle{1.5}}
	 \put(20.0,164.272){\circle*{1}}
	 \put(30.0,101.488){\circle*{1}}
	 \put(40.0,63.6){\circle*{1}}
	 \put(50.0,42.5906){\circle*{1}}
	 \put(60.0,31.5034){\circle*{1}}
	 \put(70.0,25.8051){\circle*{1}}
	 \put(80.0,22.9161){\circle*{1}}
	 \put(90.0,21.4614){\circle*{1}}
	 \put(100.0,20.7316){\circle*{1}}
	 \put(110.0,20.366){\circle*{1}}
	 \put(120.0,20.1831){\circle*{1}}
	 \put(130.0,20.0915){\circle*{1}}
	 \put(140.0,20.0458){\circle*{1}}
	 \put(150.0,20.0229){\circle*{1}}
	 \put(160.0,20.0114){\circle*{1}}
	 \put(170.0,20.0057){\circle*{1}}
	 \color{mygray1}
	 \put(45.0,160.0){\makebox(0,0)[bl]{ QUESTION: (1b)$\ \sharp\ bite_{obj}$}}
	 \put(45.0,150.0){\makebox(0,0)[bl]{ ANSWER: $PSmith$}}
	 \put(45.0,139.0){\makebox(0,0)[bl]{ NUMBER OF TRIALS: 1000}}
	 \put(45.0,127.0){\makebox(0,0)[bl]{ MEANINGFUL/NOISY BLADES: 3/1}}
	 \put(45.0,115.0){\makebox(0,0)[bl]{ (appropriate-hand-side reversed questions)}}	 
	 \put(80,78){\circle{1.5}}
	 \put(84,76){\makebox(0,0)[bl]{ test results}}
	 \put(80,69){\circle*{1}}
	 \put(84,65){\makebox(0,0)[bl]{ Equation (\ref{eq:complexexamplecase31})}}
	 \color{black}
	 \put(200.0,190.0){\line(1,0){151.0}}  
	 \put(200.0,170.0){\line(1,0){151.0}}  
	 \put(200.0,160.0){\line(1,0){151.0}}  
	 \put(200.0,150.0){\line(1,0){151.0}}  
	 \put(200.0,140.0){\line(1,0){151.0}}  
	 \put(200.0,130.0){\line(1,0){151.0}}  
	 \put(200.0,120.0){\line(1,0){151.0}}  
	 \put(200.0,110.0){\line(1,0){151.0}}  
	 \put(200.0,100.0){\line(1,0){151.0}}  
	 \put(200.0,90.0){\line(1,0){151.0}}  
	 \put(200.0,80.0){\line(1,0){151.0}}  
	 \put(200.0,70.0){\line(1,0){151.0}}  
	 \put(200.0,60.0){\line(1,0){151.0}}  
	 \put(200.0,50.0){\line(1,0){151.0}}  
	 \put(200.0,40.0){\line(1,0){151.0}}  
	 \put(200.0,30.0){\line(1,0){151.0}}  
	 \put(200.0,20.0){\line(1,0){151.0}}  
	 \put(200.0,10.0){\line(1,0){151.0}}  
	 \put(200.0,0.0){\line(1,0){151.0}}  
	 \put(200.0,0.0){\line(0,1){190.0}}  
	 \put(215.0,0.0){\line(0,1){190.0}}  
	 \put(264.0,0.0){\line(0,1){190.0}}  
	 \put(307.0,0.0){\line(0,1){190.0}}  
	 \put(351.0,0.0){\line(0,1){190.0}}  
	 \put(200,3.0){\makebox(0,0)[bl]{ 20}}
	 \put(200,13.0){\makebox(0,0)[bl]{ 19}}
	 \put(200,23.0){\makebox(0,0)[bl]{ 18}}
	 \put(200,33.0){\makebox(0,0)[bl]{ 17}}
	 \put(200,43.0){\makebox(0,0)[bl]{ 16}}
	 \put(200,53.0){\makebox(0,0)[bl]{ 15}}
	 \put(200,63.0){\makebox(0,0)[bl]{ 14}}
	 \put(200,73.0){\makebox(0,0)[bl]{ 13}}
	 \put(200,83.0){\makebox(0,0)[bl]{ 12}}
	 \put(200,93.0){\makebox(0,0)[bl]{ 11}}
	 \put(200,103.0){\makebox(0,0)[bl]{ 10}}
	 \put(204,113.0){\makebox(0,0)[bl]{ 9}}
	 \put(204,123.0){\makebox(0,0)[bl]{ 8}}
	 \put(204,133.0){\makebox(0,0)[bl]{ 7}}
	 \put(204,143.0){\makebox(0,0)[bl]{ 6}}
	 \put(204,153.0){\makebox(0,0)[bl]{ 5}}
	 \put(204,163.0){\makebox(0,0)[bl]{ 4}}
	 \put(202,172.0){\makebox(0,0)[bl]{ N}}
	 \put(200,180.0){\makebox(0,0)[bl]{ $(a)$}}
	 \put(222,170.5){\makebox(0,0)[bl]{{Eq. (\ref{eq:complexexamplecase31})}}}
	 \put(234,180.0){\makebox(0,0)[bl]{ $(b)$}}
	 \put(215,3.0){\makebox(0,0)[bl]{ 2.00057}} 
	 \put(215,13.0){\makebox(0,0)[bl]{ 2.00114}}
	 \put(215,23.0){\makebox(0,0)[bl]{ 2.00229}}
	 \put(215,33.0){\makebox(0,0)[bl]{ 2.00458}}
	 \put(215,43.0){\makebox(0,0)[bl]{ 2.00915}}
	 \put(215,53.0){\makebox(0,0)[bl]{ 2.01831}}
	 \put(215,63.0){\makebox(0,0)[bl]{ 2.0366}}
	 \put(215,73.0){\makebox(0,0)[bl]{ 2.07316}}
	 \put(215,83.0){\makebox(0,0)[bl]{ 2.14614}}
	 \put(215,93.0){\makebox(0,0)[bl]{ 2.29161}}
	 \put(215,103.0){\makebox(0,0)[bl]{ 2.58051}}
	 \put(215,113.0){\makebox(0,0)[bl]{ 3.15034}}
	 \put(215,123.0){\makebox(0,0)[bl]{ 4.25906}}
	 \put(215,133.0){\makebox(0,0)[bl]{ 6.36}}
	 \put(215,143.0){\makebox(0,0)[bl]{ 10.1488}}
	 \put(215,153.0){\makebox(0,0)[bl]{ 16.4272}}
	 \put(215,163.0){\makebox(0,0)[bl]{ 25.7459}}
	 \put(267,172.0){\makebox(0,0)[bl]{ test results}}
	 \put(279,180.0){\makebox(0,0)[bl]{ $(c)$}}
	 \put(263,3.0){\makebox(0,0)[bl]{ 2.0}} 
	 \put(263,13.0){\makebox(0,0)[bl]{ 2.0}}
	 \put(263,23.0){\makebox(0,0)[bl]{ 2.003}}
	 \put(263,33.0){\makebox(0,0)[bl]{ 2.004}}
	 \put(263,43.0){\makebox(0,0)[bl]{ 2.015}}
	 \put(263,53.0){\makebox(0,0)[bl]{ 2.021}}
	 \put(263,63.0){\makebox(0,0)[bl]{ 2.032}}
	 \put(263,73.0){\makebox(0,0)[bl]{ 2.045}}
	 \put(263,83.0){\makebox(0,0)[bl]{ 2.136}}
	 \put(263,93.0){\makebox(0,0)[bl]{ 2.249}}
	 \put(263,103.0){\makebox(0,0)[bl]{ 2.441}}
	 \put(263,113.0){\makebox(0,0)[bl]{ 2.901}}
	 \put(263,123.0){\makebox(0,0)[bl]{ 3.785}}
	 \put(263,133.0){\makebox(0,0)[bl]{ 5.368}}
	 \put(263,143.0){\makebox(0,0)[bl]{ 8.175}}
	 \put(263,153.0){\makebox(0,0)[bl]{ 12.698}}
	 \put(263,163.0){\makebox(0,0)[bl]{ 19.479}}
	 \put(309,176.0){\makebox(0,0)[bl]{ $|(b)-(c)|$}}
	 \put(306,3.0){\makebox(0,0)[bl]{ 0.00057219}} 
	 \put(306,13.0){\makebox(0,0)[bl]{ 0.00114439}}
	 \put(306,23.0){\makebox(0,0)[bl]{ 0.00071126}}
	 \put(306,33.0){\makebox(0,0)[bl]{ 0.00057730}}
	 \put(306,43.0){\makebox(0,0)[bl]{ 0.00584606}}
	 \put(306,53.0){\makebox(0,0)[bl]{ 0.0026948}}
	 \put(306,63.0){\makebox(0,0)[bl]{ 0.00459971}}
	 \put(306,73.0){\makebox(0,0)[bl]{ 0.0281567}}
	 \put(306,83.0){\makebox(0,0)[bl]{ 0.0101427}}
	 \put(306,93.0){\makebox(0,0)[bl]{ 0.0426052}}
	 \put(306,103.0){\makebox(0,0)[bl]{ 0.139507}}
	 \put(306,113.0){\makebox(0,0)[bl]{ 0.249337}}	 
	 \put(306,123.0){\makebox(0,0)[bl]{ 0.47406}}	 
	 \put(306,133.0){\makebox(0,0)[bl]{ 0.992003}}	 
	 \put(306,143.0){\makebox(0,0)[bl]{ 1.97385}}	 
	 \put(306,153.0){\makebox(0,0)[bl]{ 3.72919}}	 
	 \put(306,163.0){\makebox(0,0)[bl]{ 6.26695}}	 
  \end{picture}
  \caption{Average number of potential answers per 1000 trials with a 3:1 meaningful-to-noisy blades ratio.}
  \label{fig:complexexamplecase31}   
\end{figure}

\definecolor{mygray1}{rgb}{0.5,0.5,0.5}
\begin{figure}[p]
  \begin{picture}(0,220)(-20,-5)
	 \put(0.0,0.0){\vector(1,0){180.0}}  
	 \put(0.0,0.0){\vector(0,1){210.0}} 
	 \put(10.0,2.0){\line(0,-1){4.0}} 
	 \put(20.0,2.0){\line(0,-1){4.0}} 
	 \put(30.0,2.0){\line(0,-1){4.0}} 
	 \put(40.0,2.0){\line(0,-1){4.0}} 
	 \put(50.0,2.0){\line(0,-1){4.0}} 
	 \put(60.0,2.0){\line(0,-1){4.0}} 
	 \put(70.0,2.0){\line(0,-1){4.0}} 
	 \put(80.0,2.0){\line(0,-1){4.0}} 
	 \put(90.0,2.0){\line(0,-1){4.0}} 
	 \put(100.0,2.0){\line(0,-1){4.0}} 
	 \put(110.0,2.0){\line(0,-1){4.0}} 
	 \put(120.0,2.0){\line(0,-1){4.0}} 
	 \put(130.0,2.0){\line(0,-1){4.0}} 
	 \put(140.0,2.0){\line(0,-1){4.0}} 
	 \put(150.0,2.0){\line(0,-1){4.0}} 
	 \put(160.0,2.0){\line(0,-1){4.0}} 
	 \put(170.0,2.0){\line(0,-1){4.0}} 
	 \put(-2.0,10.0){\line(1,0){4.0}} 
	 \put(-2.0,20.0){\line(1,0){4.0}} 
	 \put(-2.0,30.0){\line(1,0){4.0}} 
	 \put(-2.0,40.0){\line(1,0){4.0}} 
	 \put(-2.0,50.0){\line(1,0){4.0}} 
	 \put(-2.0,60.0){\line(1,0){4.0}} 
	 \put(-2.0,70.0){\line(1,0){4.0}} 
	 \put(-2.0,80.0){\line(1,0){4.0}} 
	 \put(-2.0,90.0){\line(1,0){4.0}} 
	 \put(-2.0,100.0){\line(1,0){4.0}} 
	 \put(-2.0,110.0){\line(1,0){4.0}} 
	 \put(-2.0,120.0){\line(1,0){4.0}} 
	 \put(-2.0,130.0){\line(1,0){4.0}} 
	 \put(-2.0,140.0){\line(1,0){4.0}} 
	 \put(-2.0,150.0){\line(1,0){4.0}} 
	 \put(-2.0,160.0){\line(1,0){4.0}}
	 \put(-2.0,170.0){\line(1,0){4.0}} 
	 \put(-2.0,180.0){\line(1,0){4.0}} 
	 \put(-2.0,190.0){\line(1,0){4.0}} 
	 \put(-2.0,200.0){\line(1,0){4.0}} 
	 \put(18.0,-9.0){\makebox(0,0)[bl]{ 5}}
	 \put(66.0,-9.0){\makebox(0,0)[bl]{ 10}}
	 \put(116.0,-9.0){\makebox(0,0)[bl]{ 15}}
	 \put(166.0,-9.0){\makebox(0,0)[bl]{ 20}}
	 \put(183.0,-4.0){\makebox(0,0)[bl]{ N}}
	 \put(-9,8.0){\makebox(0,0)[bl]{ 1}}
	 \put(-9,48.0){\makebox(0,0)[bl]{ 5}}
	 \put(-12,98.0){\makebox(0,0)[bl]{ 10}}
	 \put(-12,148.0){\makebox(0,0)[bl]{ 15}}
	 \put(-12,198.0){\makebox(0,0)[bl]{ 20}}
	 \put(-12.0,212.0){\makebox(0,0)[bl]{ avg number of potential answers}}
	 \put(20.0,194.32){\circle{1.5}}
	 \put(30.0,136.5){\circle{1.5}}
	 \put(40.0,95.13){\circle{1.5}}
	 \put(50.0,66.01){\circle{1.5}}
	 \put(60.0,49.24){\circle{1.5}}
	 \put(70.0,40.18){\circle{1.5}}
	 \put(80.0,34.88){\circle{1.5}}
	 \put(90.0,32.46){\circle{1.5}}
	 \put(100.0,31.42){\circle{1.5}}
	 \put(110.0,30.6){\circle{1.5}}
	 \put(120.0,30.38){\circle{1.5}}
	 \put(130.0,30.17){\circle{1.5}}
	 \put(140.0,30.05){\circle{1.5}}
	 \put(150.0,30.03){\circle{1.5}}
	 \put(160.0,30){\circle{1.5}}
	 \put(170.0,30.01){\circle{1.5}}
	 \put(20.0,170.6){\circle*{1}}
	 \put(30.0,109.365){\circle*{1}}
	 \put(40.0,72.4505){\circle*{1}}
	 \put(50.0,51.9916){\circle*{1}}
	 \put(60.0,41.1975){\circle*{1}}
	 \put(70.0,35.6505){\circle*{1}}
	 \put(80.0,32.8383){\circle*{1}}
	 \put(90.0,31.4225){\circle*{1}}
	 \put(100.0,30.7121){\circle*{1}}
	 \put(110.0,30.3562){\circle*{1}}
	 \put(120.0,30.1782){\circle*{1}}
	 \put(130.0,30.0891){\circle*{1}}
	 \put(140.0,30.0446){\circle*{1}}
	 \put(150.0,30.0223){\circle*{1}}
	 \put(160.0,30.0111){\circle*{1}}
	 \put(170.0,30.0056){\circle*{1}}
	 \color{mygray1}
	 \put(45.0,160.0){\makebox(0,0)[bl]{ QUESTION: (4a)$\ \sharp\ cause_{obj}$}}
	 \put(45.0,149.0){\makebox(0,0)[bl]{ ANSWER: $flee_{agt}\ast Pat+flee_{obj}\ast Fido$}}
	 \put(45.0,139.0){\makebox(0,0)[bl]{ NUMBER OF TRIALS: 1000}}
	 \put(45.0,127.0){\makebox(0,0)[bl]{ MEANINGFUL/NOISY BLADES: 2/2}}
	 \put(45.0,115.0){\makebox(0,0)[bl]{ (appropriate-hand-side reversed questions)}}
	 \put(80,78){\circle{1.5}}
	 \put(84,76){\makebox(0,0)[bl]{ test results}}
	 \put(80,69){\circle*{1}}
	 \put(84,65){\makebox(0,0)[bl]{ Equation (\ref{eq:complexexamplecase22})}}
	 \color{black}
	 \put(200.0,190.0){\line(1,0){151.0}}  
	 \put(200.0,170.0){\line(1,0){151.0}}  
	 \put(200.0,160.0){\line(1,0){151.0}}  
	 \put(200.0,150.0){\line(1,0){151.0}}  
	 \put(200.0,140.0){\line(1,0){151.0}}  
	 \put(200.0,130.0){\line(1,0){151.0}}  
	 \put(200.0,120.0){\line(1,0){151.0}}  
	 \put(200.0,110.0){\line(1,0){151.0}}  
	 \put(200.0,100.0){\line(1,0){151.0}}  
	 \put(200.0,90.0){\line(1,0){151.0}}  
	 \put(200.0,80.0){\line(1,0){151.0}}  
	 \put(200.0,70.0){\line(1,0){151.0}}  
	 \put(200.0,60.0){\line(1,0){151.0}}  
	 \put(200.0,50.0){\line(1,0){151.0}}  
	 \put(200.0,40.0){\line(1,0){151.0}}  
	 \put(200.0,30.0){\line(1,0){151.0}}  
	 \put(200.0,20.0){\line(1,0){151.0}}  
	 \put(200.0,10.0){\line(1,0){151.0}}  
	 \put(200.0,0.0){\line(1,0){151.0}}  
	 \put(200.0,0.0){\line(0,1){190.0}}  
	 \put(215.0,0.0){\line(0,1){190.0}}  
	 \put(264.0,0.0){\line(0,1){190.0}}  
	 \put(307.0,0.0){\line(0,1){190.0}}  
	 \put(351.0,0.0){\line(0,1){190.0}}  
	 \put(200,3.0){\makebox(0,0)[bl]{ 20}}
	 \put(200,13.0){\makebox(0,0)[bl]{ 19}}
	 \put(200,23.0){\makebox(0,0)[bl]{ 18}}
	 \put(200,33.0){\makebox(0,0)[bl]{ 17}}
	 \put(200,43.0){\makebox(0,0)[bl]{ 16}}
	 \put(200,53.0){\makebox(0,0)[bl]{ 15}}
	 \put(200,63.0){\makebox(0,0)[bl]{ 14}}
	 \put(200,73.0){\makebox(0,0)[bl]{ 13}}
	 \put(200,83.0){\makebox(0,0)[bl]{ 12}}
	 \put(200,93.0){\makebox(0,0)[bl]{ 11}}
	 \put(200,103.0){\makebox(0,0)[bl]{ 10}}
	 \put(204,113.0){\makebox(0,0)[bl]{ 9}}
	 \put(204,123.0){\makebox(0,0)[bl]{ 8}}
	 \put(204,133.0){\makebox(0,0)[bl]{ 7}}
	 \put(204,143.0){\makebox(0,0)[bl]{ 6}}
	 \put(204,153.0){\makebox(0,0)[bl]{ 5}}
	 \put(204,163.0){\makebox(0,0)[bl]{ 4}}
	 \put(202,172.0){\makebox(0,0)[bl]{ N}}
	 \put(200,180.0){\makebox(0,0)[bl]{ $(a)$}}
	 \put(222,170.5){\makebox(0,0)[bl]{{Eq. (\ref{eq:complexexamplecase22})}}}
	 \put(234,180.0){\makebox(0,0)[bl]{ $(b)$}}
	 \put(215,3.0){\makebox(0,0)[bl]{ 3.00056}} 
	 \put(215,13.0){\makebox(0,0)[bl]{ 3.00111}}
	 \put(215,23.0){\makebox(0,0)[bl]{ 3.00223}}
	 \put(215,33.0){\makebox(0,0)[bl]{ 3.00446}}
	 \put(215,43.0){\makebox(0,0)[bl]{ 3.00891}}
	 \put(215,53.0){\makebox(0,0)[bl]{ 3.01782}}
	 \put(215,63.0){\makebox(0,0)[bl]{ 3.03562}}
	 \put(215,73.0){\makebox(0,0)[bl]{ 3.07121}}
	 \put(215,83.0){\makebox(0,0)[bl]{ 3.14225}}
	 \put(215,93.0){\makebox(0,0)[bl]{ 3.28383}}
	 \put(215,103.0){\makebox(0,0)[bl]{ 3.56505}}
	 \put(215,113.0){\makebox(0,0)[bl]{ 4.11975}}
	 \put(215,123.0){\makebox(0,0)[bl]{ 5.19916}}
	 \put(215,133.0){\makebox(0,0)[bl]{ 7.24505}}
	 \put(215,143.0){\makebox(0,0)[bl]{ 10.9365}}
	 \put(215,153.0){\makebox(0,0)[bl]{ 17.06}}
	 \put(215,163.0){\makebox(0,0)[bl]{ 26.1696}}
	 \put(267,172.0){\makebox(0,0)[bl]{ test results}}
	 \put(279,180.0){\makebox(0,0)[bl]{ $(c)$}}
	 \put(263,3.0){\makebox(0,0)[bl]{ 3.001}} 
	 \put(263,13.0){\makebox(0,0)[bl]{ 3.0}}
	 \put(263,23.0){\makebox(0,0)[bl]{ 3.003}}
	 \put(263,33.0){\makebox(0,0)[bl]{ 3.005}}
	 \put(263,43.0){\makebox(0,0)[bl]{ 3.017}}
	 \put(263,53.0){\makebox(0,0)[bl]{ 3.038}}
	 \put(263,63.0){\makebox(0,0)[bl]{ 3.06}}
	 \put(263,73.0){\makebox(0,0)[bl]{ 3.142}}
	 \put(263,83.0){\makebox(0,0)[bl]{ 3.246}}
	 \put(263,93.0){\makebox(0,0)[bl]{ 3.488}}
	 \put(263,103.0){\makebox(0,0)[bl]{ 4.018}}
	 \put(263,113.0){\makebox(0,0)[bl]{ 4.924}}
	 \put(263,123.0){\makebox(0,0)[bl]{ 6.601}}
	 \put(263,133.0){\makebox(0,0)[bl]{ 9.513}}
	 \put(263,143.0){\makebox(0,0)[bl]{ 13.65}}
	 \put(263,153.0){\makebox(0,0)[bl]{ 19.432}}
	 \put(263,163.0){\makebox(0,0)[bl]{ 26.908}}
	 \put(309,176.0){\makebox(0,0)[bl]{ $|(b)-(c)|$}}
	 \put(306,3.0){\makebox(0,0)[bl]{ 0.00044306}} 
	 \put(306,13.0){\makebox(0,0)[bl]{ 0.00111387}}
	 \put(306,23.0){\makebox(0,0)[bl]{ 0.00077229}}
	 \put(306,33.0){\makebox(0,0)[bl]{ 0.00054476}}
	 \put(306,43.0){\makebox(0,0)[bl]{ 0.00809016}}
	 \put(306,53.0){\makebox(0,0)[bl]{ 0.0201829}}
	 \put(306,63.0){\makebox(0,0)[bl]{ 0.0243762}}
	 \put(306,73.0){\makebox(0,0)[bl]{ 0.0707938}}
	 \put(306,83.0){\makebox(0,0)[bl]{ 0.103753}}
	 \put(306,93.0){\makebox(0,0)[bl]{ 0.204166}}
	 \put(306,103.0){\makebox(0,0)[bl]{ 0.452952}}
	 \put(306,113.0){\makebox(0,0)[bl]{ 0.804253}}	 
	 \put(306,123.0){\makebox(0,0)[bl]{ 1.40184}}	 
	 \put(306,133.0){\makebox(0,0)[bl]{ 2.26795}}	 
	 \put(306,143.0){\makebox(0,0)[bl]{ 2.71353}}	 
	 \put(306,153.0){\makebox(0,0)[bl]{ 2.37202}}	 
	 \put(306,163.0){\makebox(0,0)[bl]{ 0.738392}}	 
  \end{picture}
  \caption{Average number of potential answers per 1000 trials with a 2:2 meaningful-to-noisy blades ratio.}
  \label{fig:complexexamplecase22a}   
\end{figure}

\definecolor{mygray1}{rgb}{0.5,0.5,0.5}
\begin{figure}[p]
  \begin{picture}(0,220)(-20,-5)
	 \put(0.0,0.0){\vector(1,0){180.0}}  
	 \put(0.0,0.0){\vector(0,1){210.0}} 
	 \put(10.0,2.0){\line(0,-1){4.0}} 
	 \put(20.0,2.0){\line(0,-1){4.0}} 
	 \put(30.0,2.0){\line(0,-1){4.0}} 
	 \put(40.0,2.0){\line(0,-1){4.0}} 
	 \put(50.0,2.0){\line(0,-1){4.0}} 
	 \put(60.0,2.0){\line(0,-1){4.0}} 
	 \put(70.0,2.0){\line(0,-1){4.0}} 
	 \put(80.0,2.0){\line(0,-1){4.0}} 
	 \put(90.0,2.0){\line(0,-1){4.0}} 
	 \put(100.0,2.0){\line(0,-1){4.0}} 
	 \put(110.0,2.0){\line(0,-1){4.0}} 
	 \put(120.0,2.0){\line(0,-1){4.0}} 
	 \put(130.0,2.0){\line(0,-1){4.0}} 
	 \put(140.0,2.0){\line(0,-1){4.0}} 
	 \put(150.0,2.0){\line(0,-1){4.0}} 
	 \put(160.0,2.0){\line(0,-1){4.0}} 
	 \put(170.0,2.0){\line(0,-1){4.0}} 
	 \put(-2.0,10.0){\line(1,0){4.0}} 
	 \put(-2.0,20.0){\line(1,0){4.0}} 
	 \put(-2.0,30.0){\line(1,0){4.0}} 
	 \put(-2.0,40.0){\line(1,0){4.0}} 
	 \put(-2.0,50.0){\line(1,0){4.0}} 
	 \put(-2.0,60.0){\line(1,0){4.0}} 
	 \put(-2.0,70.0){\line(1,0){4.0}} 
	 \put(-2.0,80.0){\line(1,0){4.0}} 
	 \put(-2.0,90.0){\line(1,0){4.0}} 
	 \put(-2.0,100.0){\line(1,0){4.0}} 
	 \put(-2.0,110.0){\line(1,0){4.0}} 
	 \put(-2.0,120.0){\line(1,0){4.0}} 
	 \put(-2.0,130.0){\line(1,0){4.0}} 
	 \put(-2.0,140.0){\line(1,0){4.0}} 
	 \put(-2.0,150.0){\line(1,0){4.0}} 
	 \put(-2.0,160.0){\line(1,0){4.0}}
	 \put(-2.0,170.0){\line(1,0){4.0}} 
	 \put(-2.0,180.0){\line(1,0){4.0}} 
	 \put(-2.0,190.0){\line(1,0){4.0}} 
	 \put(-2.0,200.0){\line(1,0){4.0}} 
	 \put(18.0,-9.0){\makebox(0,0)[bl]{ 5}}
	 \put(66.0,-9.0){\makebox(0,0)[bl]{ 10}}
	 \put(116.0,-9.0){\makebox(0,0)[bl]{ 15}}
	 \put(166.0,-9.0){\makebox(0,0)[bl]{ 20}}
	 \put(183.0,-4.0){\makebox(0,0)[bl]{ N}}
	 \put(-9,8.0){\makebox(0,0)[bl]{ 1}}
	 \put(-9,48.0){\makebox(0,0)[bl]{ 5}}
	 \put(-12,98.0){\makebox(0,0)[bl]{ 10}}
	 \put(-12,148.0){\makebox(0,0)[bl]{ 15}}
	 \put(-12,198.0){\makebox(0,0)[bl]{ 20}}
	 \put(-12.0,212.0){\makebox(0,0)[bl]{ avg number of potential answers}}
	 \put(10.0,198.15){\circle{1.5}}  
	 \put(20.0,134.2){\circle{1.5}}
	 \put(30.0,88.54){\circle{1.5}}
	 \put(40.0,58.54){\circle{1.5}}
	 \put(50.0,42.65){\circle{1.5}}
	 \put(60.0,34.11){\circle{1.5}}
	 \put(70.0,29.14){\circle{1.5}}
	 \put(80.0,27.27){\circle{1.5}}
	 \put(90.0,26.45){\circle{1.5}}
	 \put(100.0,25.56){\circle{1.5}}
	 \put(110.0,25.43){\circle{1.5}}
	 \put(120.0,25.13){\circle{1.5}}
	 \put(130.0,25.32){\circle{1.5}}
	 \put(140.0,25.08){\circle{1.5}}
	 \put(150.0,24.95){\circle{1.5}}
	 \put(160.0,25.25){\circle{1.5}}
	 \put(170.0,24.96){\circle{1.5}}
	 \put(20.0,170.6){\circle*{1}}
	 \put(30.0,109.365){\circle*{1}}
	 \put(40.0,72.4505){\circle*{1}}
	 \put(50.0,51.9916){\circle*{1}}
	 \put(60.0,41.1975){\circle*{1}}
	 \put(70.0,35.6505){\circle*{1}}
	 \put(80.0,32.8383){\circle*{1}}
	 \put(90.0,31.4225){\circle*{1}}
	 \put(100.0,30.7121){\circle*{1}}
	 \put(110.0,30.3562){\circle*{1}}
	 \put(120.0,30.1782){\circle*{1}}
	 \put(130.0,30.0891){\circle*{1}}
	 \put(140.0,30.0446){\circle*{1}}
	 \put(150.0,30.0223){\circle*{1}}
	 \put(160.0,30.0111){\circle*{1}}
	 \put(170.0,30.0056){\circle*{1}}
	 \color{mygray1}
	 \put(45.0,160.0){\makebox(0,0)[bl]{ QUESTION: (4a)$\ \sharp\ cause_{obj}$}}
	 \put(45.0,149.0){\makebox(0,0)[bl]{ ANSWER: $flee_{agt}\ast Pat+flee_{obj}\ast Fido$}}
	 \put(45.0,139.0){\makebox(0,0)[bl]{ NUMBER OF TRIALS: 1000}}
	 \put(45.0,127.0){\makebox(0,0)[bl]{ MEANINGFUL/NOISY BLADES: 2/2}}
	 \put(45.0,115.0){\makebox(0,0)[bl]{ (right-hand-side questions)}}	 
	 \put(80,78){\circle{1.5}}
	 \put(84,76){\makebox(0,0)[bl]{ test results}}
	 \put(80,69){\circle*{1}}
	 \put(84,65){\makebox(0,0)[bl]{ Equation (\ref{eq:complexexamplecase22})}}
	 \color{black}
	 \put(200.0,190.0){\line(1,0){151.0}}  
	 \put(200.0,170.0){\line(1,0){151.0}}  
	 \put(200.0,160.0){\line(1,0){151.0}}  
	 \put(200.0,150.0){\line(1,0){151.0}}  
	 \put(200.0,140.0){\line(1,0){151.0}}  
	 \put(200.0,130.0){\line(1,0){151.0}}  
	 \put(200.0,120.0){\line(1,0){151.0}}  
	 \put(200.0,110.0){\line(1,0){151.0}}  
	 \put(200.0,100.0){\line(1,0){151.0}}  
	 \put(200.0,90.0){\line(1,0){151.0}}  
	 \put(200.0,80.0){\line(1,0){151.0}}  
	 \put(200.0,70.0){\line(1,0){151.0}}  
	 \put(200.0,60.0){\line(1,0){151.0}}  
	 \put(200.0,50.0){\line(1,0){151.0}}  
	 \put(200.0,40.0){\line(1,0){151.0}}  
	 \put(200.0,30.0){\line(1,0){151.0}}  
	 \put(200.0,20.0){\line(1,0){151.0}}  
	 \put(200.0,10.0){\line(1,0){151.0}}  
	 \put(200.0,0.0){\line(1,0){151.0}}  
	 \put(200.0,0.0){\line(0,1){190.0}}  
	 \put(215.0,0.0){\line(0,1){190.0}}  
	 \put(264.0,0.0){\line(0,1){190.0}}  
	 \put(307.0,0.0){\line(0,1){190.0}}  
	 \put(351.0,0.0){\line(0,1){190.0}}  
	 \put(200,3.0){\makebox(0,0)[bl]{ 20}}
	 \put(200,13.0){\makebox(0,0)[bl]{ 19}}
	 \put(200,23.0){\makebox(0,0)[bl]{ 18}}
	 \put(200,33.0){\makebox(0,0)[bl]{ 17}}
	 \put(200,43.0){\makebox(0,0)[bl]{ 16}}
	 \put(200,53.0){\makebox(0,0)[bl]{ 15}}
	 \put(200,63.0){\makebox(0,0)[bl]{ 14}}
	 \put(200,73.0){\makebox(0,0)[bl]{ 13}}
	 \put(200,83.0){\makebox(0,0)[bl]{ 12}}
	 \put(200,93.0){\makebox(0,0)[bl]{ 11}}
	 \put(200,103.0){\makebox(0,0)[bl]{ 10}}
	 \put(204,113.0){\makebox(0,0)[bl]{ 9}}
	 \put(204,123.0){\makebox(0,0)[bl]{ 8}}
	 \put(204,133.0){\makebox(0,0)[bl]{ 7}}
	 \put(204,143.0){\makebox(0,0)[bl]{ 6}}
	 \put(204,153.0){\makebox(0,0)[bl]{ 5}}
	 \put(204,163.0){\makebox(0,0)[bl]{ 4}}
	 \put(202,172.0){\makebox(0,0)[bl]{ N}}
	 \put(200,180.0){\makebox(0,0)[bl]{ $(a)$}}
	 \put(217,170.5){\makebox(0,0)[bl]{{Eq. (\ref{eq:complexexamplecase22})}}}
	 \put(234,180.0){\makebox(0,0)[bl]{ $(b)$}}
	 \put(215,3.0){\makebox(0,0)[bl]{ 3.00056}} 
	 \put(215,13.0){\makebox(0,0)[bl]{ 3.00111}}
	 \put(215,23.0){\makebox(0,0)[bl]{ 3.00223}}
	 \put(215,33.0){\makebox(0,0)[bl]{ 3.00446}}
	 \put(215,43.0){\makebox(0,0)[bl]{ 3.00891}}
	 \put(215,53.0){\makebox(0,0)[bl]{ 3.01782}}
	 \put(215,63.0){\makebox(0,0)[bl]{ 3.03562}}
	 \put(215,73.0){\makebox(0,0)[bl]{ 3.07121}}
	 \put(215,83.0){\makebox(0,0)[bl]{ 3.14225}}
	 \put(215,93.0){\makebox(0,0)[bl]{ 3.28383}}
	 \put(215,103.0){\makebox(0,0)[bl]{ 3.56505}}
	 \put(215,113.0){\makebox(0,0)[bl]{ 4.11975}}
	 \put(215,123.0){\makebox(0,0)[bl]{ 5.19916}}
	 \put(215,133.0){\makebox(0,0)[bl]{ 7.24505}}
	 \put(215,143.0){\makebox(0,0)[bl]{ 10.9365}}
	 \put(215,153.0){\makebox(0,0)[bl]{ 17.06}}
	 \put(215,163.0){\makebox(0,0)[bl]{ 26.1696}}
	 \put(267,172.0){\makebox(0,0)[bl]{ test results}}
	 \put(279,180.0){\makebox(0,0)[bl]{ $(c)$}}
	 \put(263,3.0){\makebox(0,0)[bl]{ 2.496}} 
	 \put(263,13.0){\makebox(0,0)[bl]{ 2.525}}
	 \put(263,23.0){\makebox(0,0)[bl]{ 2.495}}
	 \put(263,33.0){\makebox(0,0)[bl]{ 2.508}}
	 \put(263,43.0){\makebox(0,0)[bl]{ 2.532}}
	 \put(263,53.0){\makebox(0,0)[bl]{ 2.513}}
	 \put(263,63.0){\makebox(0,0)[bl]{ 2.543}}
	 \put(263,73.0){\makebox(0,0)[bl]{ 2.556}}
	 \put(263,83.0){\makebox(0,0)[bl]{ 2.645}}
	 \put(263,93.0){\makebox(0,0)[bl]{ 2.727}}
	 \put(263,103.0){\makebox(0,0)[bl]{ 2.914}}
	 \put(263,113.0){\makebox(0,0)[bl]{ 3.411}}
	 \put(263,123.0){\makebox(0,0)[bl]{ 4.265}}
	 \put(263,133.0){\makebox(0,0)[bl]{ 5.854}}
	 \put(263,143.0){\makebox(0,0)[bl]{ 8.854}}
	 \put(263,153.0){\makebox(0,0)[bl]{ 13.42}}
	 \put(263,163.0){\makebox(0,0)[bl]{ 19.815}}
	 \put(309,176.0){\makebox(0,0)[bl]{ $|(b)-(c)|$}}
	 \put(306,3.0){\makebox(0,0)[bl]{ 0.504557}} 
	 \put(306,13.0){\makebox(0,0)[bl]{ 0.476114}}
	 \put(306,23.0){\makebox(0,0)[bl]{ 0.507228}}
	 \put(306,33.0){\makebox(0,0)[bl]{ 0.496455}}
	 \put(306,43.0){\makebox(0,0)[bl]{ 0.47691}}
	 \put(306,53.0){\makebox(0,0)[bl]{ 0.504817}}
	 \put(306,63.0){\makebox(0,0)[bl]{ 0.492624}}
	 \put(306,73.0){\makebox(0,0)[bl]{ 0.515206}}
	 \put(306,83.0){\makebox(0,0)[bl]{ 0.497247}}
	 \put(306,93.0){\makebox(0,0)[bl]{ 0.556834}}
	 \put(306,103.0){\makebox(0,0)[bl]{ 0.651048}}
	 \put(306,113.0){\makebox(0,0)[bl]{ 0.708747}}	 
	 \put(306,123.0){\makebox(0,0)[bl]{ 0.934156}}	 
	 \put(306,133.0){\makebox(0,0)[bl]{ 1.39105}}	 
	 \put(306,143.0){\makebox(0,0)[bl]{ 2.08247}}	 
	 \put(306,153.0){\makebox(0,0)[bl]{ 3.63998}}	 
	 \put(306,163.0){\makebox(0,0)[bl]{ 6.35461}}	 
  \end{picture}
  \caption{Average number of potential answers per 1000 trials with a 2:2 meaningful-to-noisy blades ratio (right-hand-side questions).}
  \label{fig:complexexamplecase22}   
\end{figure}

$\\$\textit{Case 1.} Let $s\in S_1$ and $s \neq f_0$, in a sense that $s$ might have the same blade as $f_0$ but is remembered under a different meaning in the clean-up memory. Using basic probability methods we obtain
\begin{equation}
| \langle s | (f_0 + \tilde n_1 + \dots +  \tilde n_L) \rangle | \neq 0\quad \Leftrightarrow \quad s=f_0 \ \textrm{or}\   s=\tilde n_1 \ \textrm{or}\   \dots \ \textrm{or}\   s=\tilde n_L,
\end{equation}
\begin{equation}
\textrm{P}[ s=f_0 \ \textrm{or}\   s=\tilde n_1 \ \textrm{or}\   \dots \ \textrm{or}\   s=\tilde n_L ] = \frac{L+1}{2^N}.
\end{equation}
Since all blades in $S_1$ are chosen independently, the following is true
\begin{equation}
\sum\limits_{s\in S_1, s\neq f_0} \textrm{P}[ s=f_0 \ \textrm{or}\   s=\tilde n_1 \ \textrm{or}\   \dots \ \textrm{or}\   s=\tilde n_L ] =  \frac{(|S_1|-1)(L+1)}{2^N}.
\end{equation}

$\\$\textit{Case 2.} Let $s\in S_k$ for some $1<k\leq \omega(V)$ be a multivector made of $k$ blades, $s=s_1+\dots+s_k$. Since
\begin{equation}
\textrm{P}[s\ \textrm{does not contain any of}\ \{f_0,\tilde n_1,\dots,\tilde n_l\}] = (1-\frac{L+1}{2^N})^k,
\end{equation}
we receive the following formula
\begin{equation}
\sum\limits_{s\in S_k}\textrm{P}[s\ \textrm{contains at least one of}\ \{f_0,\tilde n_1,\dots,\tilde n_l\}] = |S_k|(1-(1-\frac{L+1}{2^N})^k).
\end{equation}
Thus, when probing for an answer of $f_0 + \tilde n_1+\dots+\tilde n_L,\ L>0,$ we are likely to receive an average of
\begin{equation}
1+\frac{(|S_1|-1)(L+1)}{2^N} + \sum\limits_{k=2}^{\omega(V)} |S_k|(1-(1-\frac{L+1}{2^N})^k) \label{eq:simpleexamplecase12}
\end{equation}
potential answers.

Figure \ref{fig:simpleexamplecase13} shows test results compared with exact values given by Equation (\ref{eq:simpleexamplecase12}) for noisy answers containing one meaningful blade and three noisy blades. Note that Equation (\ref{eq:simpleexamplecase12}) is also valid for right-hand-side questions.

$\\$
The situation becomes more complex when we are to deal with answers having more than one blade. Although items in $S_1$ are always chosen independently, we cannot say the same about items belonging to $S_i,\ 1<i\leq \omega(V)$, since the sentence set is chosen by the experimenter. 
Let us consider the following question
\begin{equation}
(bite_{agt}\ast Fido+bite_{obj}\ast PSmith)\ \sharp\ bite_{obj}
\end{equation}
yielding an answer of four blades
\begin{equation}
name\ast Pat + sex\ast male + age\ast 66 + bite_{obj}^{+}\ast bite_{agt}\ast Fido.
\end{equation}
Clearly, the correct answer ($PSmith$) belongs to $S_3$, but there is one other element of the clean-up memory listed in Table \ref{tab:chmeasurestable1} that contains a portion of $PSmith$'s blades --- $DogFido$
\begin{eqnarray}
class\ast animal+type\ast dog+taste\ast chickenlike\quad\quad \nonumber\\
+name\ast Fido+ age\ast 7 +sex\ast male+occupation\ast pet,
\end{eqnarray}
the common blade being $sex\ast male$. We have two answers that under ideal conditions will surely result in a nonzero inner product: the correct answer in $S_3$ and a potential answer in $S_7$. By calculations analogous to those leading to Equation (\ref{eq:simpleexamplecase12}), the average number of answers giving a nonzero inner product for the above example is
\begin{equation}
2 + \frac{4|S_1|}{2^N} +  \sum\limits_{k\in\{3,7\}} (|S_k|-1)(1-(1-\frac{4}{2^N})^k) + \sum\limits_{k=2,k\not\in\{3,7\}}^{\omega(V)} |S_k|(1-(1-\frac{4}{2^N})^k).
\label{eq:complexexamplecase31}
\end{equation}
The number of meaningful blades in this example is odd, therefore Equation (\ref{eq:complexexamplecase31}) is also valid for right-hand-side questions. Figure \ref{fig:complexexamplecase31} shows test results compared with exact values computed by Equation (\ref{eq:complexexamplecase31}).

$\\$
Let us consider another example --- now the question is 
\begin{equation}
(4a)\ \sharp\ cause_{obj}
\end{equation} 
and the answer has a 2:2 meaningful-to-noisy blades ratio
\begin{equation}
flee_{agt}\ast Pat + flee_{obj}\ast Fido+cause_{obj}^{+}\ast(bite_{agt}\ast Fido+bite_{obj}\ast Pat) . 
\end{equation}
Apart from the correct answer in $S_2$ ($flee_{agt}\ast Pat+flee_{obj}\ast Fido$) there are also two potential answers belonging to the clean-up memory listed in Table \ref{tab:chmeasurestable1}
\begin{itemize}
\item sentence (2b) in $S_8$ --- the common blade is $flee_{agt} \ast Pat$,
\item sentence (2c) in $S_4$ --- the common blade is $flee_{obj} \ast Fido$.
\end{itemize}
Therefore, the equation for calculating the estimated number of potential answers for this example takes the following form
\begin{equation}
3+\frac{4|S_1|}{2^N}\ +  \sum\limits_{k\in\{2,4,8\}} (|S_k|-1)(1-(1-\frac{4}{2^N})^k)\ + \sum\limits_{k=3,k\not\in\{4,8\}}^{\omega(V)} |S_k|(1-(1-\frac{4}{2^N})^k),
\label{eq:complexexamplecase22}
\end{equation}
which is illustrated by Figure \ref{fig:complexexamplecase22a}.

In this example test results for right-hand-side questions (see Figure \ref{fig:complexexamplecase22}) will differ from those obtained by formula (\ref{eq:complexexamplecase22}) by about $0.5$. That is because the scalar product of (4a)$\ \sharp\ cause_{obj}$ and the correct answer will produce two \textbf{1}s which, with probability 0.5, will have opposite signs and will cancel each other out. Potential answers (2b) and (2c) do not cause such problems, since the number of their blades is odd. In half the cases the number of potential answers will be $2$ (sentences (2b) and (2c)) and in half the cases it will be $3$ (sentences (2a), (2b) and (2c)) - achieving the average of $2.5$ potential answers.

$\\$
We are now ready to work out a more general formula describing the average number of potential answers for noisy statements with multiple meaningful blades. Let $S$ and $Q$ denote the sentence and the question respectively. Let $p_k$ be the number of potential answers to $S\ \sharp\ Q$ in the subset $S_k$ of the clean-up memory $V$, denote by $L$ the number of blades in $S\ \sharp\ Q$ and let $p=p_1+\dots+p_{\omega(V)}$. The formula for calculating the estimated number of potential answers to $S\ \sharp\ Q$ then reads
\begin{equation}
p + \frac{(|S_1|-p_1)L}{2^N} + \sum\limits_{k=2}^{\omega(V)} (|S_k|-p_k)(1-(1-\frac{L}{2^N})^k),
\end{equation}
provided, that we use appropriate-hand-side reversed questions. As far as right-hand-side questions are concerned, this equation may be regarded only as the upper bound due to cancellation --- for a closer estimate, one should investigate elements of the clean-up memory that have an even number of blades.


\section{Comparison with Previously Developed Models}
\label{sec:6}

The most important performance measure of any new distributed representation model is the comparison of its efficiency in relation to previously developed models. This Section comments on test results performed on GA, BSC and HRR. 

Naturally, the question of data size arises as a GA clean-up memory item may store information in more than one vector (blade), unlike in architectures known so far. Further, the preferred way of recognition for GA requires the usage of matrix signatures comprising up to $2^{1+2\lceil \frac{N}{2} \rceil}$ entries. However, since one only needs blades to calculate the matrix signatures, it has been assumed that tests comparing efficiency of various models should be conducted using the following sizes of data
\begin{itemize}
\item $N$ bits for a single blade in GA,
\item $KN$ bits for a single vector in BSC and HRR,
\end{itemize}
where $K$ is the maximum number of blades stored in a complex sentence belonging to GA's clean-up memory under agent-object construction with odding blades. For the data set presented in Table~\ref{tab:chmeasurestable1} the maximum number of blades is stored in items (3d) and (5b) and is equal to $13$. Such an approach to the test data size will certainly prove redundant for GA sentences having a lesser number of blades, nevertheless, it is only fair to provide relatively the same data size for all compared models.

Figures~\ref{fig:compA} through \ref{fig:compG} show comparison of performance for GA, BSC and HRR, tested sentences range in meaningful-to-noisy blades ratio from 1:2 to 7:2. Clearly, GA with the use of Hamming and Euclidean measure ensures quite a remarkable recognition percentage for sentences of great complexity and therefore --- great noise, whereas the HRR model works better for statements of low complexity. There is no significant difference in performance of the BSC model as far as complexity of tested sentences is concerned. BSC does remain the best model, provided that vector lengths for BSC are sufficiently longer than that of GA. Under uniform length of vectors and blades GA recognizes sentences better than HRR or BSC, regardless of their complexity. 

\definecolor{mygray1}{rgb}{0.5,0.5,0.5}
\definecolor{mygreen1}{rgb}{0,0.8,0}
\begin{figure}[ht!]		
  \begin{picture}(0,500)(0,-5)
	 \put(200.0,410.0){\line(1,0){129.0}}  
	 \put(200.0,400.0){\line(1,0){129.0}}  	 
	 \put(200.0,390.0){\line(1,0){129.0}}  
	 \put(200.0,380.0){\line(1,0){129.0}}  
	 \put(200.0,370.0){\line(1,0){129.0}}  
	 \put(200.0,360.0){\line(1,0){129.0}}  
	 \put(200.0,350.0){\line(1,0){129.0}}  
	 \put(200.0,340.0){\line(1,0){129.0}}  
	 \put(200.0,330.0){\line(1,0){129.0}}  
	 \put(200.0,320.0){\line(1,0){129.0}}  
	 \put(200.0,310.0){\line(1,0){129.0}}  
	 \put(200.0,300.0){\line(1,0){129.0}}  
	 \put(200.0,290.0){\line(1,0){129.0}}  
	 \put(200.0,280.0){\line(1,0){129.0}}  
	 \put(200.0,270.0){\line(1,0){129.0}}  
	 \put(200.0,260.0){\line(1,0){129.0}}  
	 \put(200.0,250.0){\line(1,0){129.0}}  
	 \put(200.0,240.0){\line(1,0){129.0}}  
	 \put(200.0,230.0){\line(1,0){129.0}}  
	 \put(200.0,180.0){\line(1,0){129.0}}  
	 \put(200.0,170.0){\line(1,0){129.0}}  
	 \put(200.0,160.0){\line(1,0){129.0}}  
	 \put(200.0,150.0){\line(1,0){129.0}}  
	 \put(200.0,140.0){\line(1,0){129.0}}  
	 \put(200.0,130.0){\line(1,0){129.0}}  
	 \put(200.0,120.0){\line(1,0){129.0}}  
	 \put(200.0,110.0){\line(1,0){129.0}}  
	 \put(200.0,100.0){\line(1,0){129.0}}  
	 \put(200.0,90.0){\line(1,0){129.0}}  
	 \put(200.0,80.0){\line(1,0){129.0}}  
	 \put(200.0,70.0){\line(1,0){129.0}}  
	 \put(200.0,60.0){\line(1,0){129.0}}  
	 \put(200.0,50.0){\line(1,0){129.0}}  
	 \put(200.0,40.0){\line(1,0){129.0}}  
	 \put(200.0,30.0){\line(1,0){129.0}}  
	 \put(200.0,20.0){\line(1,0){129.0}}  
	 \put(200.0,10.0){\line(1,0){129.0}}  
	 \put(200.0,0.0){\line(1,0){129.0}}  
	 \put(200.0,230.0){\line(0,1){180.0}}  
	 \put(215.0,230.0){\line(0,1){180.0}}  
	 \put(253.0,230.0){\line(0,1){180.0}}  
	 \put(292.0,230.0){\line(0,1){180.0}}  
	 \put(329.0,230.0){\line(0,1){180.0}}  
	 
	 \put(200.0,0.0){\line(0,1){180.0}}  
	 \put(215.0,0.0){\line(0,1){180.0}}  
	 \put(253.0,0.0){\line(0,1){180.0}}  
	 \put(292.0,0.0){\line(0,1){180.0}}  
	 \put(329.0,0.0){\line(0,1){180.0}}  
	 \put(203,402.5){\makebox(0,0)[bl]{ N}}
	 \put(217,401.0){\makebox(0,0)[bl]{{Hamming}}}
	 \put(264,403.0){\makebox(0,0)[bl]{ HRR}}
	 \put(300,403.0){\makebox(0,0)[bl]{ BSC}}	 

	 \put(203,172.5){\makebox(0,0)[bl]{ N}}
	 \put(217,171.0){\makebox(0,0)[bl]{{Hamming}}}
	 \put(264,173.0){\makebox(0,0)[bl]{ HRR}}
	 \put(300,173.0){\makebox(0,0)[bl]{ BSC}}
	 \put(203,233.0){\makebox(0,0)[bl]{ 20}}
	 \put(203,243.0){\makebox(0,0)[bl]{ 19}}
	 \put(203,253.0){\makebox(0,0)[bl]{ 18}}
	 \put(203,263.0){\makebox(0,0)[bl]{ 17}}
	 \put(203,273.0){\makebox(0,0)[bl]{ 16}}
	 \put(203,283.0){\makebox(0,0)[bl]{ 15}}
	 \put(203,293.0){\makebox(0,0)[bl]{ 14}}
	 \put(203,303.0){\makebox(0,0)[bl]{ 13}}
	 \put(203,313.0){\makebox(0,0)[bl]{ 12}}
	 \put(203,323.0){\makebox(0,0)[bl]{ 11}}
	 \put(203,333.0){\makebox(0,0)[bl]{ 10}}
	 \put(207,343.0){\makebox(0,0)[bl]{ 9}}
	 \put(207,353.0){\makebox(0,0)[bl]{ 8}}
	 \put(207,363.0){\makebox(0,0)[bl]{ 7}}
	 \put(207,373.0){\makebox(0,0)[bl]{ 6}}
	 \put(207,383.0){\makebox(0,0)[bl]{ 5}}
	 \put(207,393.0){\makebox(0,0)[bl]{ 4}}
	 \put(218,233.0){\makebox(0,0)[bl]{ 100.0\%}}
	 \put(218,243.0){\makebox(0,0)[bl]{ 100.0\%}}
	 \put(218,253.0){\makebox(0,0)[bl]{ 100.0\%}}
	 \put(218,263.0){\makebox(0,0)[bl]{ 100.0\%}}
	 \put(218,273.0){\makebox(0,0)[bl]{ 100.0\%}}
	 \put(218,283.0){\makebox(0,0)[bl]{ 100.0\%}}
	 \put(218,293.0){\makebox(0,0)[bl]{ 99.2\%}}
	 \put(218,303.0){\makebox(0,0)[bl]{ 99.0\%}}
	 \put(218,313.0){\makebox(0,0)[bl]{ 98.2\%}}
	 \put(218,323.0){\makebox(0,0)[bl]{ 97.4\%}}
	 \put(218,333.0){\makebox(0,0)[bl]{ 95.2\%}}
	 \put(218,343.0){\makebox(0,0)[bl]{ 90.2\%}}
	 \put(218,353.0){\makebox(0,0)[bl]{ 77.6\%}}
	 \put(218,363.0){\makebox(0,0)[bl]{ 60.4\%}}
	 \put(218,373.0){\makebox(0,0)[bl]{ 32.4\%}}
	 \put(218,383.0){\makebox(0,0)[bl]{ 16.0\%}}
	 \put(218,393.0){\makebox(0,0)[bl]{ 0.8\%}}
	 \put(256,233.0){\makebox(0,0)[bl]{ 49.0\%}}
	 \put(256,243.0){\makebox(0,0)[bl]{ 43.3\%}}
	 \put(256,253.0){\makebox(0,0)[bl]{ 40.1\%}}
	 \put(256,263.0){\makebox(0,0)[bl]{ 41.7\%}}
	 \put(256,273.0){\makebox(0,0)[bl]{ 37.9\%}}
	 \put(256,283.0){\makebox(0,0)[bl]{ 34.3\%}}
	 \put(256,293.0){\makebox(0,0)[bl]{ 33.2\%}}
	 \put(256,303.0){\makebox(0,0)[bl]{ 32.0\%}}
	 \put(256,313.0){\makebox(0,0)[bl]{ 25.7\%}}
	 \put(256,323.0){\makebox(0,0)[bl]{ 25.5\%}}
	 \put(256,333.0){\makebox(0,0)[bl]{ 21.4\%}}
	 \put(256,343.0){\makebox(0,0)[bl]{ 17.7\%}}
	 \put(256,353.0){\makebox(0,0)[bl]{ 13.8\%}}
	 \put(256,363.0){\makebox(0,0)[bl]{ 12.9\%}}
	 \put(256,373.0){\makebox(0,0)[bl]{ 8.9\%}}
	 \put(256,383.0){\makebox(0,0)[bl]{ 9.7\%}}
	 \put(256,393.0){\makebox(0,0)[bl]{ 3.2\%}}
	 \put(295,233.0){\makebox(0,0)[bl]{ 57.9\%}}
	 \put(295,243.0){\makebox(0,0)[bl]{ 57.8\%}}
	 \put(295,253.0){\makebox(0,0)[bl]{ 55.3\%}}
	 \put(295,263.0){\makebox(0,0)[bl]{ 52.7\%}}
	 \put(295,273.0){\makebox(0,0)[bl]{ 54.3\%}}
	 \put(295,283.0){\makebox(0,0)[bl]{ 51.4\%}}
	 \put(295,293.0){\makebox(0,0)[bl]{ 47.1\%}}
	 \put(295,303.0){\makebox(0,0)[bl]{ 46.7\%}}
	 \put(295,313.0){\makebox(0,0)[bl]{ 44.7\%}}
	 \put(295,323.0){\makebox(0,0)[bl]{ 40.6\%}}
	 \put(295,333.0){\makebox(0,0)[bl]{ 37.3\%}}
	 \put(295,343.0){\makebox(0,0)[bl]{ 37.3\%}}
	 \put(295,353.0){\makebox(0,0)[bl]{ 35.8\%}}
	 \put(295,363.0){\makebox(0,0)[bl]{ 33.5\%}}
	 \put(295,373.0){\makebox(0,0)[bl]{ 34.9\%}}
	 \put(295,383.0){\makebox(0,0)[bl]{ 29.5\%}}
	 \put(295,393.0){\makebox(0,0)[bl]{ 32.5\%}}
	 \put(203,3.0){\makebox(0,0)[bl]{ 20}}
	 \put(203,13.0){\makebox(0,0)[bl]{ 19}}
	 \put(203,23.0){\makebox(0,0)[bl]{ 18}}
	 \put(203,33.0){\makebox(0,0)[bl]{ 17}}
	 \put(203,43.0){\makebox(0,0)[bl]{ 16}}
	 \put(203,53.0){\makebox(0,0)[bl]{ 15}}
	 \put(203,63.0){\makebox(0,0)[bl]{ 14}}
	 \put(203,73.0){\makebox(0,0)[bl]{ 13}}
	 \put(203,83.0){\makebox(0,0)[bl]{ 12}}
	 \put(203,93.0){\makebox(0,0)[bl]{ 11}}
	 \put(203,103.0){\makebox(0,0)[bl]{ 10}}
	 \put(207,113.0){\makebox(0,0)[bl]{ 9}}
	 \put(207,123.0){\makebox(0,0)[bl]{ 8}}
	 \put(207,133.0){\makebox(0,0)[bl]{ 7}}
	 \put(207,143.0){\makebox(0,0)[bl]{ 6}}
	 \put(207,153.0){\makebox(0,0)[bl]{ 5}}
	 \put(207,163.0){\makebox(0,0)[bl]{ 4}}
	 \put(218,3.0){\makebox(0,0)[bl]{ 100.0\%}}
	 \put(218,13.0){\makebox(0,0)[bl]{ 100.0\%}}
	 \put(218,23.0){\makebox(0,0)[bl]{ 100.0\%}}
	 \put(218,33.0){\makebox(0,0)[bl]{ 100.0\%}}
	 \put(218,43.0){\makebox(0,0)[bl]{ 100.0\%}}
	 \put(218,53.0){\makebox(0,0)[bl]{ 100.0\%}}
	 \put(218,63.0){\makebox(0,0)[bl]{ 99.2\%}}
	 \put(218,73.0){\makebox(0,0)[bl]{ 99.0\%}}
	 \put(218,83.0){\makebox(0,0)[bl]{ 98.2\%}}
	 \put(218,93.0){\makebox(0,0)[bl]{ 97.4\%}}
	 \put(218,103.0){\makebox(0,0)[bl]{ 95.2\%}}
	 \put(218,113.0){\makebox(0,0)[bl]{ 90.2\%}}
	 \put(218,123.0){\makebox(0,0)[bl]{ 77.6\%}}
	 \put(218,133.0){\makebox(0,0)[bl]{ 60.4\%}}
	 \put(218,143.0){\makebox(0,0)[bl]{ 32.4\%}}
	 \put(218,153.0){\makebox(0,0)[bl]{ 16.0\%}}
	 \put(218,163.0){\makebox(0,0)[bl]{ 0.8\%}}
	 \put(256,3.0){\makebox(0,0)[bl]{ 100.0\%}}
	 \put(256,13.0){\makebox(0,0)[bl]{ 100.0\%}}
	 \put(256,23.0){\makebox(0,0)[bl]{ 100.0\%}}
	 \put(256,33.0){\makebox(0,0)[bl]{ 100.0\%}}
	 \put(256,43.0){\makebox(0,0)[bl]{ 99.9\%}}
	 \put(256,53.0){\makebox(0,0)[bl]{ 100.0\%}}
	 \put(256,63.0){\makebox(0,0)[bl]{ 100.0\%}}
	 \put(256,73.0){\makebox(0,0)[bl]{ 100.0\%}}
	 \put(256,83.0){\makebox(0,0)[bl]{ 99.8\%}}
	 \put(256,93.0){\makebox(0,0)[bl]{ 99.9\%}}
	 \put(256,103.0){\makebox(0,0)[bl]{ 99.6\%}}
	 \put(256,113.0){\makebox(0,0)[bl]{ 99.6\%}}
	 \put(256,123.0){\makebox(0,0)[bl]{ 99.5\%}}
	 \put(256,133.0){\makebox(0,0)[bl]{ 98.6\%}}
	 \put(256,143.0){\makebox(0,0)[bl]{ 97.7\%}}
	 \put(256,153.0){\makebox(0,0)[bl]{ 95.9\%}}
	 \put(256,163.0){\makebox(0,0)[bl]{ 90.6\%}}
	 \put(295,3.0){\makebox(0,0)[bl]{ 100.0\%}}
	 \put(295,13.0){\makebox(0,0)[bl]{ 100.0\%}}
	 \put(295,23.0){\makebox(0,0)[bl]{ 100.0\%}}
	 \put(295,33.0){\makebox(0,0)[bl]{ 100.0\%}}
	 \put(295,43.0){\makebox(0,0)[bl]{ 100.0\%}}
	 \put(295,53.0){\makebox(0,0)[bl]{ 100.0\%}}
	 \put(295,63.0){\makebox(0,0)[bl]{ 100.0\%}}
	 \put(295,73.0){\makebox(0,0)[bl]{ 100.0\%}}
	 \put(295,83.0){\makebox(0,0)[bl]{ 100.0\%}}
	 \put(295,93.0){\makebox(0,0)[bl]{ 100.0\%}}
	 \put(295,103.0){\makebox(0,0)[bl]{ 99.9\%}}
	 \put(295,113.0){\makebox(0,0)[bl]{ 99.7\%}}
	 \put(295,123.0){\makebox(0,0)[bl]{ 99.7\%}}
	 \put(295,133.0){\makebox(0,0)[bl]{ 99.1\%}}
	 \put(295,143.0){\makebox(0,0)[bl]{ 98.7\%}}
	 \put(295,153.0){\makebox(0,0)[bl]{ 97.4\%}}
	 \put(295,163.0){\makebox(0,0)[bl]{ 92.2\%}}
	 \color{mygray1}
	 \put(0,475.0){\makebox(0,0)[bl]{ QUESTION: $PSmith\ \sharp\ name$}}
	 \put(0,464.5){\makebox(0,0)[bl]{ ANSWER: $Pat$}}
	 \put(0,453.5){\makebox(0,0)[bl]{ NUMBER OF TRIALS: 1000}}
	 \put(0,441.0){\makebox(0,0)[bl]{ GA CONSTRUCTION: Agent-object with odding blades, right-hand-side questions}}
	 \put(0,431.0){\makebox(0,0)[bl]{ MEANINGFUL/NOISY BLADES: 1/2}}
	 \put(4,422){\includegraphics{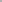}}
	 \put(8,418.5){\makebox(0,0)[bl]{ GA, Hamming measure}}	 
	 \put(4.5,413){\circle{1.5}}
	 \put(8,410){\makebox(0,0)[bl]{ BSC}}
	 \put(4.5,404){\circle*{1}}
	 \put(8,400){\makebox(0,0)[bl]{ HRR}}	 
	 \color{black}
	 \put(0.0,230.0){\vector(1,0){180.0}}  
	 \put(0.0,230.0){\vector(0,1){120.0}} 
	 \put(10.0,232){\line(0,-1){4.0}} 
	 \put(20.0,232){\line(0,-1){4.0}} 
	 \put(30.0,232){\line(0,-1){4.0}} 
	 \put(40.0,232){\line(0,-1){4.0}} 
	 \put(50.0,232){\line(0,-1){4.0}} 
	 \put(60.0,232){\line(0,-1){4.0}} 
	 \put(70.0,232){\line(0,-1){4.0}} 
	 \put(80.0,232){\line(0,-1){4.0}} 
	 \put(90.0,232){\line(0,-1){4.0}} 
	 \put(100.0,232){\line(0,-1){4.0}} 
	 \put(110.0,232){\line(0,-1){4.0}} 
	 \put(120.0,232){\line(0,-1){4.0}} 
	 \put(130.0,232){\line(0,-1){4.0}} 
	 \put(140.0,232){\line(0,-1){4.0}} 
	 \put(150.0,232){\line(0,-1){4.0}} 
	 \put(160.0,232){\line(0,-1){4.0}} 
	 \put(170.0,232){\line(0,-1){4.0}} 
	 \put(-2.0,240.0){\line(1,0){4.0}} 
	 \put(-2.0,250.0){\line(1,0){4.0}} 
	 \put(-2.0,260.0){\line(1,0){4.0}} 
	 \put(-2.0,270.0){\line(1,0){4.0}} 
	 \put(-2.0,280.0){\line(1,0){4.0}} 
	 \put(-2.0,290.0){\line(1,0){4.0}} 
	 \put(-2.0,300.0){\line(1,0){4.0}} 
	 \put(-2.0,310.0){\line(1,0){4.0}} 
	 \put(-2.0,320.0){\line(1,0){4.0}} 
	 \put(-2.0,330.0){\line(1,0){4.0}} 
	 \put(15.0,221.0){\makebox(0,0)[bl]{ 5}}
	 \put(63.0,221.0){\makebox(0,0)[bl]{ 10}}
	 \put(113.0,221.0){\makebox(0,0)[bl]{ 15}}
	 \put(163.0,221.0){\makebox(0,0)[bl]{ 20}}
	 \put(180.0,226.0){\makebox(0,0)[bl]{ N}}
	 \put(-21,237.0){\makebox(0,0)[bl]{ 10\%}}
	 \put(-21,277.0){\makebox(0,0)[bl]{ 50\%}}
	 \put(-25,327.0){\makebox(0,0)[bl]{ 100\%}}
	 \put(-7.0,352.0){\makebox(0,0)[bl]{[\%] recognition}}
	 \color{mygray1}
	 \put(9,230.3){\includegraphics{sq.jpg}}
	 \put(19,245.5){\includegraphics{sq.jpg}}
	 \put(29,261.9){\includegraphics{sq.jpg}}
	 \put(39,289.9){\includegraphics{sq.jpg}}
	 \put(49,307.1){\includegraphics{sq.jpg}}
	 \put(59,319.7){\includegraphics{sq.jpg}}
	 \put(69,324.7){\includegraphics{sq.jpg}}
	 \put(79,326.9){\includegraphics{sq.jpg}}
	 \put(89,327.7){\includegraphics{sq.jpg}}
	 \put(99,328.5){\includegraphics{sq.jpg}}
	 \put(109,328.7){\includegraphics{sq.jpg}}
	 \put(119,329.5){\includegraphics{sq.jpg}}
	 \put(129,329.5){\includegraphics{sq.jpg}}
	 \put(139,329.5){\includegraphics{sq.jpg}}
	 \put(149,329.5){\includegraphics{sq.jpg}}
	 \put(159,329.5){\includegraphics{sq.jpg}}
	 \put(169,329.5){\includegraphics{sq.jpg}}
	 \color{black}
	 \put(10.0,262.5){\circle{1.5}}
	 \put(20.0,259.5){\circle{1.5}}
	 \put(30.0,264.9){\circle{1.5}}
	 \put(40.0,263.5){\circle{1.5}}
	 \put(50.0,265.8){\circle{1.5}}
	 \put(60.0,267.3){\circle{1.5}}
	 \put(70.0,267.3){\circle{1.5}}
	 \put(80.0,270.6){\circle{1.5}}
	 \put(90.0,274.7){\circle{1.5}}
	 \put(100.0,276.7){\circle{1.5}}
	 \put(110.0,277.1){\circle{1.5}}
	 \put(120.0,281.4){\circle{1.5}}
	 \put(130.0,284.3){\circle{1.5}}
	 \put(140.0,282.7){\circle{1.5}}
	 \put(150.0,285.3){\circle{1.5}}
	 \put(160.0,287.8){\circle{1.5}}
	 \put(170.0,287.9){\circle{1.5}}
	 \put(10.0,233.2){\circle*{1}}
	 \put(20.0,239.7){\circle*{1}}
	 \put(30.0,238.9){\circle*{1}}
	 \put(40.0,242.9){\circle*{1}}
	 \put(50.0,243.8){\circle*{1}}
	 \put(60.0,247.7){\circle*{1}}
	 \put(70.0,251.4){\circle*{1}}
	 \put(80.0,255.5){\circle*{1}}
	 \put(90.0,255.7){\circle*{1}}
	 \put(100.0,262.0){\circle*{1}}
	 \put(110.0,263.2){\circle*{1}}
	 \put(120.0,264.3){\circle*{1}}
	 \put(130.0,267.9){\circle*{1}}
	 \put(140.0,271.7){\circle*{1}}
	 \put(150.0,270.1){\circle*{1}}
	 \put(160.0,273.3){\circle*{1}}
	 \put(170.0,279.0){\circle*{1}}
	 \put(0.0,0.0){\vector(1,0){180.0}}  
	 \put(0.0,0.0){\vector(0,1){120.0}} 
	 \put(10.0,2.0){\line(0,-1){4.0}} 
	 \put(20.0,2.0){\line(0,-1){4.0}} 
	 \put(30.0,2.0){\line(0,-1){4.0}} 
	 \put(40.0,2.0){\line(0,-1){4.0}} 
	 \put(50.0,2.0){\line(0,-1){4.0}} 
	 \put(60.0,2.0){\line(0,-1){4.0}} 
	 \put(70.0,2.0){\line(0,-1){4.0}} 
	 \put(80.0,2.0){\line(0,-1){4.0}} 
	 \put(90.0,2.0){\line(0,-1){4.0}} 
	 \put(100.0,2.0){\line(0,-1){4.0}} 
	 \put(110.0,2.0){\line(0,-1){4.0}} 
	 \put(120.0,2.0){\line(0,-1){4.0}} 
	 \put(130.0,2.0){\line(0,-1){4.0}} 
	 \put(140.0,2.0){\line(0,-1){4.0}} 
	 \put(150.0,2.0){\line(0,-1){4.0}} 
	 \put(160.0,2.0){\line(0,-1){4.0}} 
	 \put(170.0,2.0){\line(0,-1){4.0}} 
	 \put(-2.0,10.0){\line(1,0){4.0}} 
	 \put(-2.0,20.0){\line(1,0){4.0}} 
	 \put(-2.0,30.0){\line(1,0){4.0}} 
	 \put(-2.0,40.0){\line(1,0){4.0}} 
	 \put(-2.0,50.0){\line(1,0){4.0}} 
	 \put(-2.0,60.0){\line(1,0){4.0}} 
	 \put(-2.0,70.0){\line(1,0){4.0}} 
	 \put(-2.0,80.0){\line(1,0){4.0}} 
	 \put(-2.0,90.0){\line(1,0){4.0}} 
	 \put(-2.0,100.0){\line(1,0){4.0}} 
	 \put(15.0,-9.0){\makebox(0,0)[bl]{ 5}}
	 \put(63.0,-9.0){\makebox(0,0)[bl]{ 10}}
	 \put(113.0,-9.0){\makebox(0,0)[bl]{ 15}}
	 \put(163.0,-9.0){\makebox(0,0)[bl]{ 20}}
	 \put(180.0,-4.0){\makebox(0,0)[bl]{ N}}
	 \put(-21,7.0){\makebox(0,0)[bl]{ 10\%}}
	 \put(-21,47.0){\makebox(0,0)[bl]{ 50\%}}
	 \put(-25,97.0){\makebox(0,0)[bl]{ 100\%}}
	 \put(-7.0,122.0){\makebox(0,0)[bl]{[\%] recognition}}
	 \color{mygray1}
	 \put(9,0.3){\includegraphics{sq.jpg}}
	 \put(19,15.5){\includegraphics{sq.jpg}}
	 \put(29,31.9){\includegraphics{sq.jpg}}
	 \put(39,59.9){\includegraphics{sq.jpg}}
	 \put(49,77.1){\includegraphics{sq.jpg}}
	 \put(59,89.7){\includegraphics{sq.jpg}}
	 \put(69,94.7){\includegraphics{sq.jpg}}
	 \put(79,96.9){\includegraphics{sq.jpg}}
	 \put(89,97.7){\includegraphics{sq.jpg}}
	 \put(99,98.5){\includegraphics{sq.jpg}}
	 \put(109,98.7){\includegraphics{sq.jpg}}
	 \put(119,99.5){\includegraphics{sq.jpg}}
	 \put(129,99.5){\includegraphics{sq.jpg}}
	 \put(139,99.5){\includegraphics{sq.jpg}}
	 \put(149,99.5){\includegraphics{sq.jpg}}
	 \put(159,99.5){\includegraphics{sq.jpg}}
	 \put(169,99.5){\includegraphics{sq.jpg}}
	 \color{black}
	 \put(10.0,92.2){\circle{1.5}}
	 \put(20.0,97.4){\circle{1.5}}
	 \put(30.0,98.7){\circle{1.5}}
	 \put(40.0,99.1){\circle{1.5}}
	 \put(50.0,99.7){\circle{1.5}}
	 \put(60.0,99.7){\circle{1.5}}
	 \put(70.0,99.9){\circle{1.5}}
	 \put(80.0,100.0){\circle{1.5}}
	 \put(90.0,100.0){\circle{1.5}}
	 \put(100.0,100.0){\circle{1.5}}
	 \put(110.0,100.0){\circle{1.5}}
	 \put(120.0,100.0){\circle{1.5}}
	 \put(130.0,100.0){\circle{1.5}}
	 \put(140.0,100.0){\circle{1.5}}
	 \put(150.0,100.0){\circle{1.5}}
	 \put(160.0,100.0){\circle{1.5}}
	 \put(170.0,100.0){\circle{1.5}}
	 \put(10.0,90.6){\circle*{1}}
	 \put(20.0,95.9){\circle*{1}}
	 \put(30.0,97.7){\circle*{1}}
	 \put(40.0,98.6){\circle*{1}}
	 \put(50.0,99.5){\circle*{1}}
	 \put(60.0,99.6){\circle*{1}}
	 \put(70.0,99.6){\circle*{1}}
	 \put(80.0,99.9){\circle*{1}}
	 \put(90.0,99.8){\circle*{1}}
	 \put(100.0,100.0){\circle*{1}}
	 \put(110.0,100.0){\circle*{1}}
	 \put(120.0,100.0){\circle*{1}}
	 \put(130.0,99.9){\circle*{1}}
	 \put(140.0,100.0){\circle*{1}}
	 \put(150.0,100.0){\circle*{1}}
	 \put(160.0,100.0){\circle*{1}}
	 \put(170.0,100.0){\circle*{1}}

    \put(0,136){\makebox(0,0)[bl]{{Vector lengths: $N$ for GA, $13N$ for HRR and BSC.}}}
	 \put(0,366){\makebox(0,0)[bl]{{All vector lengths: $N$.}}}	 	 
  \end{picture}
  \caption{Comparison of recognition for GA, BSC and HRR -- $PSmith\ \sharp\ name$.}
  \label{fig:compA}   
\end{figure}

\definecolor{mygray1}{rgb}{0.5,0.5,0.5}
\begin{figure}[ht!]
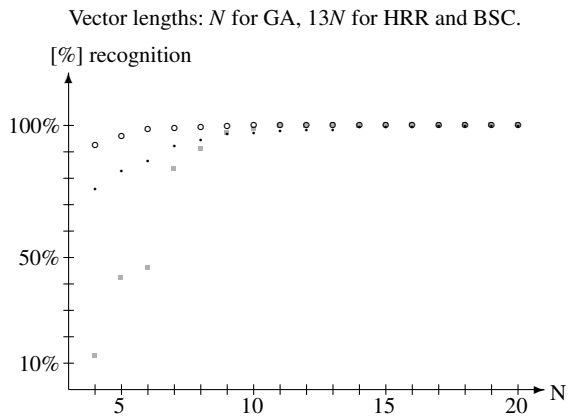
		
  \begin{picture}(0,500)(0,-5)
	 \put(200.0,410.0){\line(1,0){129.0}}  
	 \put(200.0,400.0){\line(1,0){129.0}}  	 
	 \put(200.0,390.0){\line(1,0){129.0}}  
	 \put(200.0,380.0){\line(1,0){129.0}}  
	 \put(200.0,370.0){\line(1,0){129.0}}  
	 \put(200.0,360.0){\line(1,0){129.0}}  
	 \put(200.0,350.0){\line(1,0){129.0}}  
	 \put(200.0,340.0){\line(1,0){129.0}}  
	 \put(200.0,330.0){\line(1,0){129.0}}  
	 \put(200.0,320.0){\line(1,0){129.0}}  
	 \put(200.0,310.0){\line(1,0){129.0}}  
	 \put(200.0,300.0){\line(1,0){129.0}}  
	 \put(200.0,290.0){\line(1,0){129.0}}  
	 \put(200.0,280.0){\line(1,0){129.0}}  
	 \put(200.0,270.0){\line(1,0){129.0}}  
	 \put(200.0,260.0){\line(1,0){129.0}}  
	 \put(200.0,250.0){\line(1,0){129.0}}  
	 \put(200.0,240.0){\line(1,0){129.0}}  
	 \put(200.0,230.0){\line(1,0){129.0}}  
	 \put(200.0,180.0){\line(1,0){129.0}}  
	 \put(200.0,170.0){\line(1,0){129.0}}  
	 \put(200.0,160.0){\line(1,0){129.0}}  
	 \put(200.0,150.0){\line(1,0){129.0}}  
	 \put(200.0,140.0){\line(1,0){129.0}}  
	 \put(200.0,130.0){\line(1,0){129.0}}  
	 \put(200.0,120.0){\line(1,0){129.0}}  
	 \put(200.0,110.0){\line(1,0){129.0}}  
	 \put(200.0,100.0){\line(1,0){129.0}}  
	 \put(200.0,90.0){\line(1,0){129.0}}  
	 \put(200.0,80.0){\line(1,0){129.0}}  
	 \put(200.0,70.0){\line(1,0){129.0}}  
	 \put(200.0,60.0){\line(1,0){129.0}}  
	 \put(200.0,50.0){\line(1,0){129.0}}  
	 \put(200.0,40.0){\line(1,0){129.0}}  
	 \put(200.0,30.0){\line(1,0){129.0}}  
	 \put(200.0,20.0){\line(1,0){129.0}}  
	 \put(200.0,10.0){\line(1,0){129.0}}  
	 \put(200.0,0.0){\line(1,0){129.0}}  
	 \put(200.0,230.0){\line(0,1){180.0}}  
	 \put(215.0,230.0){\line(0,1){180.0}}  
	 \put(253.0,230.0){\line(0,1){180.0}}  
	 \put(292.0,230.0){\line(0,1){180.0}}  
	 \put(329.0,230.0){\line(0,1){180.0}}  
	 
	 \put(200.0,0.0){\line(0,1){180.0}}  
	 \put(215.0,0.0){\line(0,1){180.0}}  
	 \put(253.0,0.0){\line(0,1){180.0}}  
	 \put(292.0,0.0){\line(0,1){180.0}}  
	 \put(329.0,0.0){\line(0,1){180.0}}  
	 \put(203,402.5){\makebox(0,0)[bl]{ N}}
	 \put(217,401.0){\makebox(0,0)[bl]{{Hamming}}}
	 \put(264,403.0){\makebox(0,0)[bl]{ HRR}}
	 \put(300,403.0){\makebox(0,0)[bl]{ BSC}}	 

	 \put(203,172.5){\makebox(0,0)[bl]{ N}}
	 \put(217,171.0){\makebox(0,0)[bl]{{Hamming}}}
	 \put(264,173.0){\makebox(0,0)[bl]{ HRR}}
	 \put(300,173.0){\makebox(0,0)[bl]{ BSC}}
	 \put(203,233.0){\makebox(0,0)[bl]{ 20}}
	 \put(203,243.0){\makebox(0,0)[bl]{ 19}}
	 \put(203,253.0){\makebox(0,0)[bl]{ 18}}
	 \put(203,263.0){\makebox(0,0)[bl]{ 17}}
	 \put(203,273.0){\makebox(0,0)[bl]{ 16}}
	 \put(203,283.0){\makebox(0,0)[bl]{ 15}}
	 \put(203,293.0){\makebox(0,0)[bl]{ 14}}
	 \put(203,303.0){\makebox(0,0)[bl]{ 13}}
	 \put(203,313.0){\makebox(0,0)[bl]{ 12}}
	 \put(203,323.0){\makebox(0,0)[bl]{ 11}}
	 \put(203,333.0){\makebox(0,0)[bl]{ 10}}
	 \put(207,343.0){\makebox(0,0)[bl]{ 9}}
	 \put(207,353.0){\makebox(0,0)[bl]{ 8}}
	 \put(207,363.0){\makebox(0,0)[bl]{ 7}}
	 \put(207,373.0){\makebox(0,0)[bl]{ 6}}
	 \put(207,383.0){\makebox(0,0)[bl]{ 5}}
	 \put(207,393.0){\makebox(0,0)[bl]{ 4}}
	 \put(218,233.0){\makebox(0,0)[bl]{ 100.0\%}}
	 \put(218,243.0){\makebox(0,0)[bl]{ 100.0\%}}
	 \put(218,253.0){\makebox(0,0)[bl]{ 100.0\%}}
	 \put(218,263.0){\makebox(0,0)[bl]{ 100.0\%}}
	 \put(218,273.0){\makebox(0,0)[bl]{ 100.0\%}}
	 \put(218,283.0){\makebox(0,0)[bl]{ 100.0\%}}
	 \put(218,293.0){\makebox(0,0)[bl]{ 99.8\%}}
	 \put(218,303.0){\makebox(0,0)[bl]{ 100.0\%}}
	 \put(218,313.0){\makebox(0,0)[bl]{ 99.6\%}}
	 \put(218,323.0){\makebox(0,0)[bl]{ 99.8\%}}
	 \put(218,333.0){\makebox(0,0)[bl]{ 98.6\%}}
	 \put(218,343.0){\makebox(0,0)[bl]{ 97.0\%}}
	 \put(218,353.0){\makebox(0,0)[bl]{ 91.0\%}}
	 \put(218,363.0){\makebox(0,0)[bl]{ 83.4\%}}
	 \put(218,373.0){\makebox(0,0)[bl]{ 45.8\%}}
	 \put(218,383.0){\makebox(0,0)[bl]{ 42.2\%}}
	 \put(218,393.0){\makebox(0,0)[bl]{ 12.6\%}}
	 \put(256,233.0){\makebox(0,0)[bl]{ 34.6\%}}
	 \put(256,243.0){\makebox(0,0)[bl]{ 32.9\%}}
	 \put(256,253.0){\makebox(0,0)[bl]{ 32.2\%}}
	 \put(256,263.0){\makebox(0,0)[bl]{ 30.9\%}}
	 \put(256,273.0){\makebox(0,0)[bl]{ 27.3\%}}
	 \put(256,283.0){\makebox(0,0)[bl]{ 25.3\%}}
	 \put(256,293.0){\makebox(0,0)[bl]{ 24.6\%}}
	 \put(256,303.0){\makebox(0,0)[bl]{ 22.4\%}}
	 \put(256,313.0){\makebox(0,0)[bl]{ 20.5\%}}
	 \put(256,323.0){\makebox(0,0)[bl]{ 20.7\%}}
	 \put(256,333.0){\makebox(0,0)[bl]{ 15.9\%}}
	 \put(256,343.0){\makebox(0,0)[bl]{ 18.0\%}}
	 \put(256,353.0){\makebox(0,0)[bl]{ 14.9\%}}
	 \put(256,363.0){\makebox(0,0)[bl]{ 12.7\%}}
	 \put(256,373.0){\makebox(0,0)[bl]{ 12.3\%}}
	 \put(256,383.0){\makebox(0,0)[bl]{ 11.1\%}}
	 \put(256,393.0){\makebox(0,0)[bl]{ 10.4\%}}
	 \put(295,233.0){\makebox(0,0)[bl]{ 56.5\%}}
	 \put(295,243.0){\makebox(0,0)[bl]{ 58.7\%}}
	 \put(295,253.0){\makebox(0,0)[bl]{ 56.2\%}}
	 \put(295,263.0){\makebox(0,0)[bl]{ 54.3\%}}
	 \put(295,273.0){\makebox(0,0)[bl]{ 50.6\%}}
	 \put(295,283.0){\makebox(0,0)[bl]{ 46.9\%}}
	 \put(295,293.0){\makebox(0,0)[bl]{ 47.7\%}}
	 \put(295,303.0){\makebox(0,0)[bl]{ 42.3\%}}
	 \put(295,313.0){\makebox(0,0)[bl]{ 40.2\%}}
	 \put(295,323.0){\makebox(0,0)[bl]{ 42.3\%}}
	 \put(295,333.0){\makebox(0,0)[bl]{ 39.0\%}}
	 \put(295,343.0){\makebox(0,0)[bl]{ 34.6\%}}
	 \put(295,353.0){\makebox(0,0)[bl]{ 33.5\%}}
	 \put(295,363.0){\makebox(0,0)[bl]{ 30.7\%}}
	 \put(295,373.0){\makebox(0,0)[bl]{ 28.4\%}}
	 \put(295,383.0){\makebox(0,0)[bl]{ 28.8\%}}
	 \put(295,393.0){\makebox(0,0)[bl]{ 32.0\%}}\
	 \put(203,3.0){\makebox(0,0)[bl]{ 20}}
	 \put(203,13.0){\makebox(0,0)[bl]{ 19}}
	 \put(203,23.0){\makebox(0,0)[bl]{ 18}}
	 \put(203,33.0){\makebox(0,0)[bl]{ 17}}
	 \put(203,43.0){\makebox(0,0)[bl]{ 16}}
	 \put(203,53.0){\makebox(0,0)[bl]{ 15}}
	 \put(203,63.0){\makebox(0,0)[bl]{ 14}}
	 \put(203,73.0){\makebox(0,0)[bl]{ 13}}
	 \put(203,83.0){\makebox(0,0)[bl]{ 12}}
	 \put(203,93.0){\makebox(0,0)[bl]{ 11}}
	 \put(203,103.0){\makebox(0,0)[bl]{ 10}}
	 \put(207,113.0){\makebox(0,0)[bl]{ 9}}
	 \put(207,123.0){\makebox(0,0)[bl]{ 8}}
	 \put(207,133.0){\makebox(0,0)[bl]{ 7}}
	 \put(207,143.0){\makebox(0,0)[bl]{ 6}}
	 \put(207,153.0){\makebox(0,0)[bl]{ 5}}
	 \put(207,163.0){\makebox(0,0)[bl]{ 4}}
	 \put(218,3.0){\makebox(0,0)[bl]{ 100.0\%}}
	 \put(218,13.0){\makebox(0,0)[bl]{ 100.0\%}}
	 \put(218,23.0){\makebox(0,0)[bl]{ 100.0\%}}
	 \put(218,33.0){\makebox(0,0)[bl]{ 100.0\%}}
	 \put(218,43.0){\makebox(0,0)[bl]{ 100.0\%}}
	 \put(218,53.0){\makebox(0,0)[bl]{ 100.0\%}}
	 \put(218,63.0){\makebox(0,0)[bl]{ 99.8\%}}
	 \put(218,73.0){\makebox(0,0)[bl]{ 100.0\%}}
	 \put(218,83.0){\makebox(0,0)[bl]{ 99.6\%}}
	 \put(218,93.0){\makebox(0,0)[bl]{ 99.8\%}}
	 \put(218,103.0){\makebox(0,0)[bl]{ 98.6\%}}
	 \put(218,113.0){\makebox(0,0)[bl]{ 97.0\%}}
	 \put(218,123.0){\makebox(0,0)[bl]{ 91.0\%}}
	 \put(218,133.0){\makebox(0,0)[bl]{ 83.4\%}}
	 \put(218,143.0){\makebox(0,0)[bl]{ 45.8\%}}
	 \put(218,153.0){\makebox(0,0)[bl]{ 42.2\%}}
	 \put(218,163.0){\makebox(0,0)[bl]{ 12.6\%}}
	 \put(256,3.0){\makebox(0,0)[bl]{ 99.7\%}}
	 \put(256,13.0){\makebox(0,0)[bl]{ 99.9\%}}
	 \put(256,23.0){\makebox(0,0)[bl]{ 99.8\%}}
	 \put(256,33.0){\makebox(0,0)[bl]{ 99.7\%}}
	 \put(256,43.0){\makebox(0,0)[bl]{ 99.5\%}}
	 \put(256,53.0){\makebox(0,0)[bl]{ 99.5\%}}
	 \put(256,63.0){\makebox(0,0)[bl]{ 99.5\%}}
	 \put(256,73.0){\makebox(0,0)[bl]{ 98.4\%}}
	 \put(256,83.0){\makebox(0,0)[bl]{ 98.3\%}}
	 \put(256,93.0){\makebox(0,0)[bl]{ 97.9\%}}
	 \put(256,103.0){\makebox(0,0)[bl]{ 97.2\%}}
	 \put(256,113.0){\makebox(0,0)[bl]{ 96.6\%}}
	 \put(256,123.0){\makebox(0,0)[bl]{ 94.5\%}}
	 \put(256,133.0){\makebox(0,0)[bl]{ 92.1\%}}
	 \put(256,143.0){\makebox(0,0)[bl]{ 86.7\%}}
	 \put(256,153.0){\makebox(0,0)[bl]{ 82.6\%}}
	 \put(256,163.0){\makebox(0,0)[bl]{ 75.9\%}}
	 \put(295,3.0){\makebox(0,0)[bl]{ 100.0\%}}
	 \put(295,13.0){\makebox(0,0)[bl]{ 100.0\%}}
	 \put(295,23.0){\makebox(0,0)[bl]{ 100.0\%}}
	 \put(295,33.0){\makebox(0,0)[bl]{ 100.0\%}}
	 \put(295,43.0){\makebox(0,0)[bl]{ 100.0\%}}
	 \put(295,53.0){\makebox(0,0)[bl]{ 100.0\%}}
	 \put(295,63.0){\makebox(0,0)[bl]{ 100.0\%}}
	 \put(295,73.0){\makebox(0,0)[bl]{ 100.0\%}}
	 \put(295,83.0){\makebox(0,0)[bl]{ 100.0\%}}
	 \put(295,93.0){\makebox(0,0)[bl]{ 100.0\%}}
	 \put(295,103.0){\makebox(0,0)[bl]{ 100.0\%}}
	 \put(295,113.0){\makebox(0,0)[bl]{ 99.9\%}}
	 \put(295,123.0){\makebox(0,0)[bl]{ 99.5\%}}
	 \put(295,133.0){\makebox(0,0)[bl]{ 99.1\%}}
	 \put(295,143.0){\makebox(0,0)[bl]{ 98.6\%}}
	 \put(295,153.0){\makebox(0,0)[bl]{ 95.9\%}}
	 \put(295,163.0){\makebox(0,0)[bl]{ 92.7\%}}
	 \color{mygray1}
	 \put(0,475.0){\makebox(0,0)[bl]{ QUESTION: (4a)$\ \sharp\ cause_{obj}$}}
	 \put(0,464.5){\makebox(0,0)[bl]{ ANSWER: ANSWER: (2a)}}
	 \put(0,453.5){\makebox(0,0)[bl]{ NUMBER OF TRIALS: 1000}}
	 \put(0,441.0){\makebox(0,0)[bl]{ GA CONSTRUCTION: Agent-object with odding blades, right-hand-side questions}}
	 \put(0,431.0){\makebox(0,0)[bl]{ MEANINGFUL/NOISY BLADES: 3/4}}
	 \put(4,422){\includegraphics{sq.jpg}}
	 \put(8,418.5){\makebox(0,0)[bl]{ GA, Hamming measure}}	 
	 \put(4.5,413){\circle{1.5}}
	 \put(8,410){\makebox(0,0)[bl]{ BSC}}
	 \put(4.5,404){\circle*{1}}
	 \put(8,400){\makebox(0,0)[bl]{ HRR}}	 
	 \color{black}
	 \put(0.0,230.0){\vector(1,0){180.0}}  
	 \put(0.0,230.0){\vector(0,1){120.0}} 
	 \put(10.0,232){\line(0,-1){4.0}} 
	 \put(20.0,232){\line(0,-1){4.0}} 
	 \put(30.0,232){\line(0,-1){4.0}} 
	 \put(40.0,232){\line(0,-1){4.0}} 
	 \put(50.0,232){\line(0,-1){4.0}} 
	 \put(60.0,232){\line(0,-1){4.0}} 
	 \put(70.0,232){\line(0,-1){4.0}} 
	 \put(80.0,232){\line(0,-1){4.0}} 
	 \put(90.0,232){\line(0,-1){4.0}} 
	 \put(100.0,232){\line(0,-1){4.0}} 
	 \put(110.0,232){\line(0,-1){4.0}} 
	 \put(120.0,232){\line(0,-1){4.0}} 
	 \put(130.0,232){\line(0,-1){4.0}} 
	 \put(140.0,232){\line(0,-1){4.0}} 
	 \put(150.0,232){\line(0,-1){4.0}} 
	 \put(160.0,232){\line(0,-1){4.0}} 
	 \put(170.0,232){\line(0,-1){4.0}} 
	 \put(-2.0,240.0){\line(1,0){4.0}} 
	 \put(-2.0,250.0){\line(1,0){4.0}} 
	 \put(-2.0,260.0){\line(1,0){4.0}} 
	 \put(-2.0,270.0){\line(1,0){4.0}} 
	 \put(-2.0,280.0){\line(1,0){4.0}} 
	 \put(-2.0,290.0){\line(1,0){4.0}} 
	 \put(-2.0,300.0){\line(1,0){4.0}} 
	 \put(-2.0,310.0){\line(1,0){4.0}} 
	 \put(-2.0,320.0){\line(1,0){4.0}} 
	 \put(-2.0,330.0){\line(1,0){4.0}} 
	 \put(15.0,221.0){\makebox(0,0)[bl]{ 5}}
	 \put(63.0,221.0){\makebox(0,0)[bl]{ 10}}
	 \put(113.0,221.0){\makebox(0,0)[bl]{ 15}}
	 \put(163.0,221.0){\makebox(0,0)[bl]{ 20}}
	 \put(180.0,226.0){\makebox(0,0)[bl]{ N}}
	 \put(-21,237.0){\makebox(0,0)[bl]{ 10\%}}
	 \put(-21,277.0){\makebox(0,0)[bl]{ 50\%}}
	 \put(-25,327.0){\makebox(0,0)[bl]{ 100\%}}
	 \put(-7.0,352.0){\makebox(0,0)[bl]{[\%] recognition}}
	 \color{mygray1}
	 \put(9,242.1){\includegraphics{sq.jpg}}  
	 \put(19,271.7){\includegraphics{sq.jpg}}
	 \put(29,275.3){\includegraphics{sq.jpg}}
	 \put(39,312.9){\includegraphics{sq.jpg}}
	 \put(49,320.5){\includegraphics{sq.jpg}}
	 \put(59,326.5){\includegraphics{sq.jpg}}
	 \put(69,328.1){\includegraphics{sq.jpg}}
	 \put(79,329.3){\includegraphics{sq.jpg}}
	 \put(89,329.1){\includegraphics{sq.jpg}}
	 \put(99,329.5){\includegraphics{sq.jpg}}
	 \put(109,329.3){\includegraphics{sq.jpg}}
	 \put(119,329.5){\includegraphics{sq.jpg}}
	 \put(129,329.5){\includegraphics{sq.jpg}}
	 \put(139,329.5){\includegraphics{sq.jpg}}
	 \put(149,329.5){\includegraphics{sq.jpg}}
	 \put(159,329.5){\includegraphics{sq.jpg}}
	 \put(169,329.5){\includegraphics{sq.jpg}}
	 \color{black}
	 \put(10.0,262.0){\circle{1.5}}  
	 \put(20.0,258.8){\circle{1.5}}
	 \put(30.0,258.4){\circle{1.5}}
	 \put(40.0,260.7){\circle{1.5}}
	 \put(50.0,263.5){\circle{1.5}}
	 \put(60.0,264.6){\circle{1.5}}
	 \put(70.0,269.0){\circle{1.5}}
	 \put(80.0,272.3){\circle{1.5}}
	 \put(90.0,270.2){\circle{1.5}}
	 \put(100.0,272.3){\circle{1.5}}
	 \put(110.0,277.7){\circle{1.5}}
	 \put(120.0,276.9){\circle{1.5}}
	 \put(130.0,280.6){\circle{1.5}}
	 \put(140.0,284.3){\circle{1.5}}
	 \put(150.0,286.2){\circle{1.5}}
	 \put(160.0,288.7){\circle{1.5}}
	 \put(170.0,286.5){\circle{1.5}}
	 \put(10.0,240.4){\circle*{1}}  
	 \put(20.0,241.1){\circle*{1}}
	 \put(30.0,242.3){\circle*{1}}
	 \put(40.0,242.7){\circle*{1}}
	 \put(50.0,244.9){\circle*{1}}
	 \put(60.0,248.0){\circle*{1}}
	 \put(70.0,245.9){\circle*{1}}
	 \put(80.0,250.7){\circle*{1}}
	 \put(90.0,250.5){\circle*{1}}
	 \put(100.0,252.4){\circle*{1}}
	 \put(110.0,254.6){\circle*{1}}
	 \put(120.0,255.3){\circle*{1}}
	 \put(130.0,257.3){\circle*{1}}
	 \put(140.0,260.9){\circle*{1}}
	 \put(150.0,262.2){\circle*{1}}
	 \put(160.0,262.9){\circle*{1}}
	 \put(170.0,264.6){\circle*{1}}
	 \put(0.0,0.0){\vector(1,0){180.0}}  
	 \put(0.0,0.0){\vector(0,1){120.0}} 
	 \put(10.0,2.0){\line(0,-1){4.0}} 
	 \put(20.0,2.0){\line(0,-1){4.0}} 
	 \put(30.0,2.0){\line(0,-1){4.0}} 
	 \put(40.0,2.0){\line(0,-1){4.0}} 
	 \put(50.0,2.0){\line(0,-1){4.0}} 
	 \put(60.0,2.0){\line(0,-1){4.0}} 
	 \put(70.0,2.0){\line(0,-1){4.0}} 
	 \put(80.0,2.0){\line(0,-1){4.0}} 
	 \put(90.0,2.0){\line(0,-1){4.0}} 
	 \put(100.0,2.0){\line(0,-1){4.0}} 
	 \put(110.0,2.0){\line(0,-1){4.0}} 
	 \put(120.0,2.0){\line(0,-1){4.0}} 
	 \put(130.0,2.0){\line(0,-1){4.0}} 
	 \put(140.0,2.0){\line(0,-1){4.0}} 
	 \put(150.0,2.0){\line(0,-1){4.0}} 
	 \put(160.0,2.0){\line(0,-1){4.0}} 
	 \put(170.0,2.0){\line(0,-1){4.0}} 
	 \put(-2.0,10.0){\line(1,0){4.0}} 
	 \put(-2.0,20.0){\line(1,0){4.0}} 
	 \put(-2.0,30.0){\line(1,0){4.0}} 
	 \put(-2.0,40.0){\line(1,0){4.0}} 
	 \put(-2.0,50.0){\line(1,0){4.0}} 
	 \put(-2.0,60.0){\line(1,0){4.0}} 
	 \put(-2.0,70.0){\line(1,0){4.0}} 
	 \put(-2.0,80.0){\line(1,0){4.0}} 
	 \put(-2.0,90.0){\line(1,0){4.0}} 
	 \put(-2.0,100.0){\line(1,0){4.0}} 
	 \put(15.0,-9.0){\makebox(0,0)[bl]{ 5}}
	 \put(63.0,-9.0){\makebox(0,0)[bl]{ 10}}
	 \put(113.0,-9.0){\makebox(0,0)[bl]{ 15}}
	 \put(163.0,-9.0){\makebox(0,0)[bl]{ 20}}
	 \put(180.0,-4.0){\makebox(0,0)[bl]{ N}}
	 \put(-21,7.0){\makebox(0,0)[bl]{ 10\%}}
	 \put(-21,47.0){\makebox(0,0)[bl]{ 50\%}}
	 \put(-25,97.0){\makebox(0,0)[bl]{ 100\%}}
	 \put(-7.0,122.0){\makebox(0,0)[bl]{[\%] recognition}}
	 \color{mygray1}
	 \put(9,12.1){\includegraphics{sq.jpg}}  
	 \put(19,41.7){\includegraphics{sq.jpg}}
	 \put(29,45.3){\includegraphics{sq.jpg}}
	 \put(39,82.9){\includegraphics{sq.jpg}}
	 \put(49,90.5){\includegraphics{sq.jpg}}
	 \put(59,96.5){\includegraphics{sq.jpg}}
	 \put(69,98.1){\includegraphics{sq.jpg}}
	 \put(79,99.3){\includegraphics{sq.jpg}}
	 \put(89,99.1){\includegraphics{sq.jpg}}
	 \put(99,99.5){\includegraphics{sq.jpg}}
	 \put(109,99.3){\includegraphics{sq.jpg}}
	 \put(119,99.5){\includegraphics{sq.jpg}}
	 \put(129,99.5){\includegraphics{sq.jpg}}
	 \put(139,99.5){\includegraphics{sq.jpg}}
	 \put(149,99.5){\includegraphics{sq.jpg}}
	 \put(159,99.5){\includegraphics{sq.jpg}}
	 \put(169,99.5){\includegraphics{sq.jpg}}
	 \color{black}
	 \put(10.0,92.7){\circle{1.5}} 
	 \put(20.0,95.9){\circle{1.5}}
	 \put(30.0,98.6){\circle{1.5}}
	 \put(40.0,99.1){\circle{1.5}}
	 \put(50.0,99.5){\circle{1.5}}
	 \put(60.0,99.9){\circle{1.5}}
	 \put(70.0,100.0){\circle{1.5}}
	 \put(80.0,100.0){\circle{1.5}}
	 \put(90.0,100.0){\circle{1.5}}
	 \put(100.0,100.0){\circle{1.5}}
	 \put(110.0,100.0){\circle{1.5}}
	 \put(120.0,100.0){\circle{1.5}}
	 \put(130.0,100.0){\circle{1.5}}
	 \put(140.0,100.0){\circle{1.5}}
	 \put(150.0,100.0){\circle{1.5}}
	 \put(160.0,100.0){\circle{1.5}}
	 \put(170.0,100.0){\circle{1.5}}
	 \put(10.0,75.9){\circle*{1}} 
	 \put(20.0,82.6){\circle*{1}}
	 \put(30.0,86.7){\circle*{1}}
	 \put(40.0,92.1){\circle*{1}}
	 \put(50.0,94.5){\circle*{1}}
	 \put(60.0,96.6){\circle*{1}}
	 \put(70.0,97.2){\circle*{1}}
	 \put(80.0,97.9){\circle*{1}}
	 \put(90.0,98.3){\circle*{1}}
	 \put(100.0,98.4){\circle*{1}}
	 \put(110.0,99.5){\circle*{1}}
	 \put(120.0,99.5){\circle*{1}}
	 \put(130.0,99.5){\circle*{1}}
	 \put(140.0,99.7){\circle*{1}}
	 \put(150.0,99.8){\circle*{1}}
	 \put(160.0,99.9){\circle*{1}}
	 \put(170.0,99.7){\circle*{1}}
    \put(0,136){\makebox(0,0)[bl]{{Vector lengths: $N$ for GA, $13N$ for HRR and BSC.}}}
	 \put(0,366){\makebox(0,0)[bl]{{All vector lengths: $N$.}}}	 	 
  \end{picture}
  \caption{Comparison of recognition for GA, BSC and HRR -- (4a)$\ \sharp\ cause_{obj}$.}
  \label{fig:compF}   
\end{figure}

\definecolor{mygray1}{rgb}{0.5,0.5,0.5}
\definecolor{mygreen1}{rgb}{0,0.8,0}
\begin{figure}[ht!]		
  \begin{picture}(0,500)(0,-5)
	 \put(200.0,410.0){\line(1,0){129.0}}  
	 \put(200.0,400.0){\line(1,0){129.0}}  	 
	 \put(200.0,390.0){\line(1,0){129.0}}  
	 \put(200.0,380.0){\line(1,0){129.0}}  
	 \put(200.0,370.0){\line(1,0){129.0}}  
	 \put(200.0,360.0){\line(1,0){129.0}}  
	 \put(200.0,350.0){\line(1,0){129.0}}  
	 \put(200.0,340.0){\line(1,0){129.0}}  
	 \put(200.0,330.0){\line(1,0){129.0}}  
	 \put(200.0,320.0){\line(1,0){129.0}}  
	 \put(200.0,310.0){\line(1,0){129.0}}  
	 \put(200.0,300.0){\line(1,0){129.0}}  
	 \put(200.0,290.0){\line(1,0){129.0}}  
	 \put(200.0,280.0){\line(1,0){129.0}}  
	 \put(200.0,270.0){\line(1,0){129.0}}  
	 \put(200.0,260.0){\line(1,0){129.0}}  
	 \put(200.0,250.0){\line(1,0){129.0}}  
	 \put(200.0,240.0){\line(1,0){129.0}}  
	 \put(200.0,230.0){\line(1,0){129.0}}  
	 \put(200.0,180.0){\line(1,0){129.0}}  
	 \put(200.0,170.0){\line(1,0){129.0}}  
	 \put(200.0,160.0){\line(1,0){129.0}}  
	 \put(200.0,150.0){\line(1,0){129.0}}  
	 \put(200.0,140.0){\line(1,0){129.0}}  
	 \put(200.0,130.0){\line(1,0){129.0}}  
	 \put(200.0,120.0){\line(1,0){129.0}}  
	 \put(200.0,110.0){\line(1,0){129.0}}  
	 \put(200.0,100.0){\line(1,0){129.0}}  
	 \put(200.0,90.0){\line(1,0){129.0}}  
	 \put(200.0,80.0){\line(1,0){129.0}}  
	 \put(200.0,70.0){\line(1,0){129.0}}  
	 \put(200.0,60.0){\line(1,0){129.0}}  
	 \put(200.0,50.0){\line(1,0){129.0}}  
	 \put(200.0,40.0){\line(1,0){129.0}}  
	 \put(200.0,30.0){\line(1,0){129.0}}  
	 \put(200.0,20.0){\line(1,0){129.0}}  
	 \put(200.0,10.0){\line(1,0){129.0}}  
	 \put(200.0,0.0){\line(1,0){129.0}}  
	 \put(200.0,230.0){\line(0,1){180.0}}  
	 \put(215.0,230.0){\line(0,1){180.0}}  
	 \put(253.0,230.0){\line(0,1){180.0}}  
	 \put(292.0,230.0){\line(0,1){180.0}}  
	 \put(329.0,230.0){\line(0,1){180.0}}  
	 
	 \put(200.0,0.0){\line(0,1){180.0}}  
	 \put(215.0,0.0){\line(0,1){180.0}}  
	 \put(253.0,0.0){\line(0,1){180.0}}  
	 \put(292.0,0.0){\line(0,1){180.0}}  
	 \put(329.0,0.0){\line(0,1){180.0}}  
	 \put(203,402.5){\makebox(0,0)[bl]{ N}}
	 \put(217,401.0){\makebox(0,0)[bl]{{Hamming}}}
	 \put(264,403.0){\makebox(0,0)[bl]{ HRR}}
	 \put(300,403.0){\makebox(0,0)[bl]{ BSC}}	 

	 \put(203,172.5){\makebox(0,0)[bl]{ N}}
	 \put(217,171.0){\makebox(0,0)[bl]{{Hamming}}}
	 \put(264,173.0){\makebox(0,0)[bl]{ HRR}}
	 \put(300,173.0){\makebox(0,0)[bl]{ BSC}}
	 \put(203,233.0){\makebox(0,0)[bl]{ 20}}
	 \put(203,243.0){\makebox(0,0)[bl]{ 19}}
	 \put(203,253.0){\makebox(0,0)[bl]{ 18}}
	 \put(203,263.0){\makebox(0,0)[bl]{ 17}}
	 \put(203,273.0){\makebox(0,0)[bl]{ 16}}
	 \put(203,283.0){\makebox(0,0)[bl]{ 15}}
	 \put(203,293.0){\makebox(0,0)[bl]{ 14}}
	 \put(203,303.0){\makebox(0,0)[bl]{ 13}}
	 \put(203,313.0){\makebox(0,0)[bl]{ 12}}
	 \put(203,323.0){\makebox(0,0)[bl]{ 11}}
	 \put(203,333.0){\makebox(0,0)[bl]{ 10}}
	 \put(207,343.0){\makebox(0,0)[bl]{ 9}}
	 \put(207,353.0){\makebox(0,0)[bl]{ 8}}
	 \put(207,363.0){\makebox(0,0)[bl]{ 7}}
	 \put(207,373.0){\makebox(0,0)[bl]{ 6}}
	 \put(207,383.0){\makebox(0,0)[bl]{ 5}}
	 \put(207,393.0){\makebox(0,0)[bl]{ 4}}
	 \put(218,233.0){\makebox(0,0)[bl]{ 100.0\%}}
	 \put(218,243.0){\makebox(0,0)[bl]{ 100.0\%}}
	 \put(218,253.0){\makebox(0,0)[bl]{ 100.0\%}}
	 \put(218,263.0){\makebox(0,0)[bl]{ 100.0\%}}
	 \put(218,273.0){\makebox(0,0)[bl]{ 100.0\%}}
	 \put(218,283.0){\makebox(0,0)[bl]{ 100.0\%}}
	 \put(218,293.0){\makebox(0,0)[bl]{ 100.0\%}}
	 \put(218,303.0){\makebox(0,0)[bl]{ 100.0\%}}
	 \put(218,313.0){\makebox(0,0)[bl]{ 100.0\%}}
	 \put(218,323.0){\makebox(0,0)[bl]{ 99.6\%}}
	 \put(218,333.0){\makebox(0,0)[bl]{ 99.0\%}}
	 \put(218,343.0){\makebox(0,0)[bl]{ 99.4\%}}
	 \put(218,353.0){\makebox(0,0)[bl]{ 98.0\%}}
	 \put(218,363.0){\makebox(0,0)[bl]{ 96.2\%}}
	 \put(218,373.0){\makebox(0,0)[bl]{ 90.0\%}}
	 \put(218,383.0){\makebox(0,0)[bl]{ 89.4\%}}
	 \put(218,393.0){\makebox(0,0)[bl]{ 55.2\%}}
	 \put(256,233.0){\makebox(0,0)[bl]{ 26.6\%}}
	 \put(256,243.0){\makebox(0,0)[bl]{ 29.0\%}}
	 \put(256,253.0){\makebox(0,0)[bl]{ 27.3\%}}
	 \put(256,263.0){\makebox(0,0)[bl]{ 22.3\%}}
	 \put(256,273.0){\makebox(0,0)[bl]{ 22.2\%}}
	 \put(256,283.0){\makebox(0,0)[bl]{ 18.8\%}}
	 \put(256,293.0){\makebox(0,0)[bl]{ 19.7\%}}
	 \put(256,303.0){\makebox(0,0)[bl]{ 21.1\%}}
	 \put(256,313.0){\makebox(0,0)[bl]{ 17.5\%}}
	 \put(256,323.0){\makebox(0,0)[bl]{ 16.6\%}}
	 \put(256,333.0){\makebox(0,0)[bl]{ 15.6\%}}
	 \put(256,343.0){\makebox(0,0)[bl]{ 13.4\%}}
	 \put(256,353.0){\makebox(0,0)[bl]{ 13.0\%}}
	 \put(256,363.0){\makebox(0,0)[bl]{ 10.2\%}}
	 \put(256,373.0){\makebox(0,0)[bl]{ 8.6\%}}
	 \put(256,383.0){\makebox(0,0)[bl]{ 8.4\%}}
	 \put(256,393.0){\makebox(0,0)[bl]{ 7.6\%}}
	 \put(295,233.0){\makebox(0,0)[bl]{ 56.4\%}}
	 \put(295,243.0){\makebox(0,0)[bl]{ 59.7\%}}
	 \put(295,253.0){\makebox(0,0)[bl]{ 53.6\%}}
	 \put(295,263.0){\makebox(0,0)[bl]{ 53.8\%}}
	 \put(295,273.0){\makebox(0,0)[bl]{ 49.9\%}}
	 \put(295,283.0){\makebox(0,0)[bl]{ 46.5\%}}
	 \put(295,293.0){\makebox(0,0)[bl]{ 44.8\%}}
	 \put(295,303.0){\makebox(0,0)[bl]{ 46.0\%}}
	 \put(295,313.0){\makebox(0,0)[bl]{ 42.6\%}}
	 \put(295,323.0){\makebox(0,0)[bl]{ 40.4\%}}
	 \put(295,333.0){\makebox(0,0)[bl]{ 35.3\%}}
	 \put(295,343.0){\makebox(0,0)[bl]{ 35.1\%}}
	 \put(295,353.0){\makebox(0,0)[bl]{ 32.9\%}}
	 \put(295,363.0){\makebox(0,0)[bl]{ 31.0\%}}
	 \put(295,373.0){\makebox(0,0)[bl]{ 32.5\%}}
	 \put(295,383.0){\makebox(0,0)[bl]{ 26.1\%}}
	 \put(295,393.0){\makebox(0,0)[bl]{ 32.9\%}}
	 \put(203,3.0){\makebox(0,0)[bl]{ 20}}
	 \put(203,13.0){\makebox(0,0)[bl]{ 19}}
	 \put(203,23.0){\makebox(0,0)[bl]{ 18}}
	 \put(203,33.0){\makebox(0,0)[bl]{ 17}}
	 \put(203,43.0){\makebox(0,0)[bl]{ 16}}
	 \put(203,53.0){\makebox(0,0)[bl]{ 15}}
	 \put(203,63.0){\makebox(0,0)[bl]{ 14}}
	 \put(203,73.0){\makebox(0,0)[bl]{ 13}}
	 \put(203,83.0){\makebox(0,0)[bl]{ 12}}
	 \put(203,93.0){\makebox(0,0)[bl]{ 11}}
	 \put(203,103.0){\makebox(0,0)[bl]{ 10}}
	 \put(207,113.0){\makebox(0,0)[bl]{ 9}}
	 \put(207,123.0){\makebox(0,0)[bl]{ 8}}
	 \put(207,133.0){\makebox(0,0)[bl]{ 7}}
	 \put(207,143.0){\makebox(0,0)[bl]{ 6}}
	 \put(207,153.0){\makebox(0,0)[bl]{ 5}}
	 \put(207,163.0){\makebox(0,0)[bl]{ 4}}
	 \put(218,3.0){\makebox(0,0)[bl]{ 100.0\%}}
	 \put(218,13.0){\makebox(0,0)[bl]{ 100.0\%}}
	 \put(218,23.0){\makebox(0,0)[bl]{ 100.0\%}}
	 \put(218,33.0){\makebox(0,0)[bl]{ 100.0\%}}
	 \put(218,43.0){\makebox(0,0)[bl]{ 100.0\%}}
	 \put(218,53.0){\makebox(0,0)[bl]{ 100.0\%}}
	 \put(218,63.0){\makebox(0,0)[bl]{ 100.0\%}}
	 \put(218,73.0){\makebox(0,0)[bl]{ 100.0\%}}
	 \put(218,83.0){\makebox(0,0)[bl]{ 100.0\%}}
	 \put(218,93.0){\makebox(0,0)[bl]{ 99.6\%}}
	 \put(218,103.0){\makebox(0,0)[bl]{ 99.0\%}}
	 \put(218,113.0){\makebox(0,0)[bl]{ 99.4\%}}
	 \put(218,123.0){\makebox(0,0)[bl]{ 98.0\%}}
	 \put(218,133.0){\makebox(0,0)[bl]{ 96.2\%}}
	 \put(218,143.0){\makebox(0,0)[bl]{ 90.0\%}}
	 \put(218,153.0){\makebox(0,0)[bl]{ 89.4\%}}
	 \put(218,163.0){\makebox(0,0)[bl]{ 55.2\%}}
	 \put(256,3.0){\makebox(0,0)[bl]{ 95.5\%}}
	 \put(256,13.0){\makebox(0,0)[bl]{ 95.8\%}}
	 \put(256,23.0){\makebox(0,0)[bl]{ 94.4\%}}
	 \put(256,33.0){\makebox(0,0)[bl]{ 95.4\%}}
	 \put(256,43.0){\makebox(0,0)[bl]{ 94.1\%}}
	 \put(256,53.0){\makebox(0,0)[bl]{ 93.3\%}}
	 \put(256,63.0){\makebox(0,0)[bl]{ 91.8\%}}
	 \put(256,73.0){\makebox(0,0)[bl]{ 91.5\%}}
	 \put(256,83.0){\makebox(0,0)[bl]{ 89.7\%}}
	 \put(256,93.0){\makebox(0,0)[bl]{ 91.6\%}}
	 \put(256,103.0){\makebox(0,0)[bl]{ 86.9\%}}
	 \put(256,113.0){\makebox(0,0)[bl]{ 85.9\%}}
	 \put(256,123.0){\makebox(0,0)[bl]{ 81.5\%}}
	 \put(256,133.0){\makebox(0,0)[bl]{ 81.6\%}}
	 \put(256,143.0){\makebox(0,0)[bl]{ 77.6\%}}
	 \put(256,153.0){\makebox(0,0)[bl]{ 74.0\%}}
	 \put(256,163.0){\makebox(0,0)[bl]{ 65.3\%}}
	 \put(295,3.0){\makebox(0,0)[bl]{ 100.0\%}}
	 \put(295,13.0){\makebox(0,0)[bl]{ 100.0\%}}
	 \put(295,23.0){\makebox(0,0)[bl]{ 100.0\%}}
	 \put(295,33.0){\makebox(0,0)[bl]{ 100.0\%}}
	 \put(295,43.0){\makebox(0,0)[bl]{ 100.0\%}}
	 \put(295,53.0){\makebox(0,0)[bl]{ 100.0\%}}
	 \put(295,63.0){\makebox(0,0)[bl]{ 100.0\%}}
	 \put(295,73.0){\makebox(0,0)[bl]{ 100.0\%}}
	 \put(295,83.0){\makebox(0,0)[bl]{ 100.0\%}}
	 \put(295,93.0){\makebox(0,0)[bl]{ 100.0\%}}
	 \put(295,103.0){\makebox(0,0)[bl]{ 100.0\%}}
	 \put(295,113.0){\makebox(0,0)[bl]{ 99.9\%}}
	 \put(295,123.0){\makebox(0,0)[bl]{ 100.0\%}}
	 \put(295,133.0){\makebox(0,0)[bl]{ 99.3\%}}
	 \put(295,143.0){\makebox(0,0)[bl]{ 98.0\%}}
	 \put(295,153.0){\makebox(0,0)[bl]{ 96.8\%}}
	 \put(295,163.0){\makebox(0,0)[bl]{ 93.0\%}}
	 \color{mygray1}
	 \put(0,475.0){\makebox(0,0)[bl]{ QUESTION: (5a)$\ \sharp\ see_{obj}$}}
	 \put(0,464.5){\makebox(0,0)[bl]{ ANSWER: (4a)}}
	 \put(0,453.5){\makebox(0,0)[bl]{ NUMBER OF TRIALS: 1000}}
	 \put(0,441.0){\makebox(0,0)[bl]{ GA CONSTRUCTION: Agent-object with odding blades, right-hand-side questions}}
	 \put(0,431.0){\makebox(0,0)[bl]{ MEANINGFUL/NOISY BLADES: 7/2}}
	 \put(4,422){\includegraphics{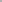}}
	 \put(8,418.5){\makebox(0,0)[bl]{ GA, Hamming measure}}	 
	 \put(4.5,413){\circle{1.5}}
	 \put(8,410){\makebox(0,0)[bl]{ BSC}}
	 \put(4.5,404){\circle*{1}}
	 \put(8,400){\makebox(0,0)[bl]{ HRR}}	 
	 \color{black}
	 \put(0.0,230.0){\vector(1,0){180.0}}  
	 \put(0.0,230.0){\vector(0,1){120.0}} 
	 \put(10.0,232){\line(0,-1){4.0}} 
	 \put(20.0,232){\line(0,-1){4.0}} 
	 \put(30.0,232){\line(0,-1){4.0}} 
	 \put(40.0,232){\line(0,-1){4.0}} 
	 \put(50.0,232){\line(0,-1){4.0}} 
	 \put(60.0,232){\line(0,-1){4.0}} 
	 \put(70.0,232){\line(0,-1){4.0}} 
	 \put(80.0,232){\line(0,-1){4.0}} 
	 \put(90.0,232){\line(0,-1){4.0}} 
	 \put(100.0,232){\line(0,-1){4.0}} 
	 \put(110.0,232){\line(0,-1){4.0}} 
	 \put(120.0,232){\line(0,-1){4.0}} 
	 \put(130.0,232){\line(0,-1){4.0}} 
	 \put(140.0,232){\line(0,-1){4.0}} 
	 \put(150.0,232){\line(0,-1){4.0}} 
	 \put(160.0,232){\line(0,-1){4.0}} 
	 \put(170.0,232){\line(0,-1){4.0}} 
	 \put(-2.0,240.0){\line(1,0){4.0}} 
	 \put(-2.0,250.0){\line(1,0){4.0}} 
	 \put(-2.0,260.0){\line(1,0){4.0}} 
	 \put(-2.0,270.0){\line(1,0){4.0}} 
	 \put(-2.0,280.0){\line(1,0){4.0}} 
	 \put(-2.0,290.0){\line(1,0){4.0}} 
	 \put(-2.0,300.0){\line(1,0){4.0}} 
	 \put(-2.0,310.0){\line(1,0){4.0}} 
	 \put(-2.0,320.0){\line(1,0){4.0}} 
	 \put(-2.0,330.0){\line(1,0){4.0}} 
	 \put(15.0,221.0){\makebox(0,0)[bl]{ 5}}
	 \put(63.0,221.0){\makebox(0,0)[bl]{ 10}}
	 \put(113.0,221.0){\makebox(0,0)[bl]{ 15}}
	 \put(163.0,221.0){\makebox(0,0)[bl]{ 20}}
	 \put(180.0,226.0){\makebox(0,0)[bl]{ N}}
	 \put(-21,237.0){\makebox(0,0)[bl]{ 10\%}}
	 \put(-21,277.0){\makebox(0,0)[bl]{ 50\%}}
	 \put(-25,327.0){\makebox(0,0)[bl]{ 100\%}}
	 \put(-7.0,352.0){\makebox(0,0)[bl]{[\%] recognition}}
	 \color{mygray1}
	 \put(9,284.7){\includegraphics{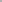}}  
	 \put(19,318.9){\includegraphics{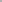}}
	 \put(29,319.5){\includegraphics{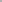}}
	 \put(39,325.7){\includegraphics{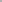}}
	 \put(49,327.5){\includegraphics{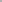}}
	 \put(59,328.9){\includegraphics{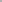}}
	 \put(69,328.5){\includegraphics{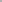}}
	 \put(79,329.1){\includegraphics{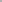}}
	 \put(89,329.5){\includegraphics{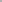}}
	 \put(99,329.5){\includegraphics{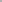}}
	 \put(109,329.5){\includegraphics{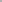}}
	 \put(119,329.5){\includegraphics{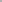}}
	 \put(129,329.5){\includegraphics{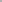}}
	 \put(139,329.5){\includegraphics{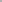}}
	 \put(149,329.5){\includegraphics{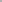}}
	 \put(159,329.5){\includegraphics{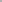}}
	 \put(169,329.5){\includegraphics{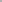}}
	 \color{black}
	 \put(10.0,262.9){\circle{1.5}}  
	 \put(20.0,256.1){\circle{1.5}}
	 \put(30.0,262.5){\circle{1.5}}
	 \put(40.0,261.0){\circle{1.5}}
	 \put(50.0,262.9){\circle{1.5}}
	 \put(60.0,265.1){\circle{1.5}}
	 \put(70.0,265.3){\circle{1.5}}
	 \put(80.0,270.4){\circle{1.5}}
	 \put(90.0,272.6){\circle{1.5}}
	 \put(100.0,276.0){\circle{1.5}}
	 \put(110.0,274.8){\circle{1.5}}
	 \put(120.0,276.5){\circle{1.5}}
	 \put(130.0,279.9){\circle{1.5}}
	 \put(140.0,283.8){\circle{1.5}}
	 \put(150.0,283.6){\circle{1.5}}
	 \put(160.0,289.7){\circle{1.5}}
	 \put(170.0,286.4){\circle{1.5}}
	 \put(10.0,237.6){\circle*{1}}  
	 \put(20.0,238.4){\circle*{1}}
	 \put(30.0,238.6){\circle*{1}}
	 \put(40.0,240.2){\circle*{1}}
	 \put(50.0,243.0){\circle*{1}}
	 \put(60.0,243.4){\circle*{1}}
	 \put(70.0,245.6){\circle*{1}}
	 \put(80.0,246.6){\circle*{1}}
	 \put(90.0,247.5){\circle*{1}}
	 \put(100.0,251.1){\circle*{1}}
	 \put(110.0,249.7){\circle*{1}}
	 \put(120.0,248.8){\circle*{1}}
	 \put(130.0,252.2){\circle*{1}}
	 \put(140.0,252.3){\circle*{1}}
	 \put(150.0,257.3){\circle*{1}}
	 \put(160.0,259.0){\circle*{1}}
	 \put(170.0,256.6){\circle*{1}}
	 \put(0.0,0.0){\vector(1,0){180.0}}  
	 \put(0.0,0.0){\vector(0,1){120.0}} 
	 \put(10.0,2.0){\line(0,-1){4.0}} 
	 \put(20.0,2.0){\line(0,-1){4.0}} 
	 \put(30.0,2.0){\line(0,-1){4.0}} 
	 \put(40.0,2.0){\line(0,-1){4.0}} 
	 \put(50.0,2.0){\line(0,-1){4.0}} 
	 \put(60.0,2.0){\line(0,-1){4.0}} 
	 \put(70.0,2.0){\line(0,-1){4.0}} 
	 \put(80.0,2.0){\line(0,-1){4.0}} 
	 \put(90.0,2.0){\line(0,-1){4.0}} 
	 \put(100.0,2.0){\line(0,-1){4.0}} 
	 \put(110.0,2.0){\line(0,-1){4.0}} 
	 \put(120.0,2.0){\line(0,-1){4.0}} 
	 \put(130.0,2.0){\line(0,-1){4.0}} 
	 \put(140.0,2.0){\line(0,-1){4.0}} 
	 \put(150.0,2.0){\line(0,-1){4.0}} 
	 \put(160.0,2.0){\line(0,-1){4.0}} 
	 \put(170.0,2.0){\line(0,-1){4.0}} 
	 \put(-2.0,10.0){\line(1,0){4.0}} 
	 \put(-2.0,20.0){\line(1,0){4.0}} 
	 \put(-2.0,30.0){\line(1,0){4.0}} 
	 \put(-2.0,40.0){\line(1,0){4.0}} 
	 \put(-2.0,50.0){\line(1,0){4.0}} 
	 \put(-2.0,60.0){\line(1,0){4.0}} 
	 \put(-2.0,70.0){\line(1,0){4.0}} 
	 \put(-2.0,80.0){\line(1,0){4.0}} 
	 \put(-2.0,90.0){\line(1,0){4.0}} 
	 \put(-2.0,100.0){\line(1,0){4.0}} 
	 \put(15.0,-9.0){\makebox(0,0)[bl]{ 5}}
	 \put(63.0,-9.0){\makebox(0,0)[bl]{ 10}}
	 \put(113.0,-9.0){\makebox(0,0)[bl]{ 15}}
	 \put(163.0,-9.0){\makebox(0,0)[bl]{ 20}}
	 \put(180.0,-4.0){\makebox(0,0)[bl]{ N}}
	 \put(-21,7.0){\makebox(0,0)[bl]{ 10\%}}
	 \put(-21,47.0){\makebox(0,0)[bl]{ 50\%}}
	 \put(-25,97.0){\makebox(0,0)[bl]{ 100\%}}
	 \put(-7.0,122.0){\makebox(0,0)[bl]{[\%] recognition}}
	 \color{mygray1}
	 \put(9,54.7){\includegraphics{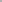}}  
	 \put(19,88.9){\includegraphics{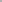}}
	 \put(29,89.5){\includegraphics{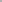}}
	 \put(39,95.7){\includegraphics{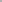}}
	 \put(49,97.5){\includegraphics{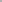}}
	 \put(59,98.9){\includegraphics{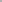}}
	 \put(69,98.5){\includegraphics{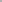}}
	 \put(79,99.1){\includegraphics{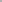}}
	 \put(89,99.5){\includegraphics{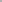}}
	 \put(99,99.5){\includegraphics{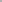}}
	 \put(109,99.5){\includegraphics{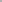}}
	 \put(119,99.5){\includegraphics{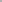}}
	 \put(129,99.5){\includegraphics{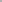}}
	 \put(139,99.5){\includegraphics{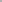}}
	 \put(149,99.5){\includegraphics{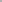}}
	 \put(159,99.5){\includegraphics{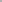}}
	 \put(169,99.5){\includegraphics{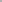}}
	 \color{black}
	 \put(10.0,93.0){\circle{1.5}} 
	 \put(20.0,96.8){\circle{1.5}}
	 \put(30.0,98.0){\circle{1.5}}
	 \put(40.0,99.3){\circle{1.5}}
	 \put(50.0,100.0){\circle{1.5}}
	 \put(60.0,99.9){\circle{1.5}}
	 \put(70.0,100.0){\circle{1.5}}
	 \put(80.0,100.0){\circle{1.5}}
	 \put(90.0,100.0){\circle{1.5}}
	 \put(100.0,100.0){\circle{1.5}}
	 \put(110.0,100.0){\circle{1.5}}
	 \put(120.0,100.0){\circle{1.5}}
	 \put(130.0,100.0){\circle{1.5}}
	 \put(140.0,100.0){\circle{1.5}}
	 \put(150.0,100.0){\circle{1.5}}
	 \put(160.0,100.0){\circle{1.5}}
	 \put(170.0,100.0){\circle{1.5}}
	 \put(10.0,65.3){\circle*{1}} 
	 \put(20.0,74.0){\circle*{1}}
	 \put(30.0,77.6){\circle*{1}}
	 \put(40.0,81.6){\circle*{1}}
	 \put(50.0,81.5){\circle*{1}}
	 \put(60.0,85.9){\circle*{1}}
	 \put(70.0,86.9){\circle*{1}}
	 \put(80.0,91.6){\circle*{1}}
	 \put(90.0,89.7){\circle*{1}}
	 \put(100.0,91.5){\circle*{1}}
	 \put(110.0,91.8){\circle*{1}}
	 \put(120.0,93.3){\circle*{1}}
	 \put(130.0,94.1){\circle*{1}}
	 \put(140.0,95.4){\circle*{1}}
	 \put(150.0,94.4){\circle*{1}}
	 \put(160.0,95.8){\circle*{1}}
	 \put(170.0,95.5){\circle*{1}}

    \put(0,136){\makebox(0,0)[bl]{{Vector lengths: $N$ for GA, $13N$ for HRR and BSC.}}}
	 \put(0,366){\makebox(0,0)[bl]{{All vector lengths: $N$.}}}	 	 
  \end{picture}
  \caption{Comparison of recognition for GA, BSC and HRR -- (5a)$\ \sharp\ see_{obj}$.}
  \label{fig:compG}   
\end{figure}

\section{Conclusion}
\label{sec:7}

We have presented a new model of distributed representation that is based on the way humans
think, while models developed so far were designed to use arrays of numbers mainly in order to be easily simulated by computers.

After a brief recollection of the main ideas behind the GA model, we investigated three types of 
sentence constructions, namely the Plate construction, the agent-object construction and the agent-object construction with odding blades. Two methods of asking questions were also investigated. As a result, in face of shortcomings of recognition based solely on the inner product, matrix representation has been employed as a recognition tool for the GA model. Using test results computed on a toy model, we have shown that Hamming and Euclidean measures of similarity perform very well under the agent-object construction with odding blades.

We also studied the ways in which the number of potential answers is affected by situations in which the system draws at random identical blades denoting different atomic objects or in which identical sentence chunks are produced from different blades. A formula estimating the number of potential counterparts of a noisy piece of information has been derived. Finally, the performance of the GA model has been compared with that of BSC and HRR models using sentences of various complexity.


\begin{acknowledgement}
I would like to thank Rafa\l{} Ab\l{}amowicz and Ian Bell for many inspiring conversations that took place during the AGACSE conference in Grimma in August 2008. Furthermore, my participation in that conference would not have been possible without the support from Centrum Leo Apostel (CLEA) at the Vrije Universiteit in Brussels. I also acknowledge support from the LFPPI network.
\end{acknowledgement}

\end{document}